\documentclass{article}

\usepackage[final,nonatbib]{neurips_2019}
\usepackage[numbers]{natbib}

\usepackage[utf8]{inputenc} 
\usepackage[T1]{fontenc}    
\usepackage{hyperref}       
\usepackage{url}            
\usepackage{booktabs}       
\usepackage{amsfonts}       
\usepackage{nicefrac}       
\usepackage{microtype}      
\usepackage{listings}       
\usepackage{float}
\usepackage{graphicx}
\usepackage{color}
\usepackage{amsmath}

\title{GANspection}

\author{
  Hammad A. Ayyubi \\
  Department of Computer Science\\
  University of California, San Diego\\
  \texttt{hayyubi@ucsd.edu} \\
}

\begin{document}

\maketitle
\begin{abstract}
  Generative Adversarial Networks (GANs) have been used extensively and quite successfully for unsupervised learning. As GANs don't approximate an explicit probability distribution, it's an interesting study to inspect the latent space representations learned by GANs. The current work seeks to push the boundaries of such inspection methods to further understand in more detail the manifold being learned by GANs. Various interpolation and extrapolation techniques along with vector arithmetic is used to understand the learned manifold. We show through experiments that GANs indeed learn a data probability distribution rather than memorize images/data. Further, we prove that GANs encode semantically relevant information in the learned probability distribution. The experiments have been performed on two publicly available datasets - Large Scale Scene Understanding (LSUN) and CelebA.
\end{abstract}

\section{Introduction}

Unsupervised learning has gained a lot of traction off late and is an exponentially growing field. This is primarily because of two reasons - the sheer amount of data that we collect on a daily basis and the humongous effort required to label them. It becomes practically impossible to label millions and billions of data being collected on a daily basis. 

On one hand we have deep supervised learning models \cite{resnet}, \cite{alexnet}, \cite{maskrcnn} which work quite well, but also which which require huge amount of labelled data to be trained. On the other hand we have huge amounts of labelled data - eg. youtube videos - which are not leveraged because of the difficulty/effort of labelling them. It is this divide/gap that unsupervised learning methods \cite{dcgan} or semi-supervised learning methods \cite{semi_supervised} seek to bridge.

One of the powerful unsupervised learning methods involve learning the probability distribution of the source data. Multiple approaches have been employed to achieve such a learning task. One set of methods seek to explicity learn the probability distribution of data. These approaches include Fully visible belief networks \cite{belief_net}, \cite{wavenet}, Variational Autoencoders (VAE) \cite{vae}, \cite{vae2} and Boltzman Machine \cite{boltzman} among others. The other set of approach seeks to learn the probability distribution implicitly. Generative adversarial networks (GANs) \cite{gan} fall into this category. GANs learn this data distribution implicitly by constantly sampling from the learned distribution and ensuring that the sample looks like the real data.

GANs \cite{gan} have achieved remarkable success in tasks such as unsupervised learning \cite{dcgan}, domain adaptation \cite{cycada}, style transfer \cite{stargan}, realistic sample generation \cite{biggan} etc. Despite the growing application of GANs, it's difficult to interpret the distribution being learned as they don't approximate any explicit probability distribution. We seek to gain this understanding by inspecting the learned latent representations. More specifically, we will investigate the manifold learned by the generator of DCGAN \cite{dcgan}. We will also explore the generic nature of these representation by applying the same inspection methods on PgGAN \cite{pggan}.

To summarize, we make the following contributions through this work:

\begin{itemize}
    \item We inspect the manifold learned by the generator of DCGAN through various interpolation and extrapolation techniques to figures out whether GANs are indeed learning some useful representations or simply memorizing images.
    \item We inspect the kind of semantic relations encoded in the learned manifold by checking the result of various vector arithmetic operations.
    \item We also perform similar experiments of PgGAN to check whether such representations and semantic relations generalize across GANs or not.
\end{itemize}

\begin{figure}[H]
    \includegraphics[width=\textwidth, height=250pt]{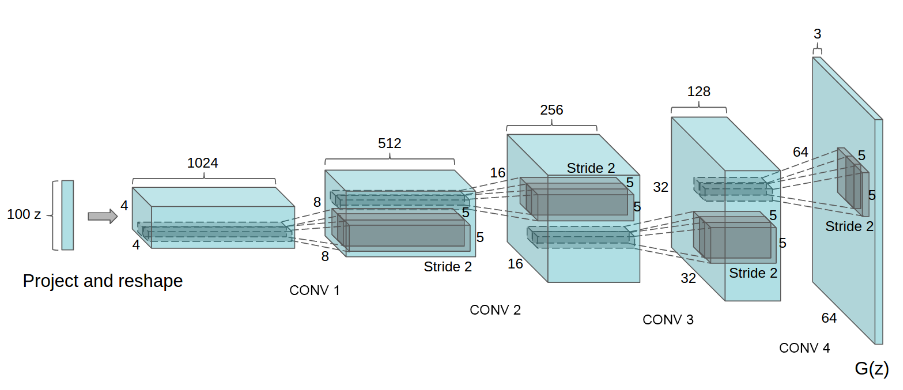}
    \caption{DCGAN architecture}
    \label{dcgan}
\end{figure}

\section{Related Work}

\subsection{Generative Adversarial Networks}
Generative Adversarial Networks (GANs) \cite{gan} are a powerful pool of generative models which have been widely successful in generating realistic samples. GANs approximate the data probability distribution implicitly and constantly draw samples from it. The ultimate objective is to make this drawn sample indistinguishable from the real samples. The discriminator tries to distinguish real samples from the fake ones and the generator tries to fool the discriminator to pass the counterfeit ones as real. Thus, they work on the principles of game theory where one player tries to better the other. 

\subsection{DCGAN}
GANs always used convolutional neural net but they could never generate high quality realistic samples. DCGAN \cite{dcgan} for the first time showed that GANs could indeed generate high quality samples. To acheive this they made certain structural changes in the neural network, namely - using batch normalization \cite{batchnorm}, using ReLu \cite{relu} activation in generator and Leaky ReLu \cite{leaky_relu} in the discriminator and replacing all pooling operations with strided convolutions. The network architecture is shown in figure \ref{dcgan}.

Further, they inspected the learned discriminator representations and latent space representation learned by the generator. Through various experiments they concluded that the generator learns meaning latent space representations encoding semantically relevant information while the learned discriminator representations is quite useful for a number of downstream supervised learning task.

\begin{figure}[H]
    \includegraphics[width=\textwidth, height=250pt]{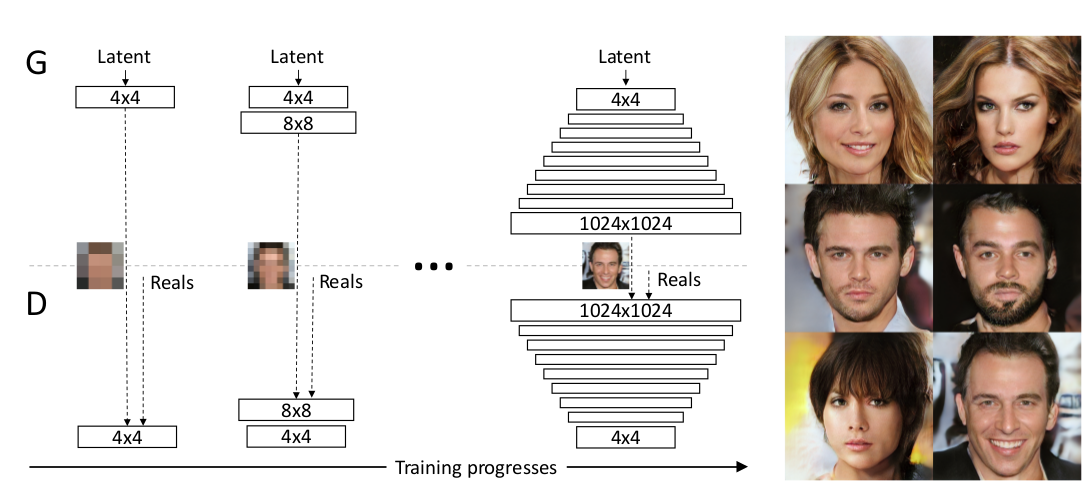}
    \caption{PgGAN architecture}
    \label{pggan}
\end{figure}

\subsection{PgGAN}

PgGANs have been immensely successful in generating high quality and high resolution images. The key idea utilized is progressive/gradual growing of the model complexity while training. At the beginning of training, both the generator and discriminator have only a few layers and thus only model low spatial resolution. As the training progresses, more layers are added to both the generator and discriminator network. As such, they start modeling finer details at a much higher resolution. The author argues that this approach not only speeds up the training process but also greatly stabilizes it. The network architecture is shown in figure \ref{pggan}.

\section{Method/Approach}

\subsection{DCGAN}

We will first inspect the latent representation learned by DCGAN. We employ the following methods to investigate it.

\subsubsection{Circular interpolation}

The paper \cite{dcgan} argues that meaningful representation are learned in the latent space since a linear interpolation between two random latent vector generates smooth transition between generated images. We seek to ask whether a similar smooth transition holds if we interpolate circularly instead of linearly.

    \[ x = r * \cos(\theta) \]
    \[ y = r * \sin(\theta) \]
    
where $ r $ is radius and $ \theta  \in (0,\pi) $.

The torch lua code for the above experiment is given below:

\begin{lstlisting}
noiseL = torch.FloatTensor(100):uniform(-1, 1)
noiseR = torch.FloatTensor(100):uniform(-1, 1)

theta = torch.linspace(0, math.pi, 16)
mid = (noiseL + noiseR ) / 2
rx = mid[1]
ry = mid[2]
for i = 1, 16 do
    x = rx + math.cos(theta[i])
    y = ry + math.sin(theta[i])
    noise_cur = noiseL * (1-x) + noiseR * x
    noise_cur[1] = x
    noise_cur[2] = y
    noise:select(1, i):copy(noise_cur)
end

\end{lstlisting}

\subsubsection{Extrapolation}

Similar to circular interpolation, we seek to investigate what kind of samples are generated if we extrapolate instead of interpolation.

The torch lua code for this experiments is provided below:

\begin{lstlisting}
noiseL = torch.FloatTensor(100):uniform(-1, 1)
noiseR = torch.FloatTensor(100):uniform(-1, 1)

line  = torch.linspace(0, 1, 16)
for i = 1, 16 do
	if i <= math.floor(16/2) then
	    el = -line[i]
	else
	    el = 1 + line[i]
	end
        noise:select(1, i):copy(noiseL * el + noiseR * (1 - el))
    end

\end{lstlisting}

We extrapolate on one side for half the samples and for the other half we extrapolate on the other side.

\subsubsection{Vector Arithmetic}

The paper \cite{dcgan} shows two kind of vector relations - smiling and wearing glasses - hold across men and woman through vector arithmetic. We seek to find what other relations - eg. blonde hair - hold true.

The generic approach followed is described as below:

\begin{enumerate}
    \item The attribute to be inspected is found by vector subtraction $ A - B $, where $A$ represents presence of the concerned attribute in one gender and B vector represents absence of the same attribute in the same gender. For example, for smile attribute $A$ vector may represent smiling woman while $B$ may represent neutral woman. In this case the vector $ A - B $ gives us the smile attribute.
    \item We inspect the semantic relationship by adding a vector $C$ representing absence of the concerned attribute in the opposite gender. For example in the above example, $C$ may represent neutral man.
    \item By vector relationship, in the above example $ A - B + C $, should give us representations for smiling man.
\end{enumerate}

We follow the approach described above to inspect various attributes like wearing glasses, smiling, blonde hair and curly/brown hair.

Presented below is the torch lua code for experimenting with vector arithmetic:

\begin{lstlisting}
local Aavg = (A[1] + A[2] + A[3]) / 3
local Bavg = (B[1] + B[2] + B[3]) / 3
local Cavg = (C[1] + C[2] + C[3]) / 3

local final_noise = Aavg - Bavg + Cavg

-- generate images
images = net:forward(noise)

\end{lstlisting}

\subsection{PgGAN}

We explore the generalization ability of such learned representations and semantic relationships across different GANs. We will do so by inspecting the manifold learned by the generator of PgGAN.

We perform similar experiments on PgGAN as described above:
\begin{itemize}
    \item Interpolation
    \item Circular Interpolation
    \item Vector arithmetic
\end{itemize}

\section{Experimental Details}

\subsection{Datasets}

\begin{enumerate}
    \item Large Scale Scene Understanding Dataset (LSUN) \cite{lsun} \\
    This dataset contains scenes from 10 categories like bedroom, kitchen, living room, dining room etc. We will be using the bedroom category which has over 3 million data samples.\\
    
    \item CelebA \cite{celeba} \\
    CelebA dataset contains large scale face attributes dataset with over 200K images. There are 10,177 different identities and exactly 202,599 number of face images with different pose variations and background clutter.
    
\end{enumerate}

\subsection{Experimental setup}
\subsubsection{DCGAN}
We use the official code repository of DCGAN. The trained generator weights are provided by the authors and we use the same weights for all our experiments.

\subsubsection{PgGAN}
For PgGAN as well, we use the official code repository provided by the authors. Here as well, trained generator weights are provided and we use the same for all the experiments.

\subsection{Results}

\subsubsection{DCGAN}
In the first part, we perform our experiments on the generator learned by DCGAN.

\begin{figure}[H]
    \includegraphics[width=0.5\textwidth, height=250pt]{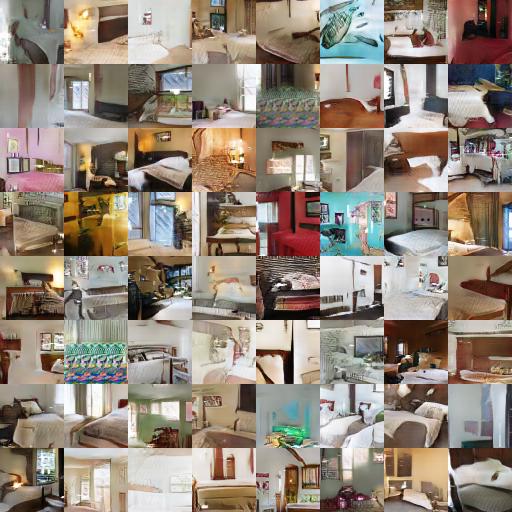}
    \includegraphics[width=0.5\textwidth, height=250pt]{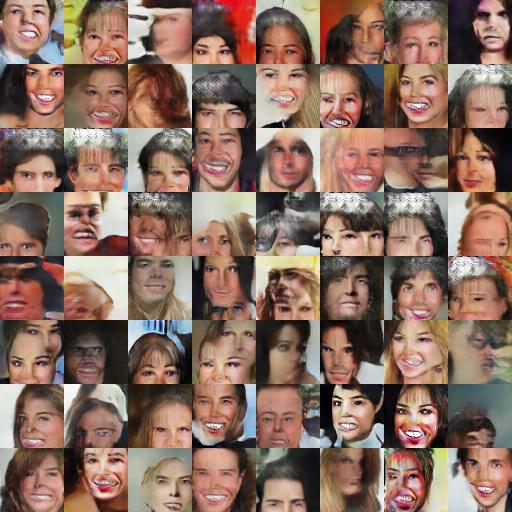}
    \caption{Left: Samples generated from the generator trained on LSUN dataset. Right: Samples generated by the generator trained on CelebA dataset.}
    \label{gen_samples}
\end{figure}

\begin{figure}[H]
    \includegraphics[width=0.5\textwidth, height=250pt]{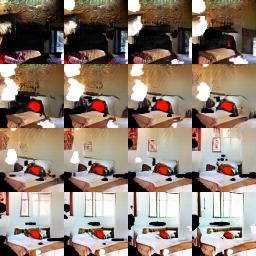}
    \includegraphics[width=0.5\textwidth, height=250pt]{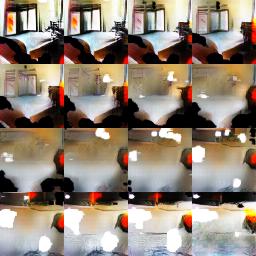}
    \caption{Two samples generated by interpolation. For each image the samples are generated by interpolating between the top left image and bottom right image.}
    \label{interpolate}
\end{figure}

\textbf{Generator Samples}

In this experiment we look at the samples generated by the generator \ref{gen_samples}. While the samples generated from the LSUN dataset looks realistic enough, the samples generated from the CelebA dataset have a lot of artifacts. This is in contrast to the samples reported in the paper \cite{gan}, which leads us to believe that the samples reported in the paper were highly selective and cherry picked.\\

\begin{figure}[H]
    \includegraphics[width=0.5\textwidth, height=250pt]{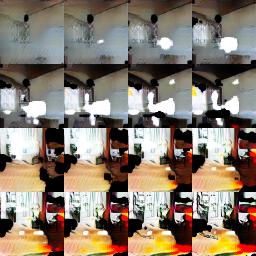}
    \includegraphics[width=0.5\textwidth, height=250pt]{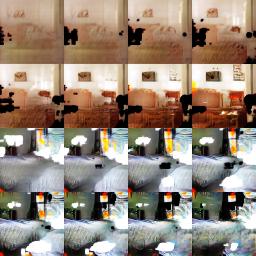}
    \caption{Two samples generated by extrapolation. For each image the samples are generated by interpolating between the top left image and bottom right image. There is a significant domain shift after half the generated samples. This is so because half of the samples were generated by extrapolating on one side and the other half was generated by extrapolating on the other side.}
    \label{extrapolate}
\end{figure}

\textbf{Interpolate}

For this experiment, we take two random vectors in the latent space and interpolate between them, generating samples at each interpolated points. \ref{interpolate} shows two such interpolation experiments. The images have been interpolated between top left and bottom right images. Although, the generated samples are not of high quality, we do see objects gradually forming and image transitioning smoothly. This gives us some proof that the latent representation learned by DCGAN has not just memorized images.\\

\textbf{Extrapolate}

For this experiment, we randomly selected two vectors in the latent space and extrapolated them linearly on either side. Figure \ref{extrapolate} shows the result of two such extrapolation experiments.

In each image, there is a significant domain shift after half the generated samples. This is because half of the samples were generated by extrapolating on the side of one vector and the other half were generated by extrapolating on the side of the other vector. This domain shift also falls nicely into our intuitive reasoning since there's a big gap between two vectors and so this domain shift is to be expected. Apart from this domain shift, all the other vectors are close to one of the randomly selected vector and we can see from the generated samples that samples generated from them make smooth transitions. This also serves up further proof that the DCGAN is indeed learning useful representations rather than memorizing images.\\

\begin{figure}[H]
    \includegraphics[width=0.5\textwidth, height=250pt]{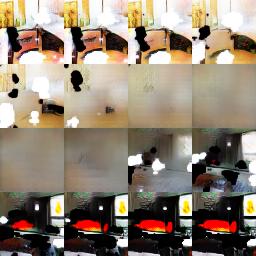}
    \includegraphics[width=0.5\textwidth, height=250pt]{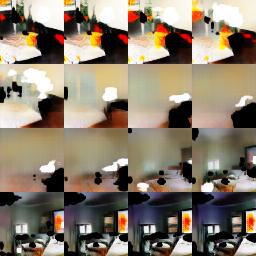}
    \caption{Two samples generated by circular interpolation. The circular interpolation is performed between top left and bottom sample in each image.}
    \label{circle}
\end{figure}

\textbf{Circular Interpolation}

For this experiment, we again selected two random vectors in the latent space. We then interpolated between them circularly in a total arc of $\pi$. The difference of the two vectors was kept as diameter and intermediate vectors were interpolated on a semi-circle over this diameter. The results can be seen in \ref{circle}.

We see that the samples indeed make a smooth transition over the arc of $\pi$. There are some indistinguishable samples generated in the middle, probably over the angle of $\pi/2$. We hypothesize it is because the latent representation might be making domain transitions over angles of $\pi/2$. Regardless, all the samples generated make smooth transitions without any abrupt change. This further leads us to believe that the latent representation learned is indeed meaningful.

\textbf{Vector Arithmetic}

\begin{figure}[H]
    \includegraphics[width=\textwidth, height=250pt]{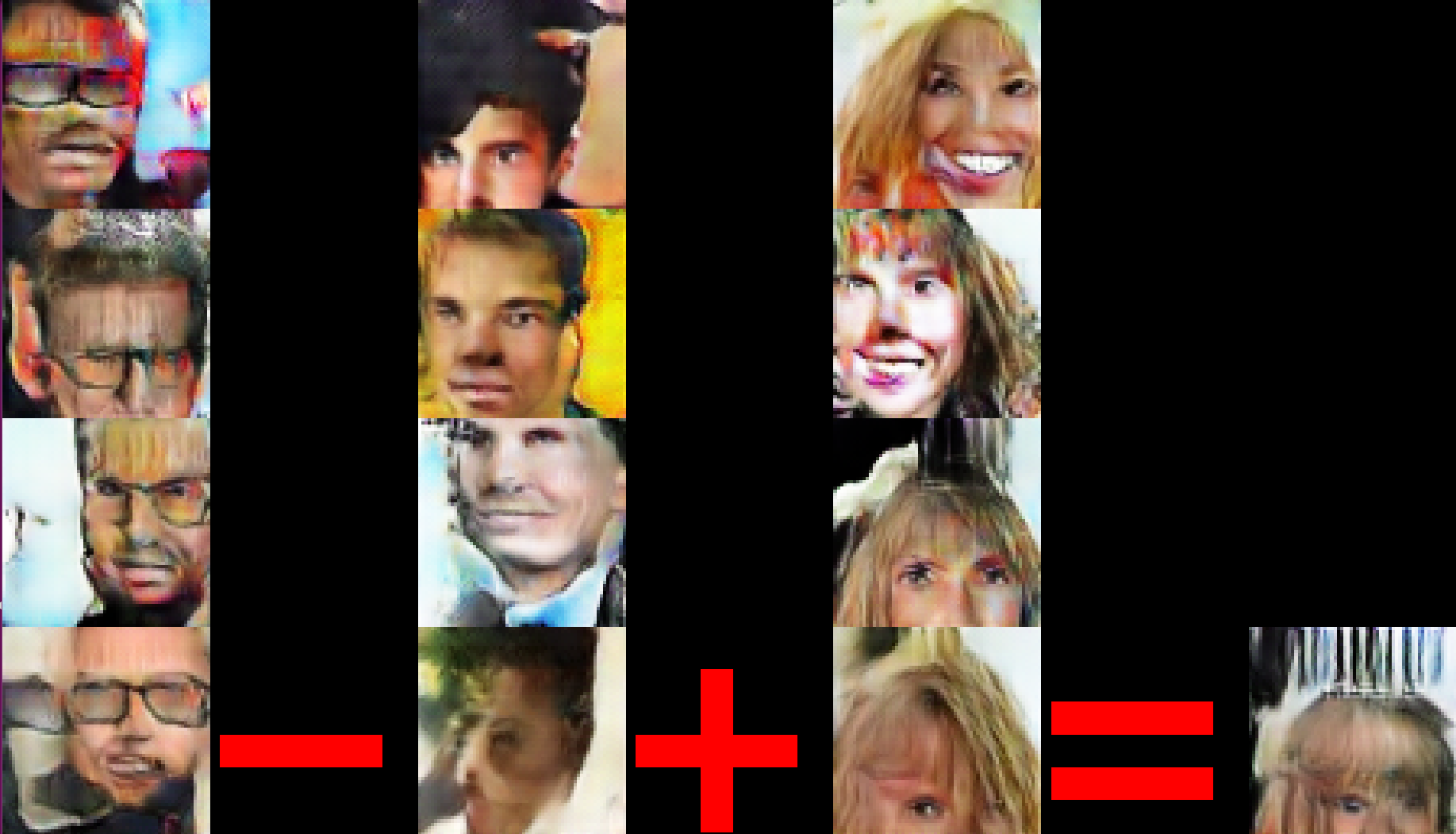}
    \caption{Vector arithmetic for wearing glasses.}
    \label{glass}
\end{figure}

\begin{figure}[H]
    \includegraphics[width=\textwidth, height=250pt]{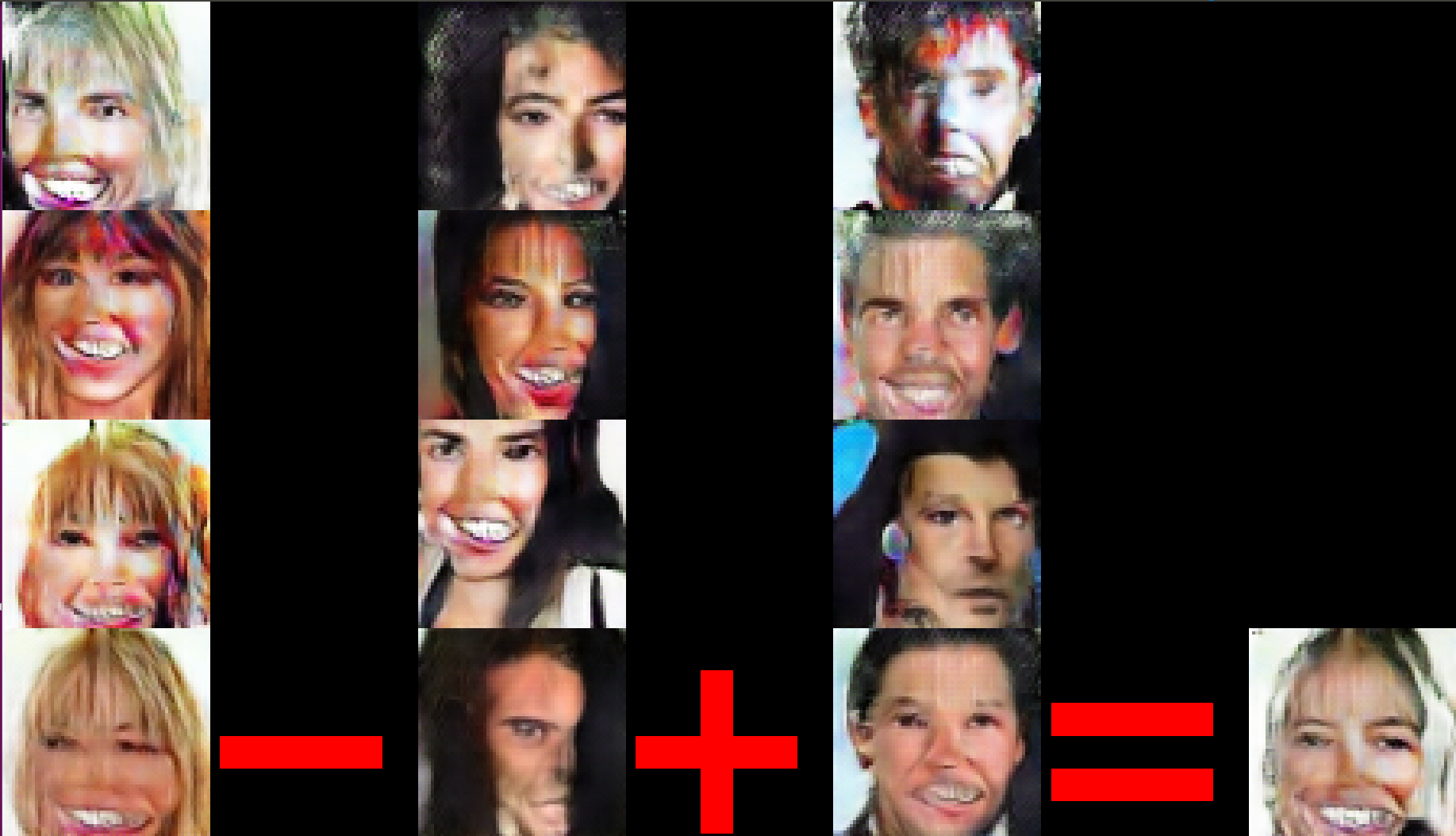}
    \caption{Vector arithmetic for having blonde hair.}
    \label{blond}
\end{figure}

We check for two kinds of vector relationship in the learned latent space - wearing glasses and having blonde hair. The results are shown in \ref{glass} and \ref{blond}.

In the first experiment latent vector representing men not wearing glasses was subtracted by the vector representing men wearing glass. This gives us the relationship of wearing glass. This relationship was added to the vector representing women not wearing glass. The resulting vector was passed through the generator and the resulting sample has some semblance of woman wearing glasses. It is difficult to say because of the poor quality of generated samples.

In the second experiment, we investigated the relationship of having blonde hair in a manner similar to one described above. The result of woman with blonde hair minus woman with black hair added to man with black hair was indeed a person with a blonde. This person does look like a man but again it's difficult to say because of the poor quality of generated samples.

This set of experiment gave us some indication that the DCGAN was indeed encoding useful semantic relationships but it's difficult to say because of the poor quality of generated smamples. We will verify these experiments with PgGAN which generates images of much higher quality.

\begin{figure}[H]
    \includegraphics[width=0.2\textwidth, height=110pt]{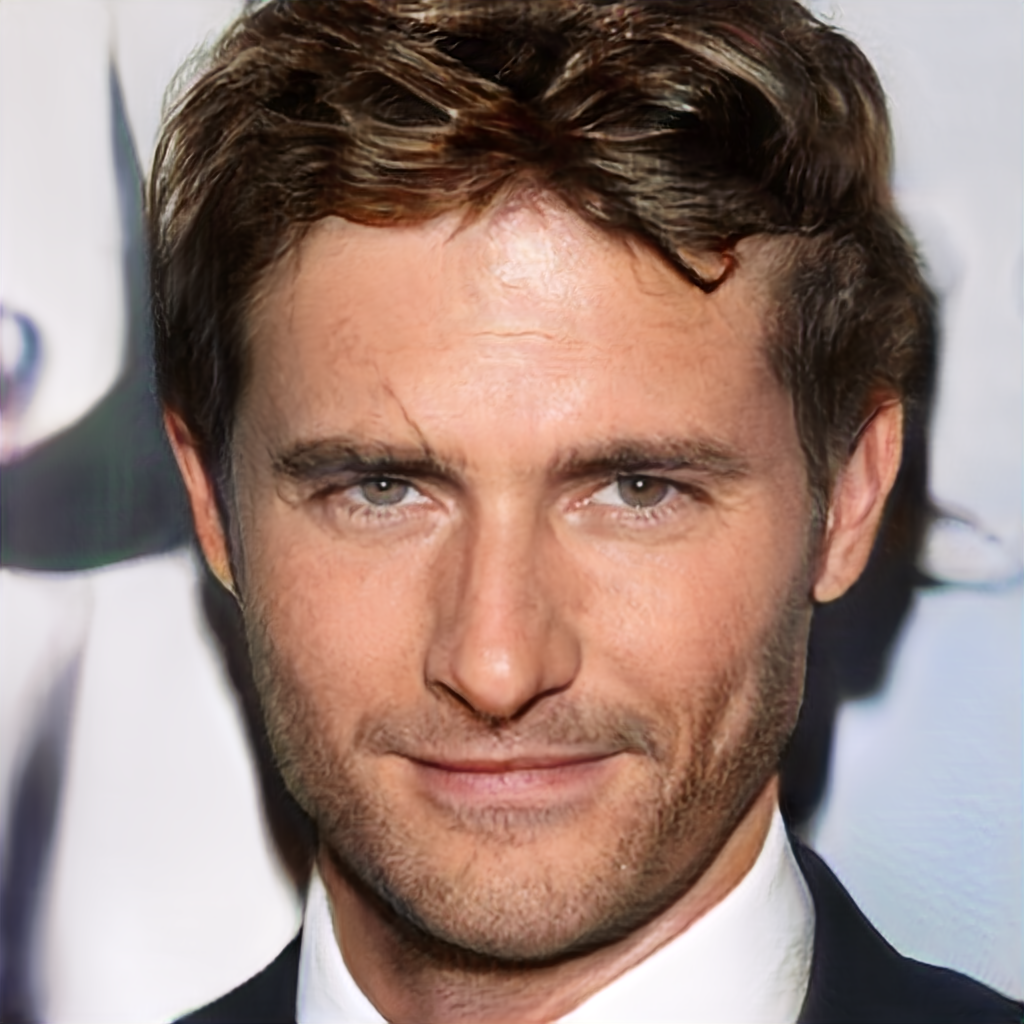}\hfill
    \includegraphics[width=0.2\textwidth, height=110pt]{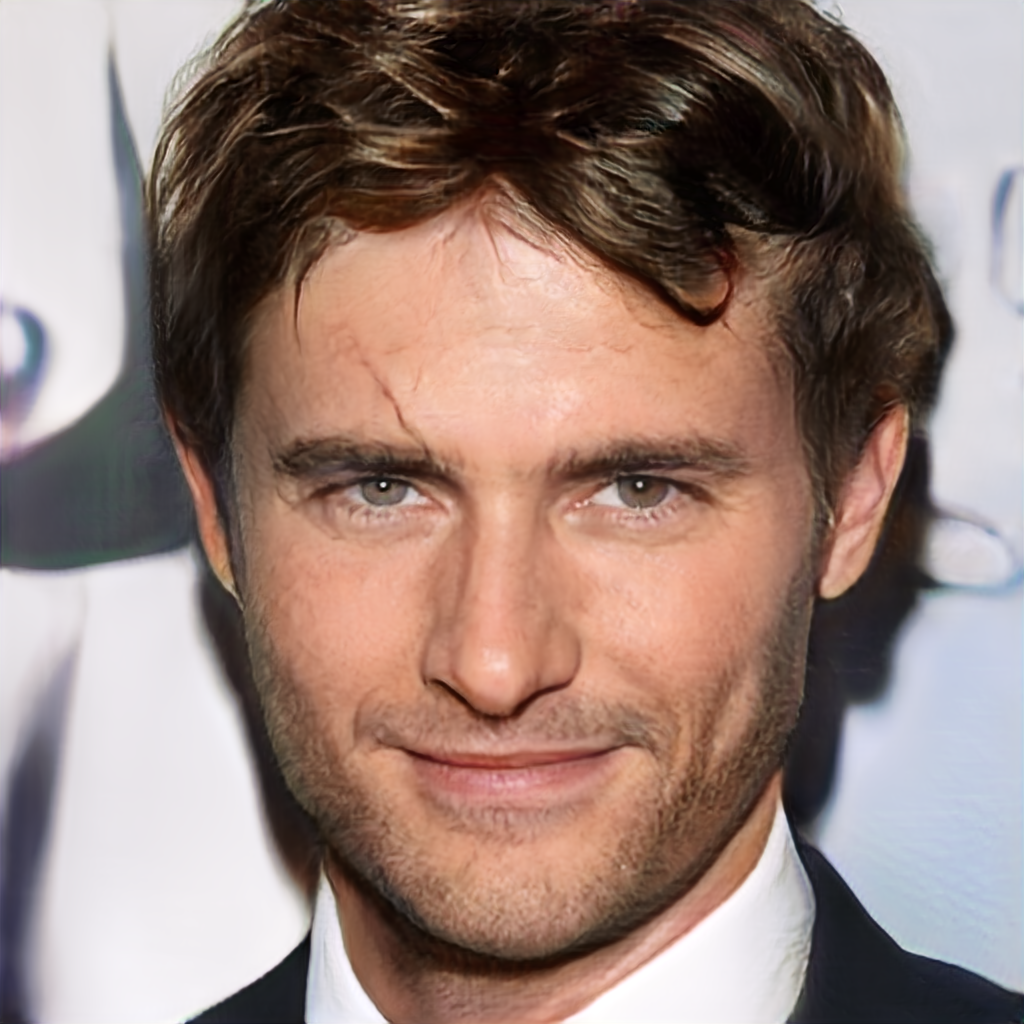}\hfill
    \includegraphics[width=0.2\textwidth, height=110pt]{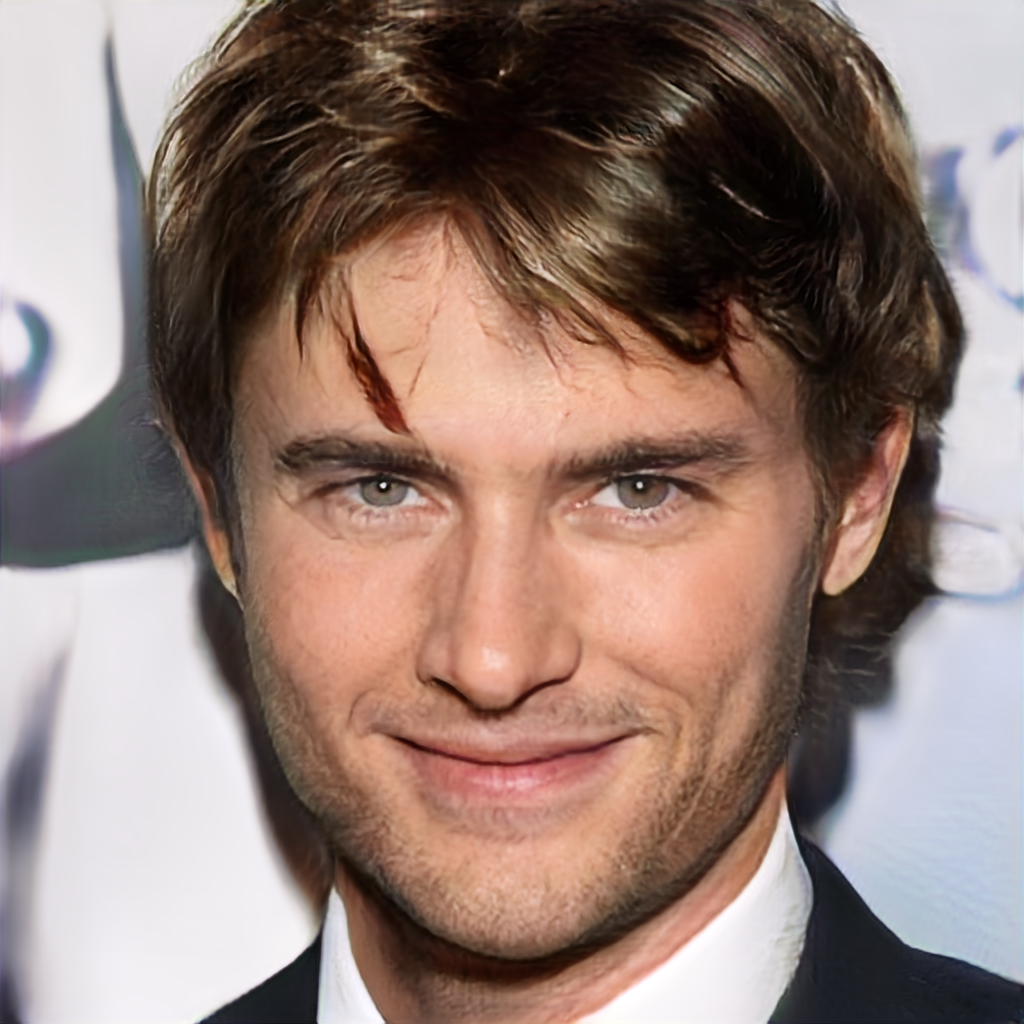}\hfill
    \includegraphics[width=0.2\textwidth, height=110pt]{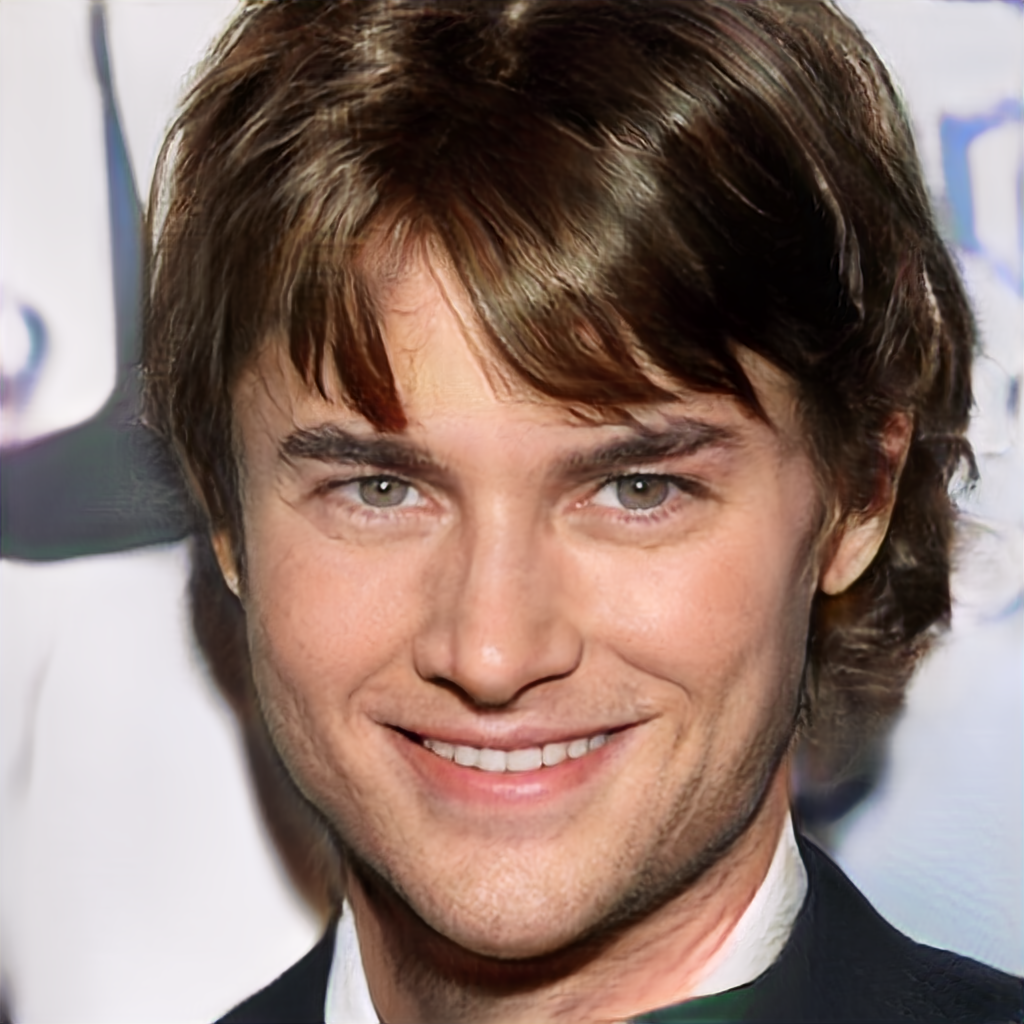}\hfill
    \includegraphics[width=0.2\textwidth, height=110pt]{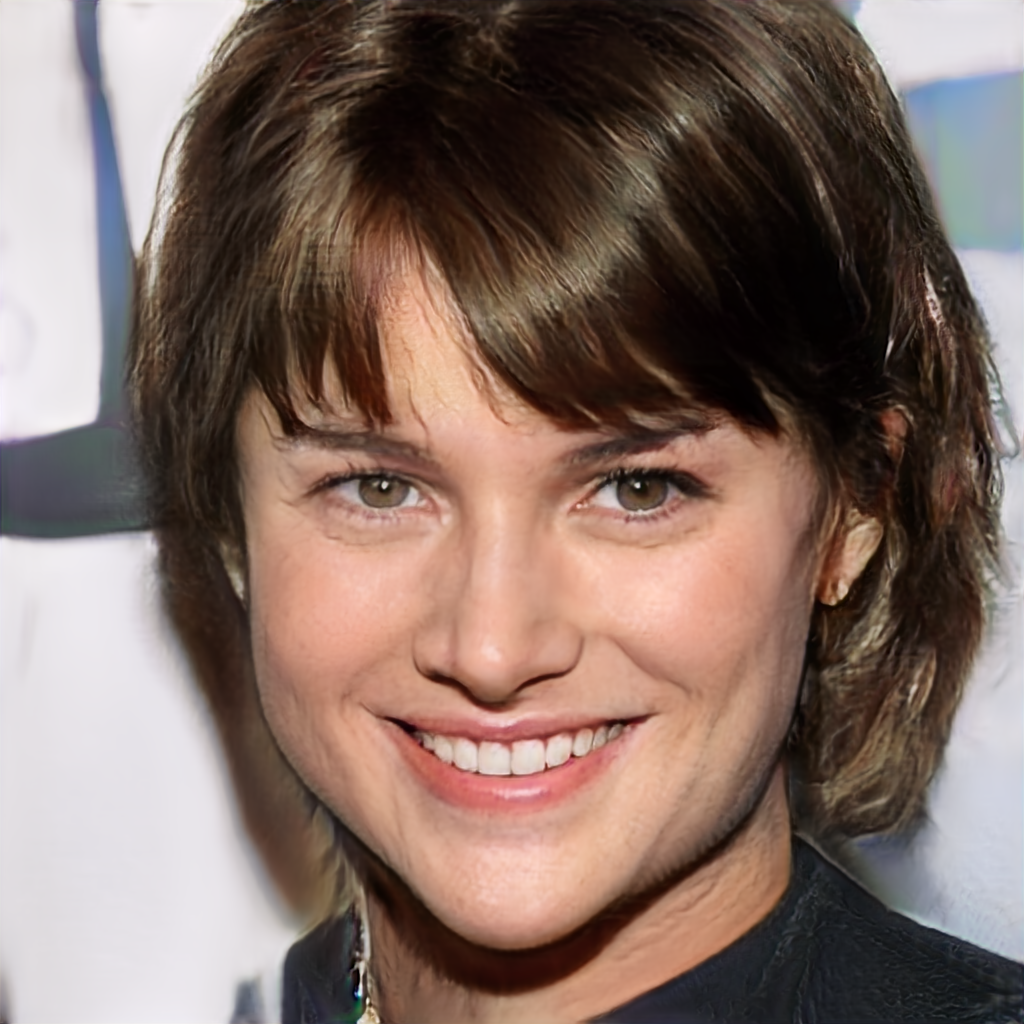}\hfill
    \includegraphics[width=0.2\textwidth, height=110pt]{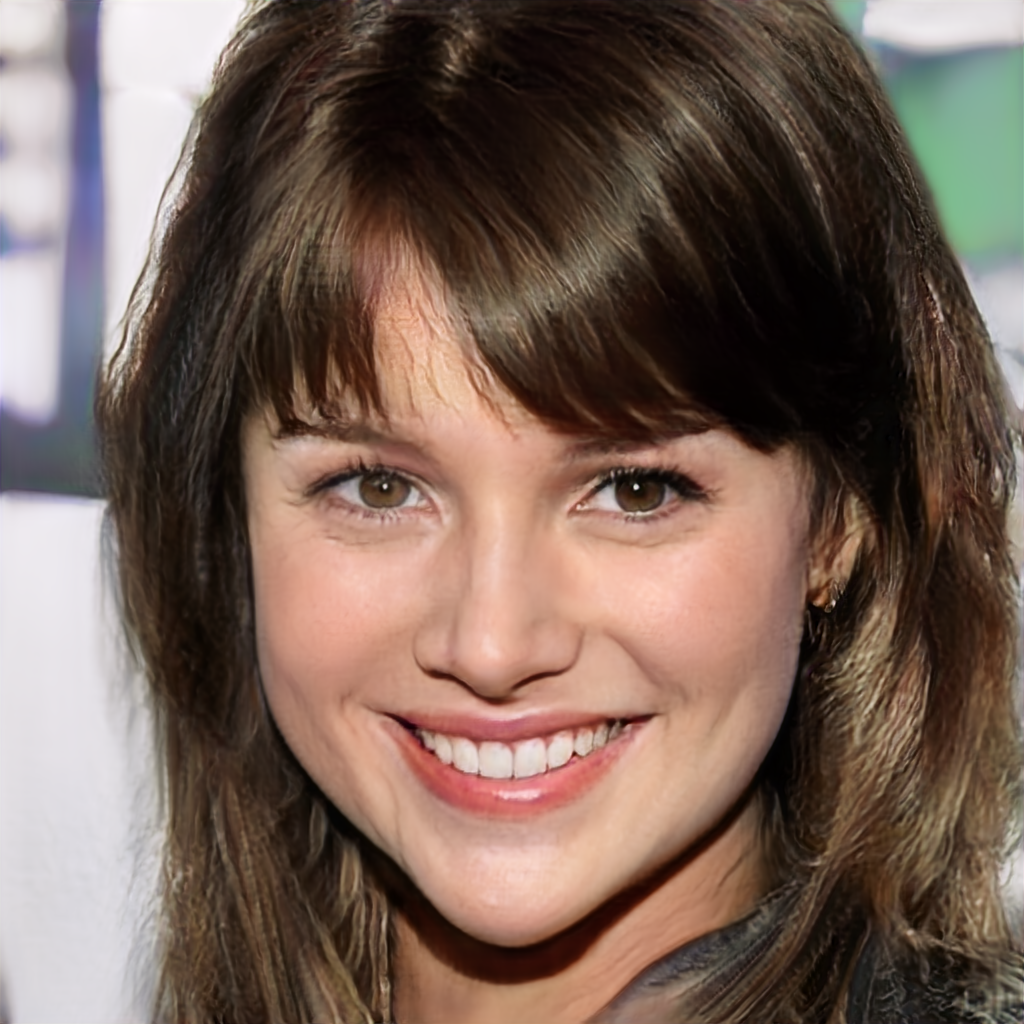}\hfill
    \includegraphics[width=0.2\textwidth, height=110pt]{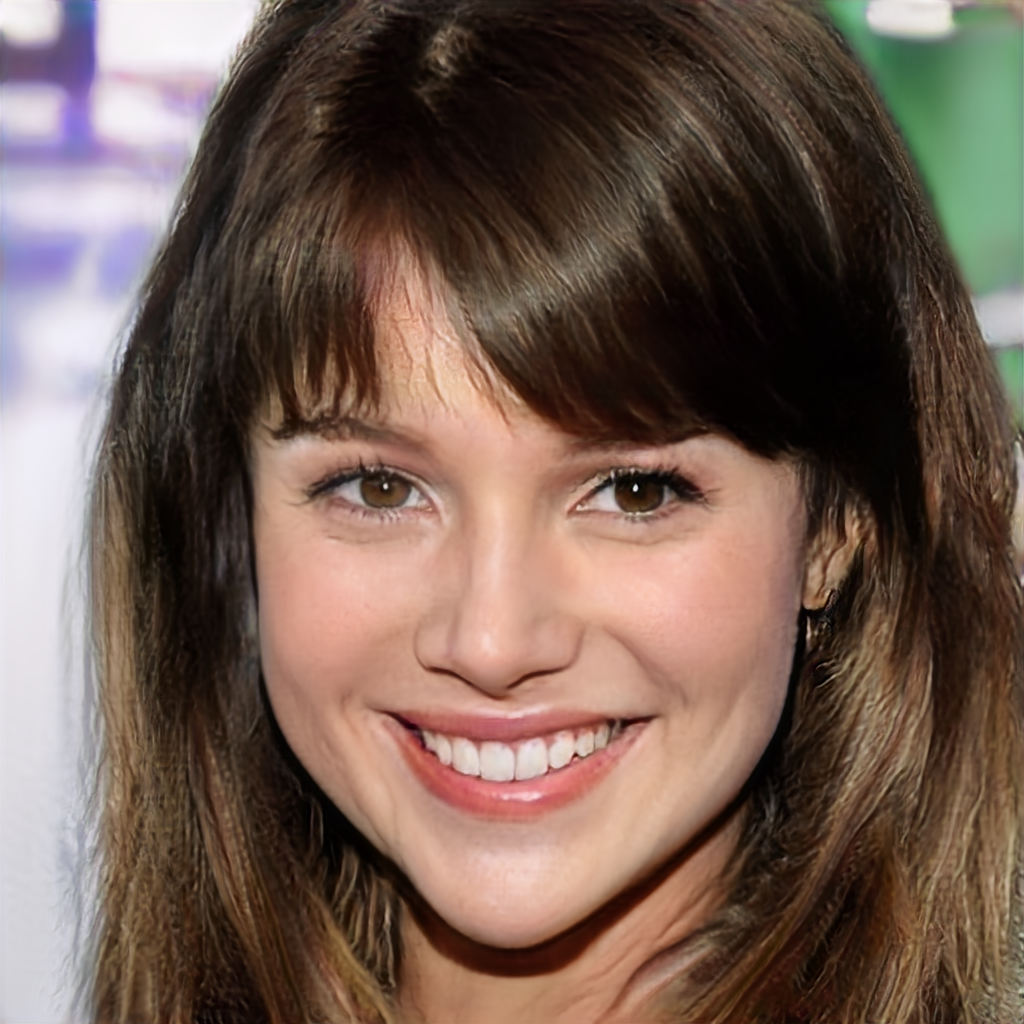}\hfill
    \includegraphics[width=0.2\textwidth, height=110pt]{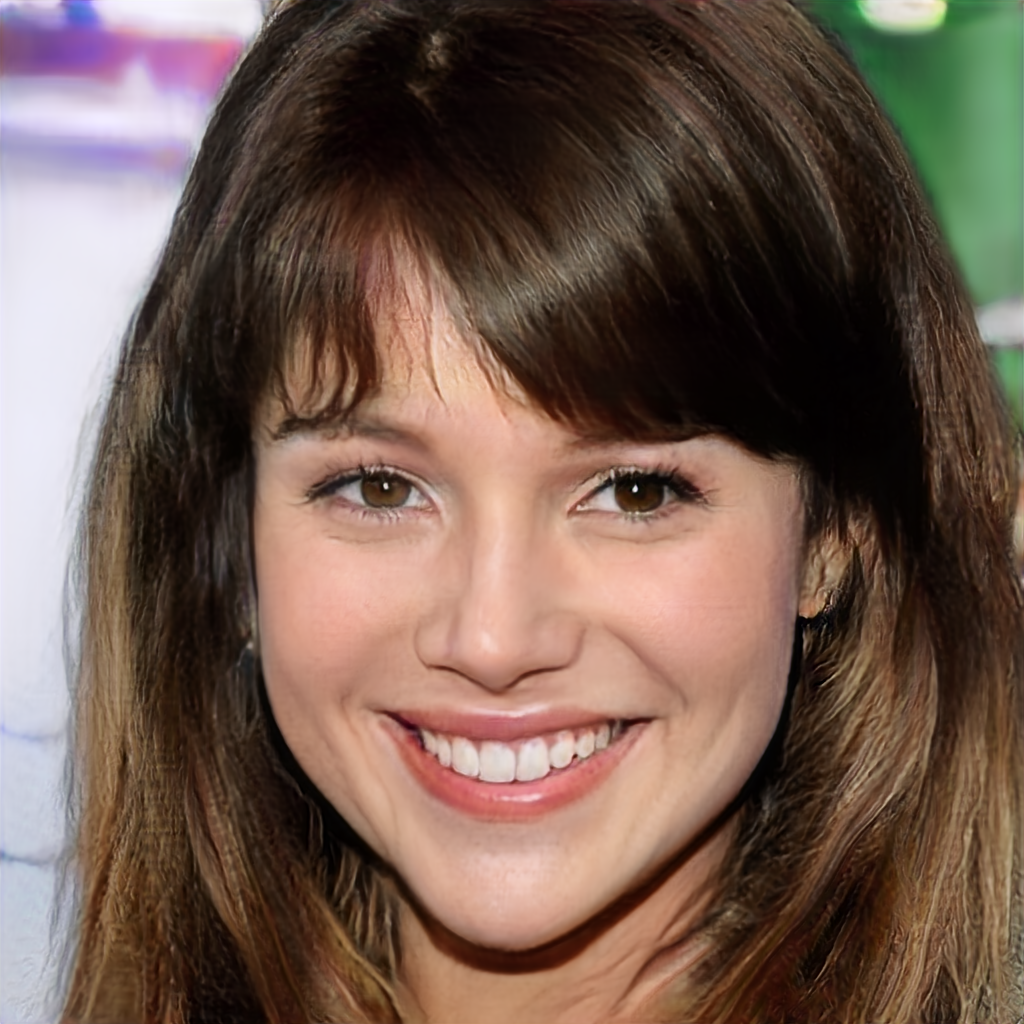}\hfill
    \includegraphics[width=0.2\textwidth, height=110pt]{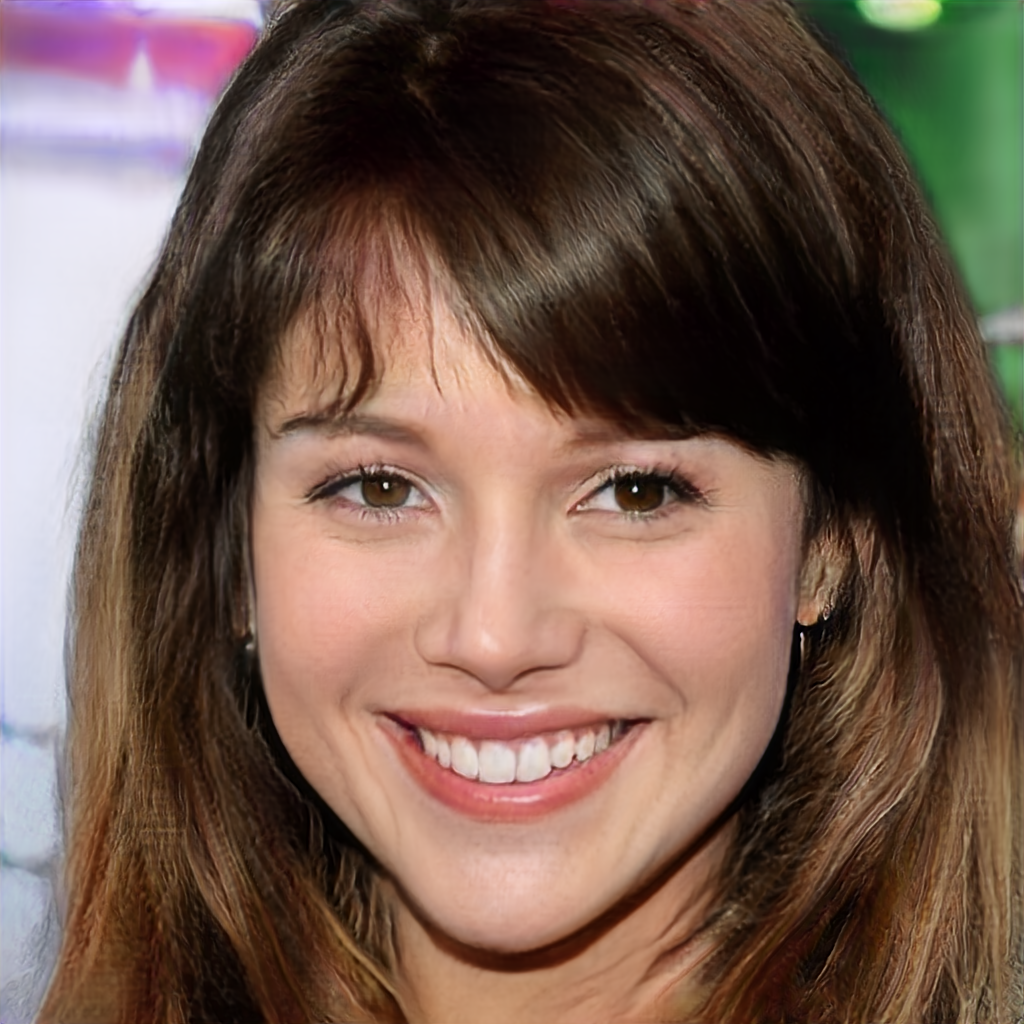}\hfill
    \includegraphics[width=0.2\textwidth, height=110pt]{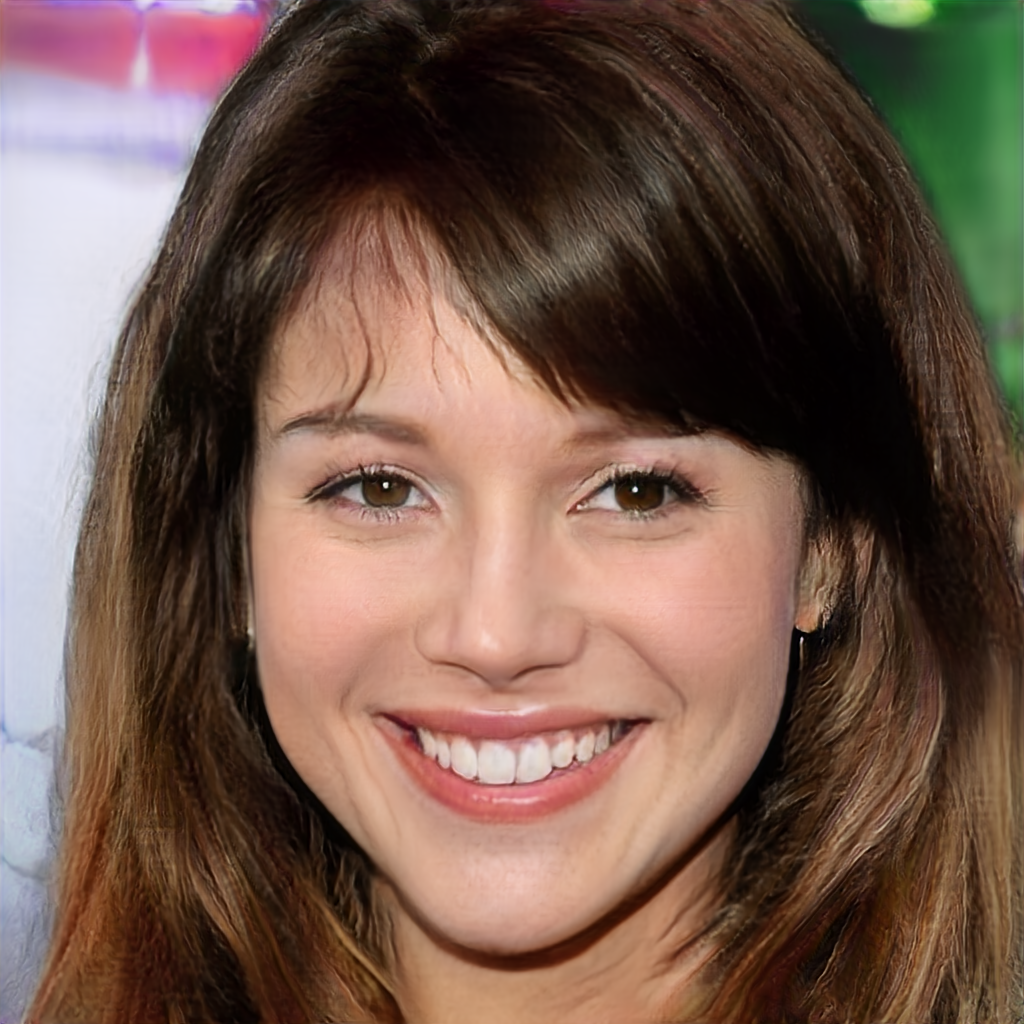}\hfill
    \caption{Interpolation from top left to bottom right generated by PgGAN.}
    \label{pggan_interpolate}
\end{figure}

\begin{figure}[H]
    \includegraphics[width=0.2\textwidth, height=110pt]{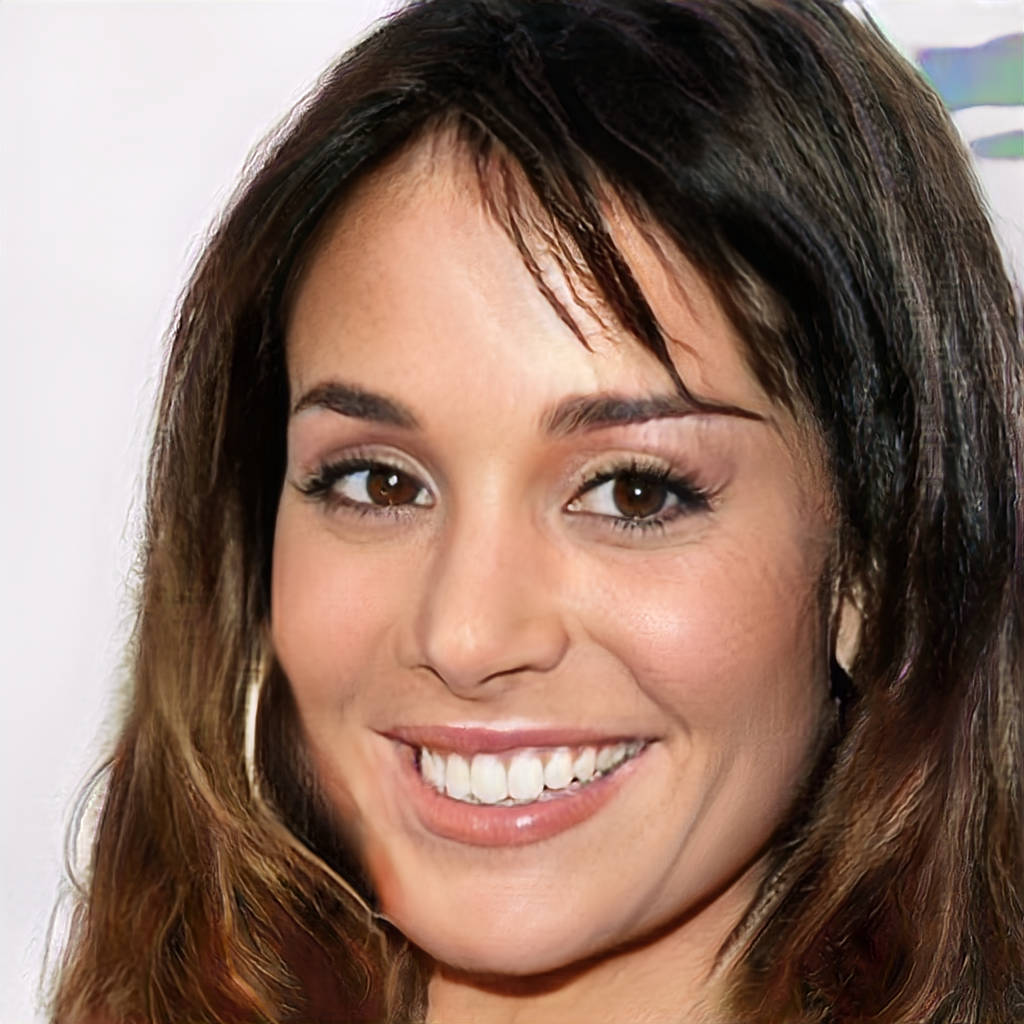}\hfill
    \includegraphics[width=0.2\textwidth, height=110pt]{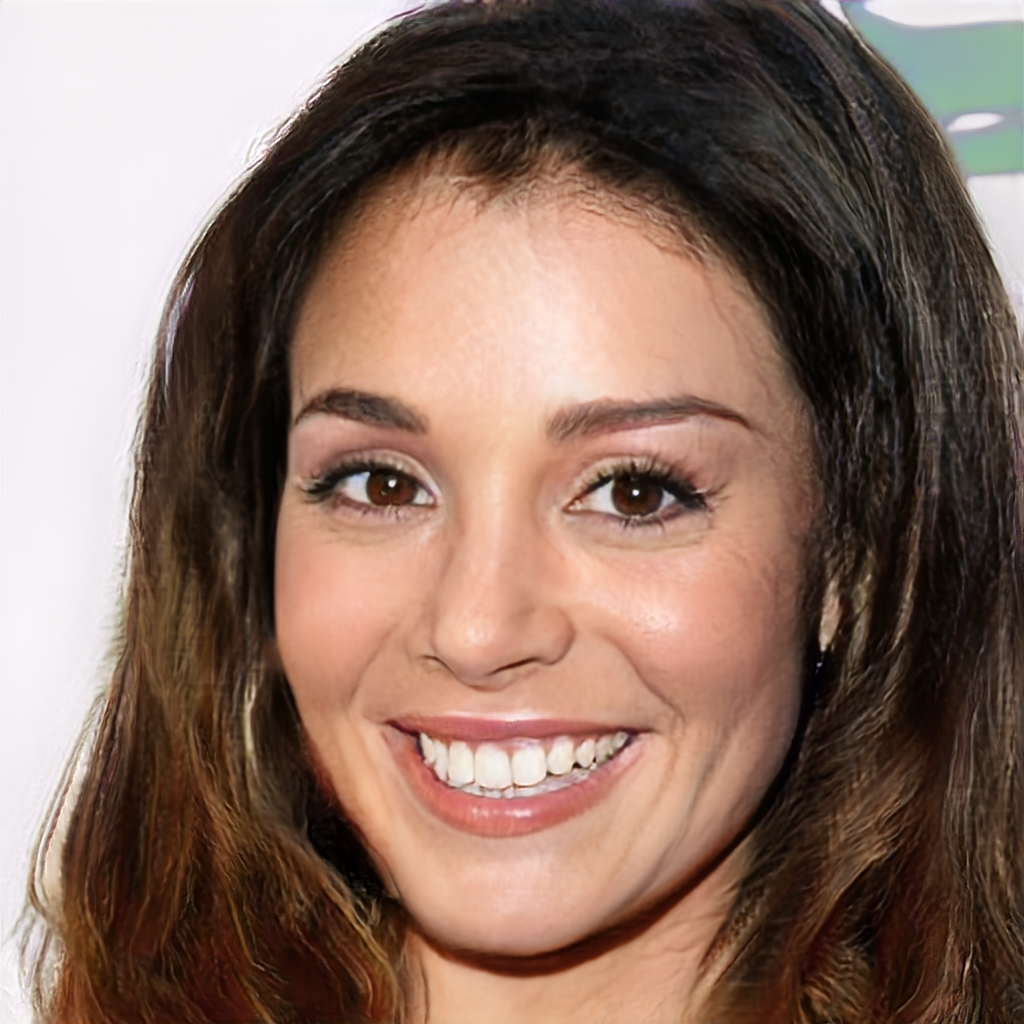}\hfill
    \includegraphics[width=0.2\textwidth, height=110pt]{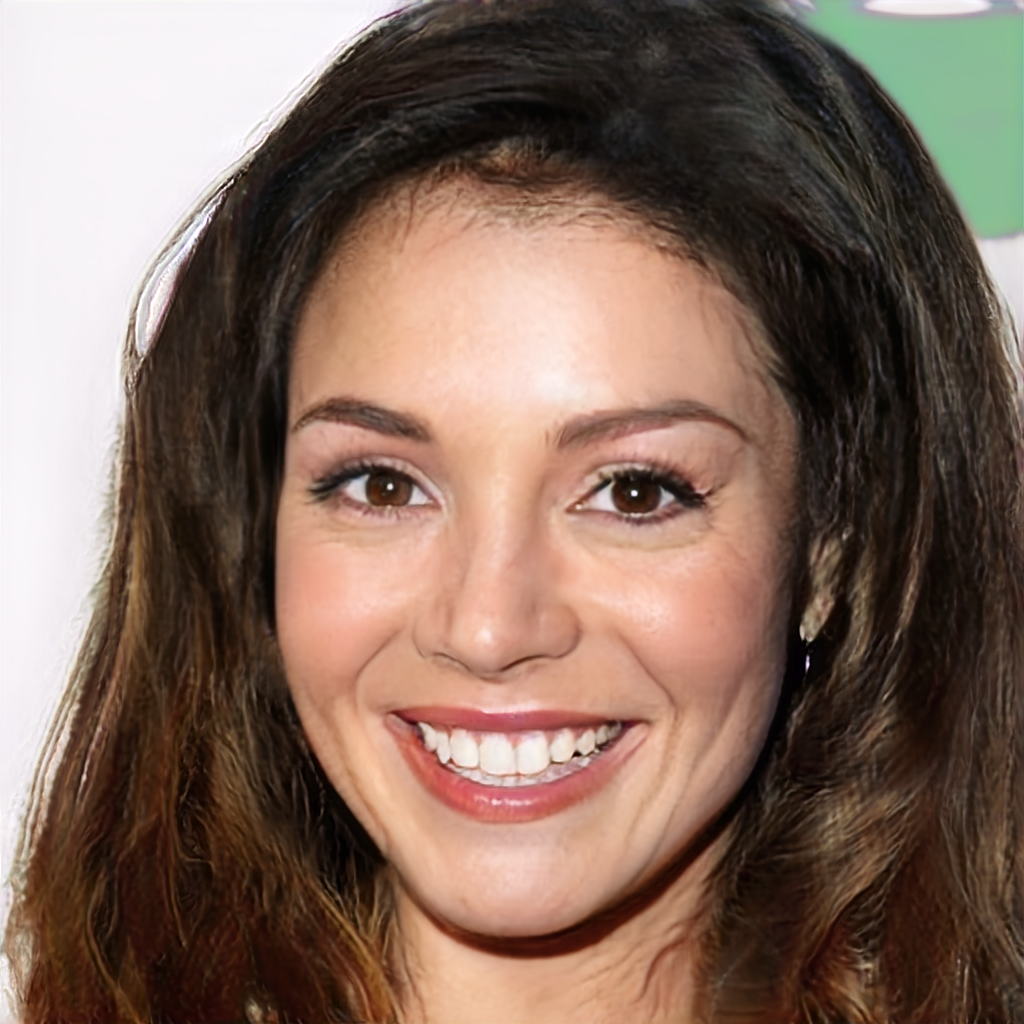}\hfill
    \includegraphics[width=0.2\textwidth, height=110pt]{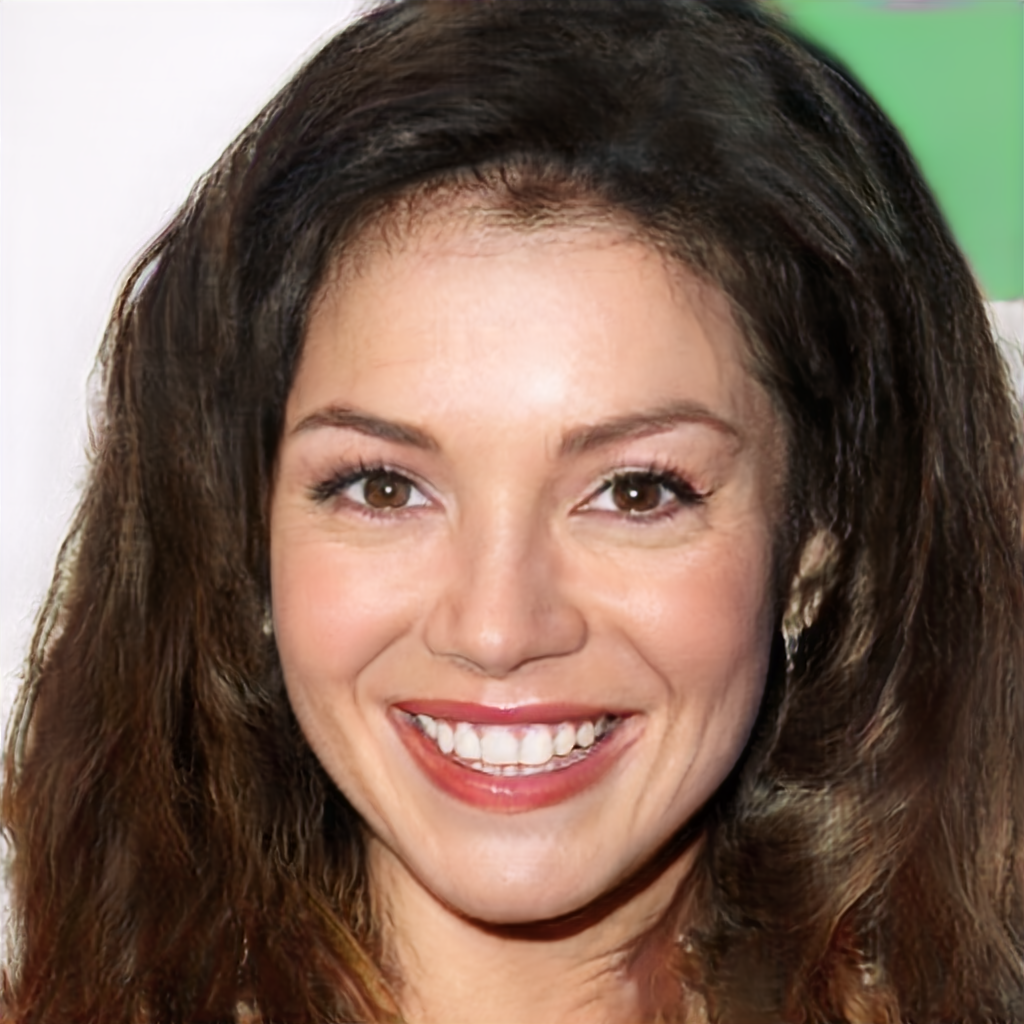}\hfill
    \includegraphics[width=0.2\textwidth, height=110pt]{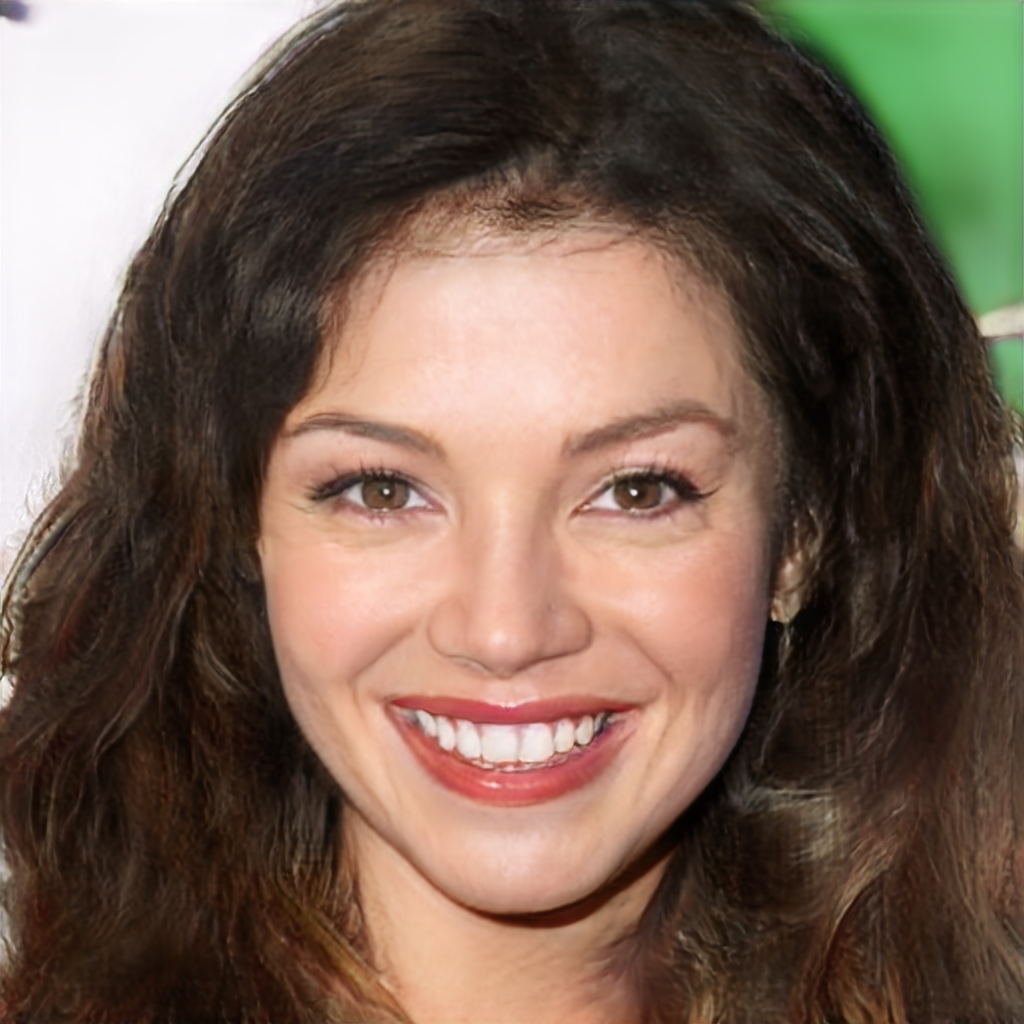}\hfill
    \includegraphics[width=0.2\textwidth, height=110pt]{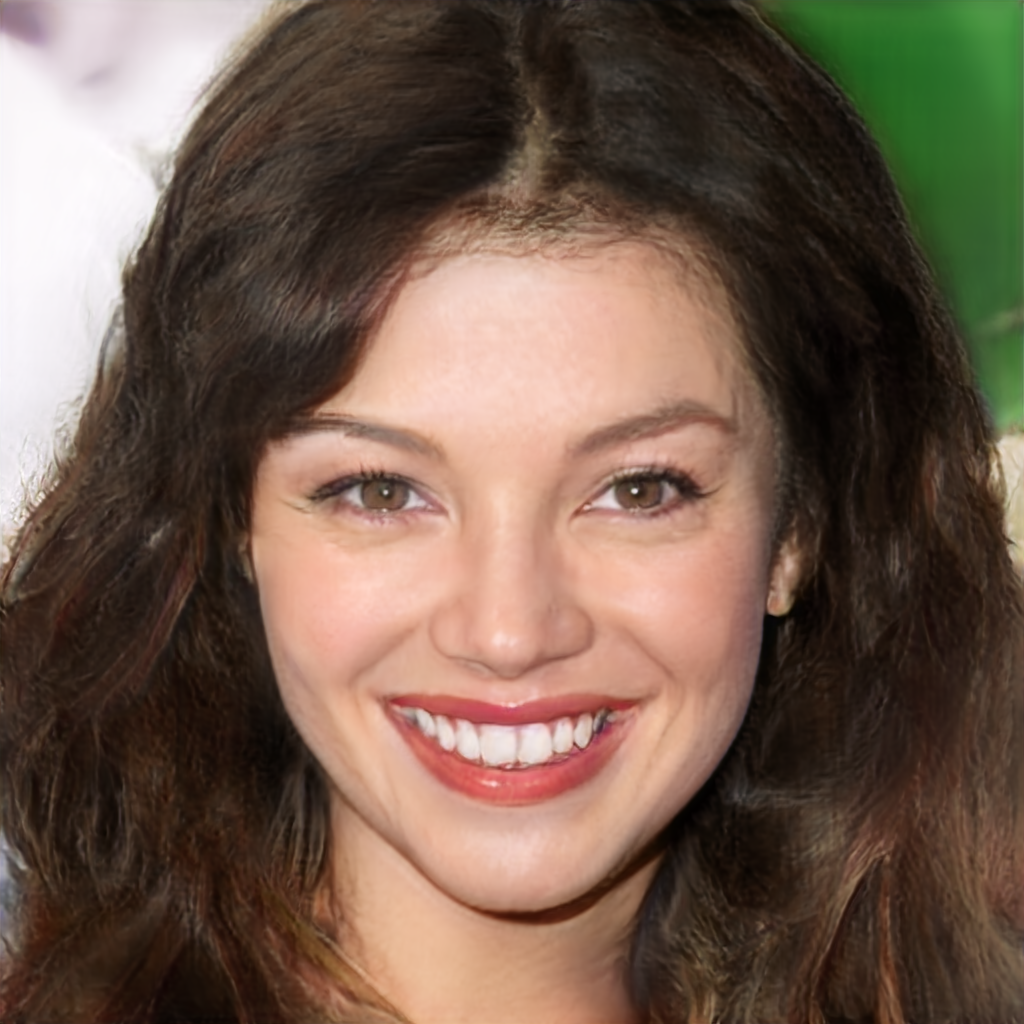}\hfill
    \includegraphics[width=0.2\textwidth, height=110pt]{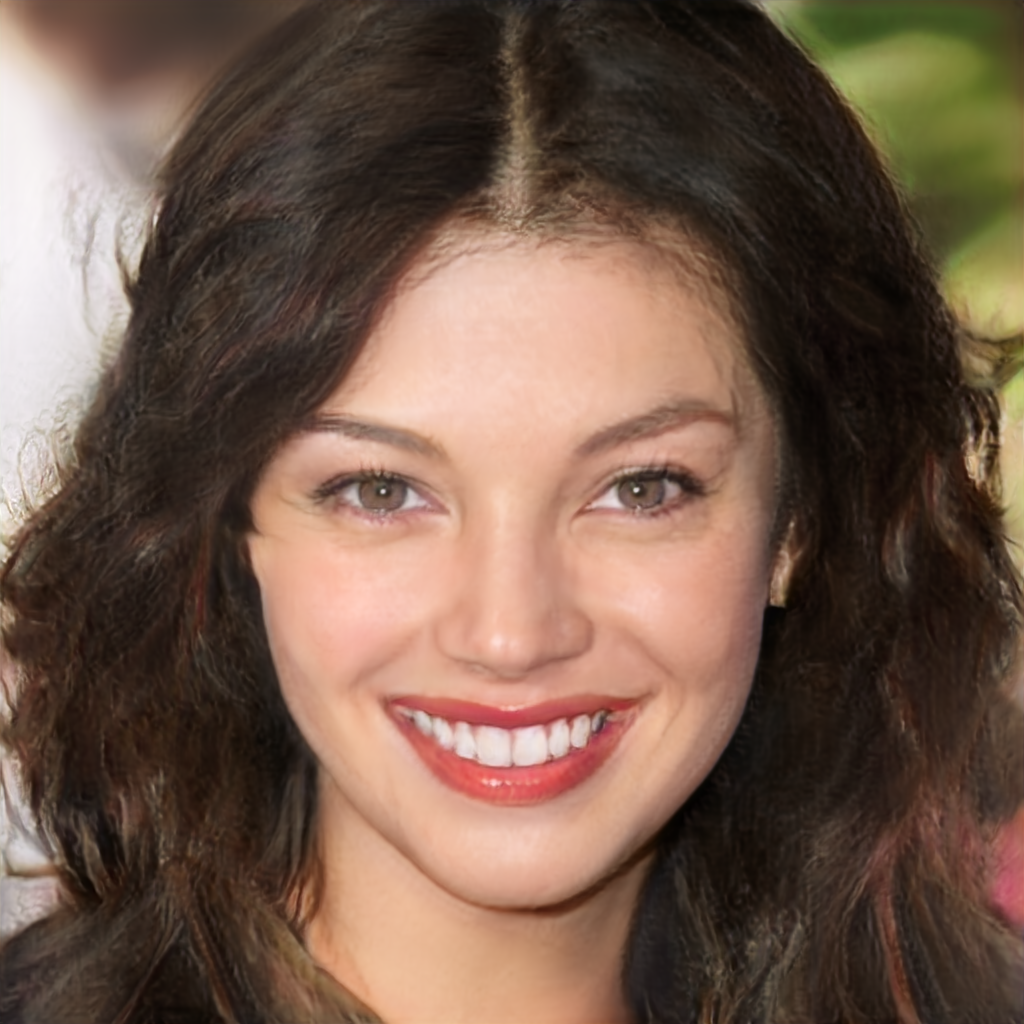}\hfill
    \includegraphics[width=0.2\textwidth, height=110pt]{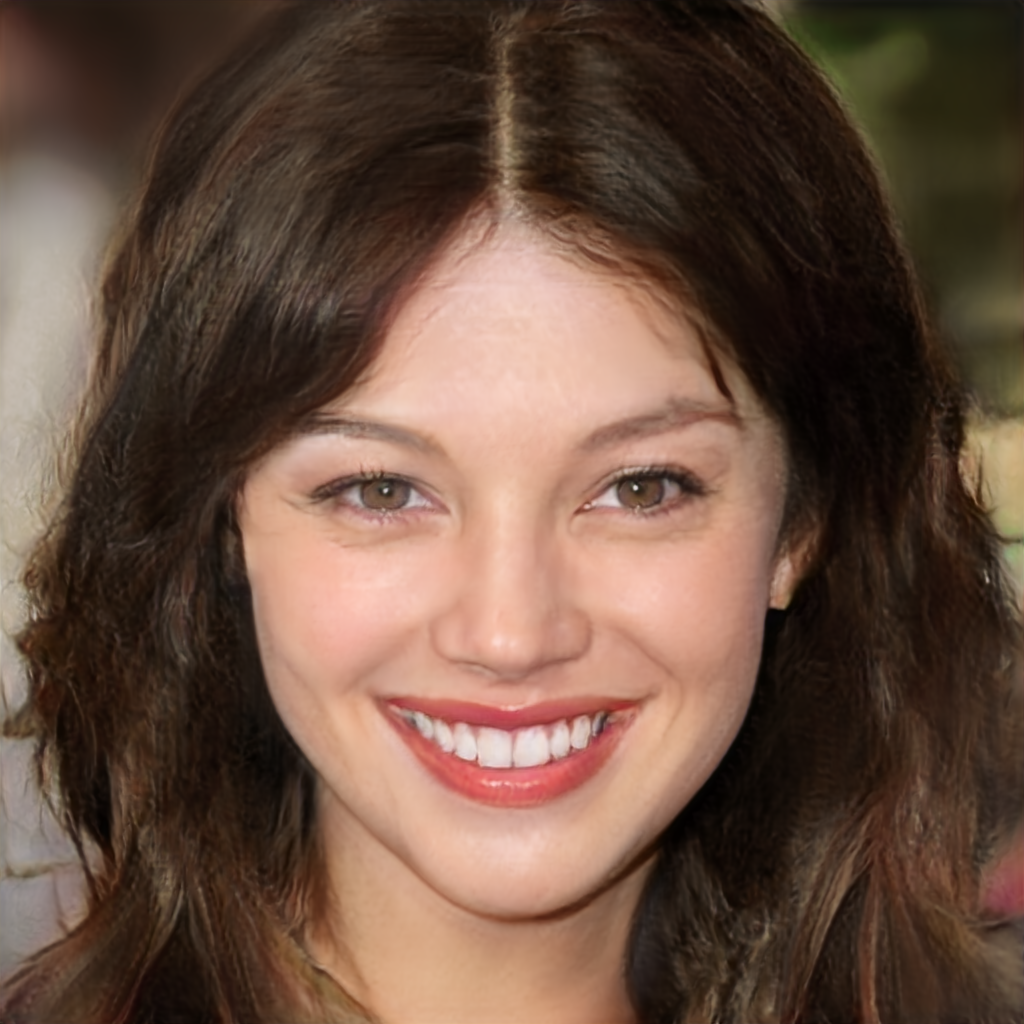}\hfill
    \includegraphics[width=0.2\textwidth, height=110pt]{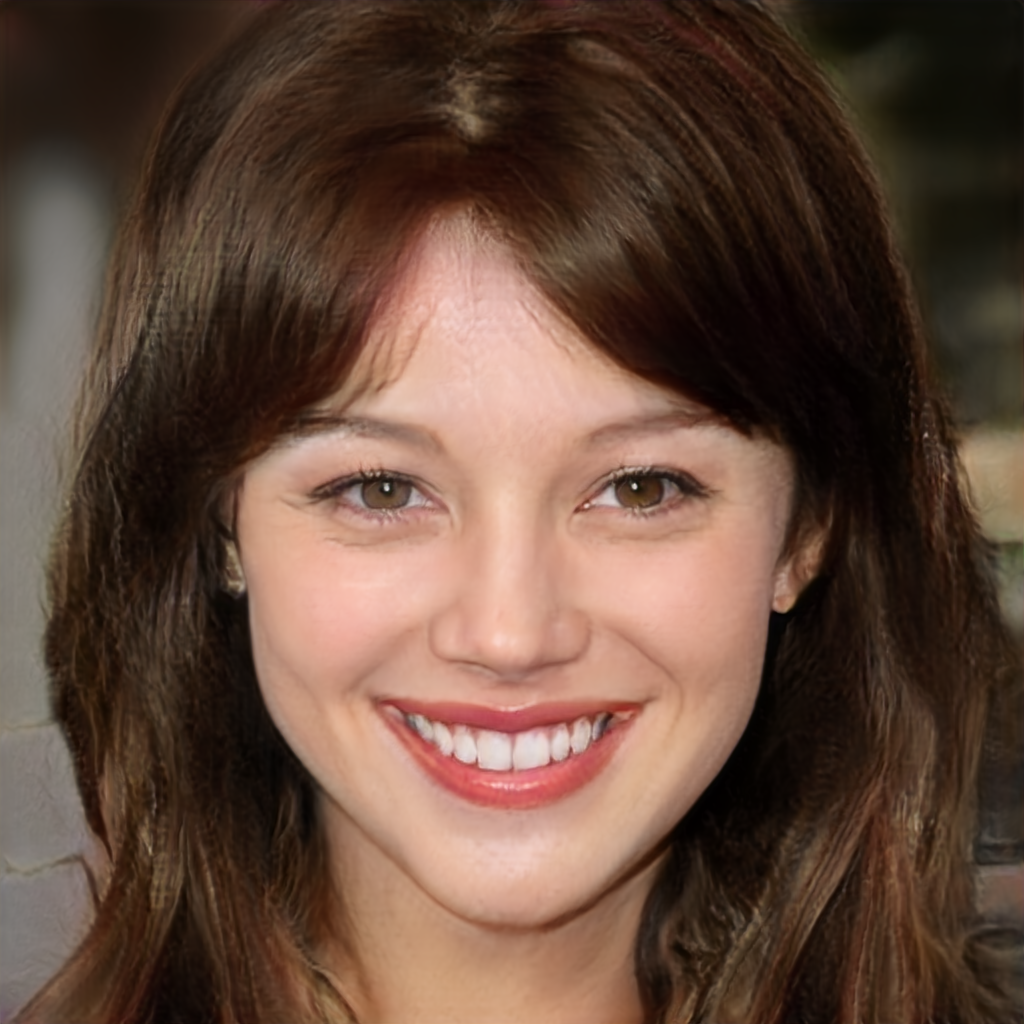}\hfill
    \includegraphics[width=0.2\textwidth, height=110pt]{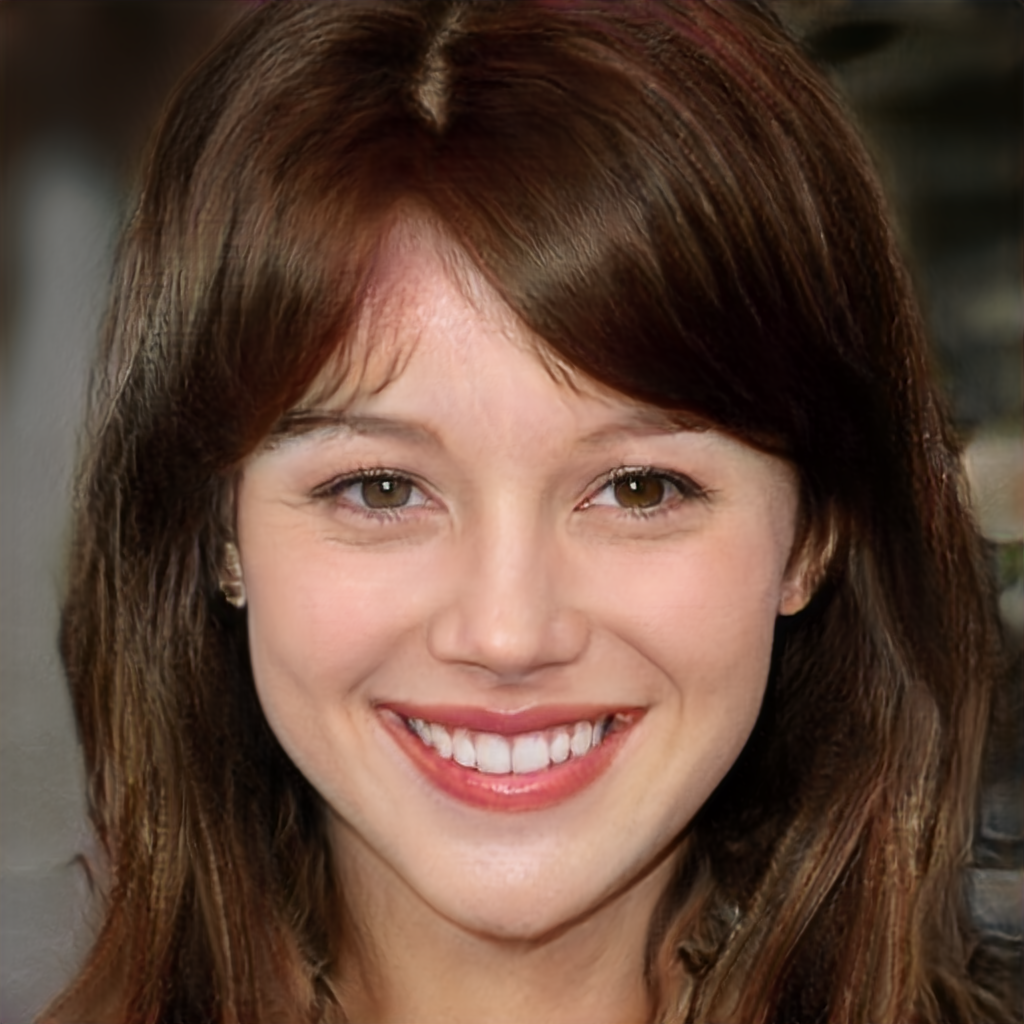}\hfill
    \caption{Circular interpolation over angle of $\pi$ from top left to bottom right generated by PgGAN.}
    \label{pggan_circle_interpolate}
\end{figure}

\subsubsection{PgGAN}

We next perform experiments on PgGAN to check whether the useful latent representations and semantic relations learnt by GANs generalize across GANs or is it a specific property of DCGAN.

\textbf{Interpolation}

The interpolation results can be visualized in figure \ref{pggan_interpolate}. We can clearly see a series of high quality samples are generated along the interpolated vectors which transition smoothly from one sample to another. This gives us proof that the latent representation learned by GANs indeed generalize across different GANs and are not a property/fluke from one GAN.

\textbf{Circular Interpolation}

The results of circular interpolation can be seen in figure \ref{pggan_circle_interpolate}. Again, we see that a smooth transition happens from top left sample to the bottom right sample. This provides further proof that GANs do learn meaningful latent representations and they generalize across different GANs.

\begin{figure}[H]
    \includegraphics[width=\textwidth, height=140pt]{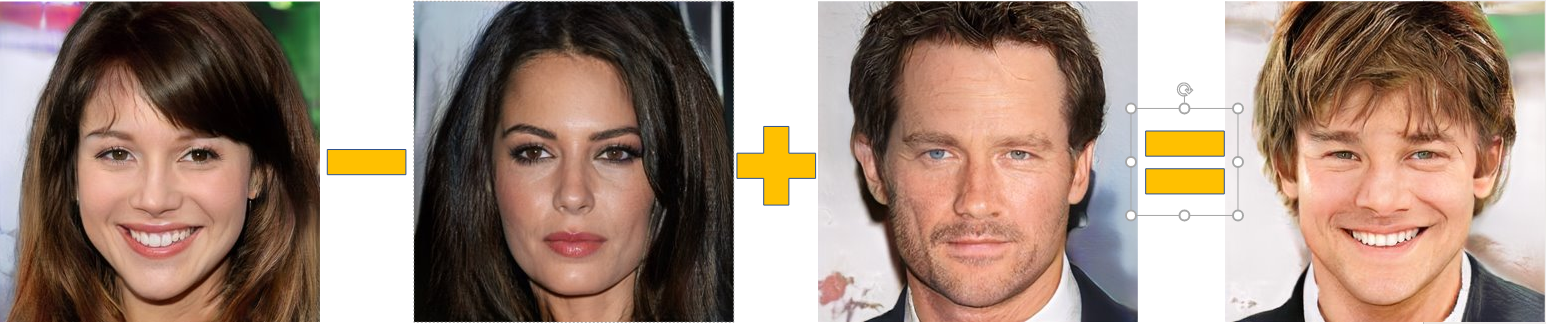}
    \caption{Vector arithmetic for the attribute smiling.}
    \label{pggan_smile}
\end{figure}

\begin{figure}[H]
    \includegraphics[width=\textwidth, height=140pt]{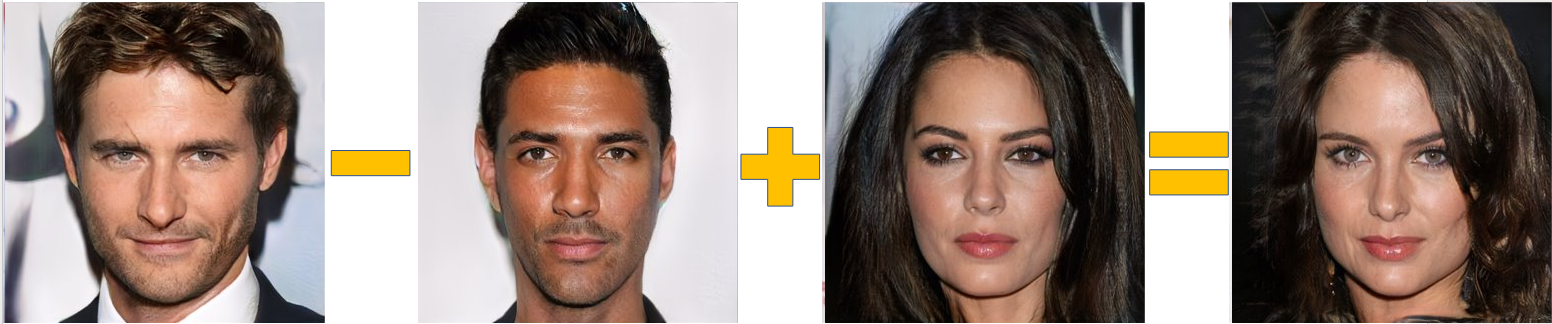}
    \caption{Vector arithmetic for the attribute of brown and curly hair.}
    \label{pggan_brown_curly}
\end{figure}

\textbf{Vector Arithmetic}

We investigate semantic relations encoded in the latent space of PgGAN generator by performing vector arithmetic on two attributes - smiling and having brown/curly hair. The results are shown in figure \ref{pggan_smile} and \ref{pggan_brown_curly} respectively.

We see here clearly that high quality samples are generated which conform to the expectations. Smiling woman minus neutral woman plus neutral man does indeed equal smiling man. Similarly, man with brown/curly hair minus man with black hair plus woman with black hair does equal woman with woman with brown/curly hair.

This further proves the generalized nature of the ability of GANs to learn meaningful representations and encode semantic relationships.

\section{Conclusion}

We have shown and proved in this work that GANs do indeed learn meaningful latent representaions and encode semantically relevant information. We first applied different investigation techniques on DCGAN, like linear interpolation/extrapolation and circular interpolation of latent vector, and found that a smooth sample transition is produced in each of the experiments. This proved that a meaningful latent space representation is indeed being learned rather than image memorization. Secondly, we tested DCGAN with different vector arithmetic operations and found they encode semantically relevant attributes like smiling, wearing glasses and having blonde hair.

Further, we applied the aforementioned investigation techniques on PgGAN. We got a similar smooth transition while investigating the learned manifold. Also, we found attributes like smiling and having brown/curly hair was being encoded in the latent space. We conclude that GANs not only meaningful representation but also these properties generalize across different GANs.

\bibliographystyle{plainnat}
\bibliography{neurips_2019}

\begin{thebibliography}{20}
\providecommand{\natexlab}[1]{#1}
\providecommand{\url}[1]{\texttt{#1}}
\expandafter\ifx\csname urlstyle\endcsname\relax
  \providecommand{\doi}[1]{doi: #1}\else
  \providecommand{\doi}{doi: \begingroup \urlstyle{rm}\Url}\fi

\bibitem[Alec~Radford(2016)]{dcgan}
Soumith~Chintala Alec~Radford, Luke~Metz.
\newblock Unsupervised representation learning with deep convolutional
  generative adversarial networks.
\newblock \emph{ICLR}, 2016.

\bibitem[Andrew~Brock(2018)]{biggan}
Karen~Simonyan Andrew~Brock, Jeff~Donahue.
\newblock Large scale gan training for high fidelity natural image synthesis.
\newblock \emph{arXiv:1809.11096}, 2018.

\bibitem[Ben~Athiwaratkun(2019)]{semi_supervised}
Pavel Izmailov Andrew Gordon~Wilson Ben~Athiwaratkun, Marc~Finzi.
\newblock There are many consistent explanations of unlabeled data: Why you
  shoudl average.
\newblock \emph{ICLR}, 2019.

\bibitem[Fisher~Yu and Xiao(2015)]{lsun}
Yinda Zhang Shuran Song Thomas~Funkhouser Fisher~Yu, Ari~Seff and Jianxiong
  Xiao.
\newblock Lsun: Construction of a large-scale image dataset using deep learning
  with humans in the loop.
\newblock \emph{arXiv:1506.03365}, 2015.

\bibitem[Frey and Dayan(1995)]{belief_net}
Hinton G.~E. Frey, B.~J. and P.~Dayan.
\newblock Does the wake-sleep algorithm learn good density estimators?
\newblock \emph{NIPS}, 1995.

\bibitem[He et~al.(2017)He, Gkioxari, Doll\'{a}r, and Girshick]{maskrcnn}
Kaiming He, Georgia Gkioxari, Piotr Doll\'{a}r, and Ross Girshick.
\newblock {Mask R-CNN}.
\newblock In \emph{Proceedings of the International Conference on Computer
  Vision ({ICCV})}, 2017.

\bibitem[Hinton and Teh(2006)]{boltzman}
Osindero~S. Hinton, G.~E. and Y.~Teh.
\newblock A fast learning algorithm for deep belief nets.
\newblock \emph{Neural Computation, 18, 1527–1554}, 2006.

\bibitem[Ian J.~Goodfellow(2014)]{gan}
Mehdi Mirza Bing Xu David Warde-Farley Sherjil Ozair Aaron Courville
  Yoshua~Bengio Ian J.~Goodfellow, Jean Pouget-Abadie.
\newblock Generative adversarial networks.
\newblock \emph{arXiv:1406.2661}, 2014.

\bibitem[Ioffe and Szegedy(2015)]{batchnorm}
Sergey Ioffe and Christian Szegedy.
\newblock Batch normalization: Accelerating deep network training by reducing
  internal covariate shift.
\newblock \emph{arXiv preprint arXiv:1502.03167}, 2015.

\bibitem[Judy~Hoffman(2017)]{cycada}
Taesung Park Jun-Yan Zhu Phillip Isola Kate Saenko Alexei A. Efros
  Trevor~Darrell Judy~Hoffman, Eric~Tzeng.
\newblock Cycada: Cycle-consistent adversarial domain adaptation.
\newblock \emph{arXiv:1711.03213}, 2017.

\bibitem[Kaiming~He(2015)]{resnet}
Shaoqing Ren Jian~Sun Kaiming~He, Xiangyu~Zhang.
\newblock Deep residual learning for image recognition.
\newblock \emph{arXiv:1512.03385}, 2015.

\bibitem[Kingma(2013)]{vae}
D.~P. Kingma.
\newblock Fast gradient-based inference with continuous latent variable models
  in auxiliary form.
\newblock \emph{Technical report, arxiv:1306.0733}, 2013.

\bibitem[Krizhevsky et~al.(2012)Krizhevsky, Sutskever, and Hinton]{alexnet}
Alex Krizhevsky, Ilya Sutskever, and Geoffrey~E. Hinton.
\newblock Imagenet classification with deep convolutional neural networks.
\newblock \emph{NIPS'12}, 2012.

\bibitem[Liu et~al.(2015)Liu, Luo, Wang, and Tang]{celeba}
Ziwei Liu, Ping Luo, Xiaogang Wang, and Xiaoou Tang.
\newblock Deep learning face attributes in the wild.
\newblock In \emph{Proceedings of International Conference on Computer Vision
  (ICCV)}, December 2015.

\bibitem[Maas and Ng(2013)]{leaky_relu}
Hannun Awni~Y Maas, Andrew~L and Andrew~Y. Ng.
\newblock Rectifier nonlinearities improve neural network acoustic models.
\newblock \emph{Proc. ICML, volume 30}, 2013.

\bibitem[Nair and Hinton(2010)]{relu}
Vinod Nair and Geoffrey~E Hinton.
\newblock Rectified linear units improve restricted boltzmann machines.
\newblock \emph{Proceedings of the 27th International Conference on Machine
  Learning (ICML-10)}, 2010.

\bibitem[Oord and Kavukcuoglu(2016)]{wavenet}
Dieleman S. Zen H. Simonyan K. Vinyals O. Graves A. Kalchbrenner N. Senior~A.
  Oord, A. v.~d. and K.~Kavukcuoglu.
\newblock Wavenet: A generative model for raw audio.
\newblock \emph{arXiv preprint arXiv:1609.03499}, 2016.

\bibitem[Rezende and Wierstra(2014)]{vae2}
Mohamed~S. Rezende, D.~J. and D.~Wierstra.
\newblock Stochastic backpropagation and approximate inference in deep
  generative models.
\newblock \emph{ICML}, 2014.

\bibitem[Tero~Karras(2018)]{pggan}
Samuli Laine Jaakko~Lehtinen Tero~Karras, Timo~Aila.
\newblock Progressive growing of gans for improved quality, stability, and
  variation.
\newblock \emph{ICLR}, 2018.

\bibitem[Yunjey~Choi(2018)]{stargan}
Munyoung Kim Jung-Woo Ha Sunghun Kim Jaegul~Choo Yunjey~Choi, Minje~Choi.
\newblock Stargan: Unified generative adversarial networks for multi-domain
  image-to-image translation.
\newblock \emph{CVPR}, 2018.

\end{thebibliography}

\section{Supplementary Material}

\subsection{Generated samples from PgGAN}

\begin{figure}[H]
    \includegraphics[width=0.2\textwidth, height=110pt]{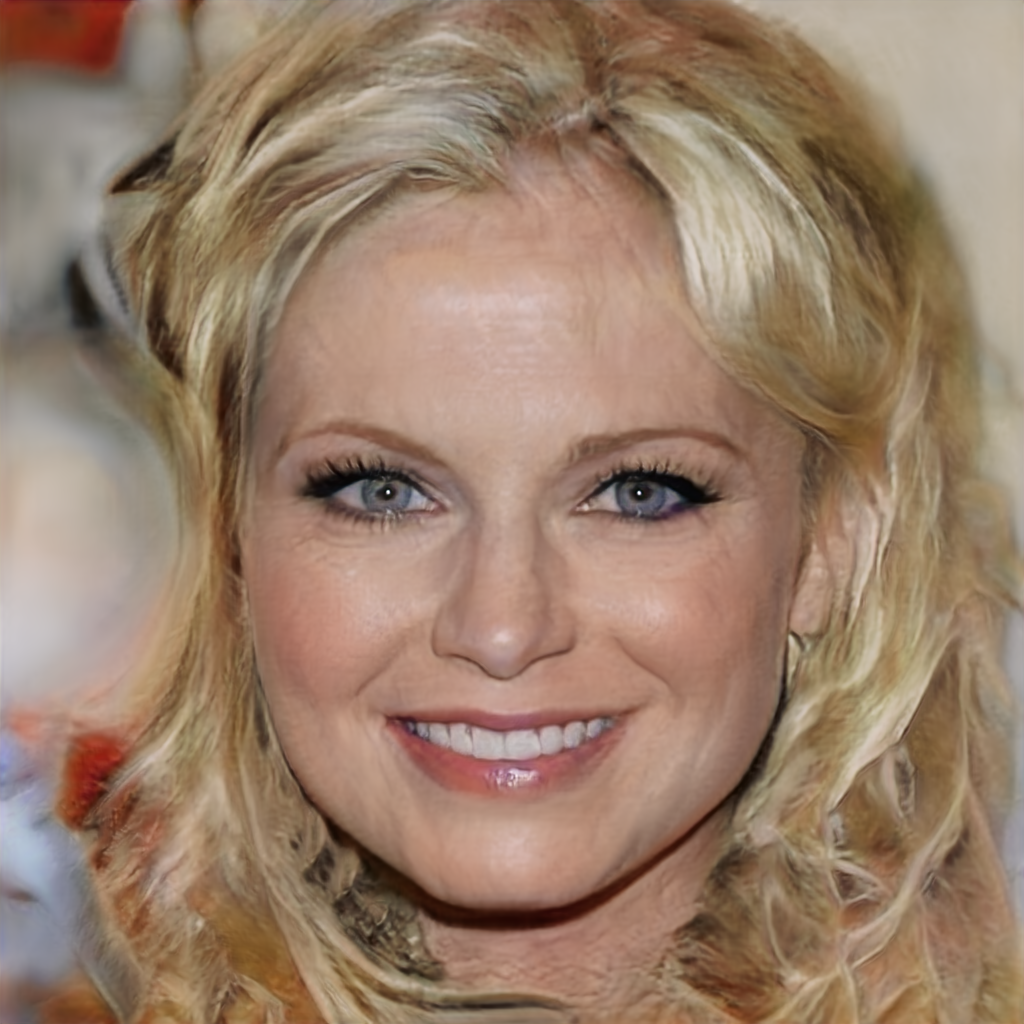}\hfill
    \includegraphics[width=0.2\textwidth, height=110pt]{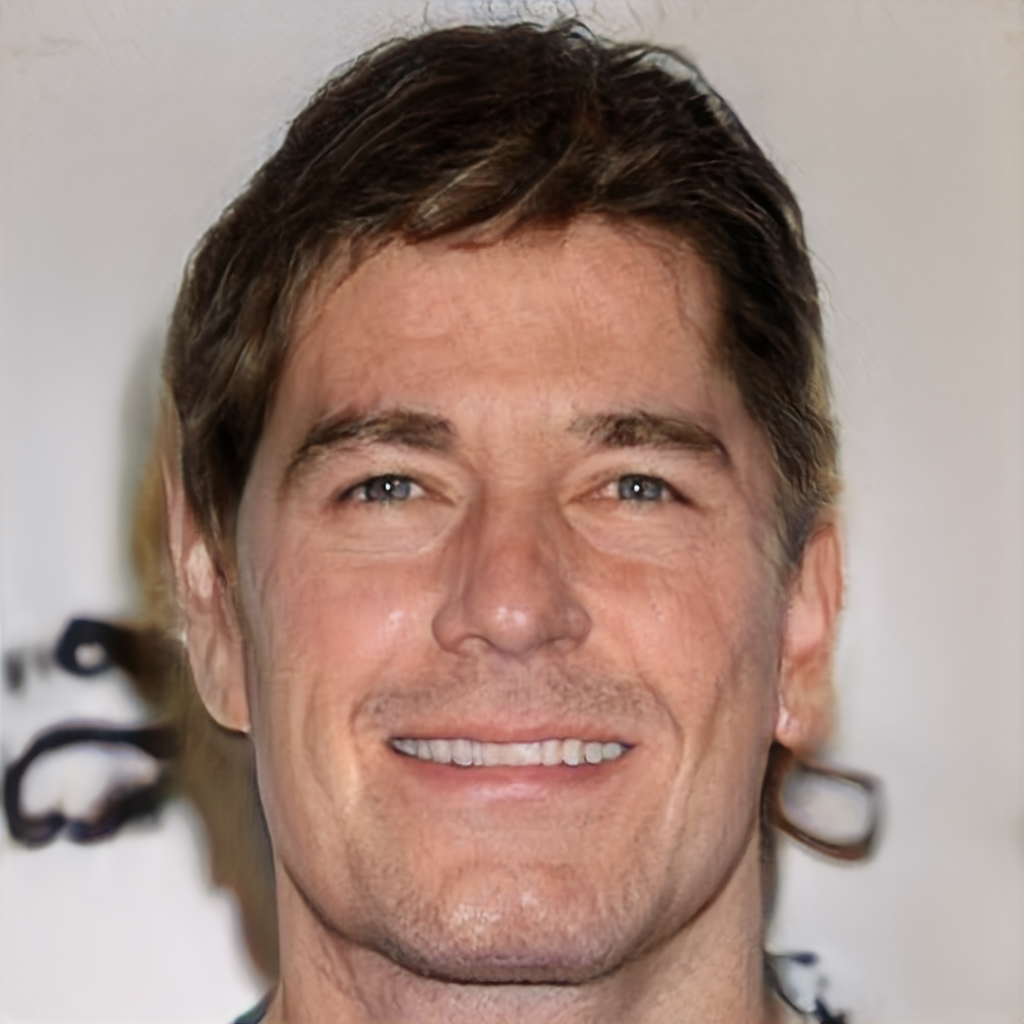}\hfill
    \includegraphics[width=0.2\textwidth, height=110pt]{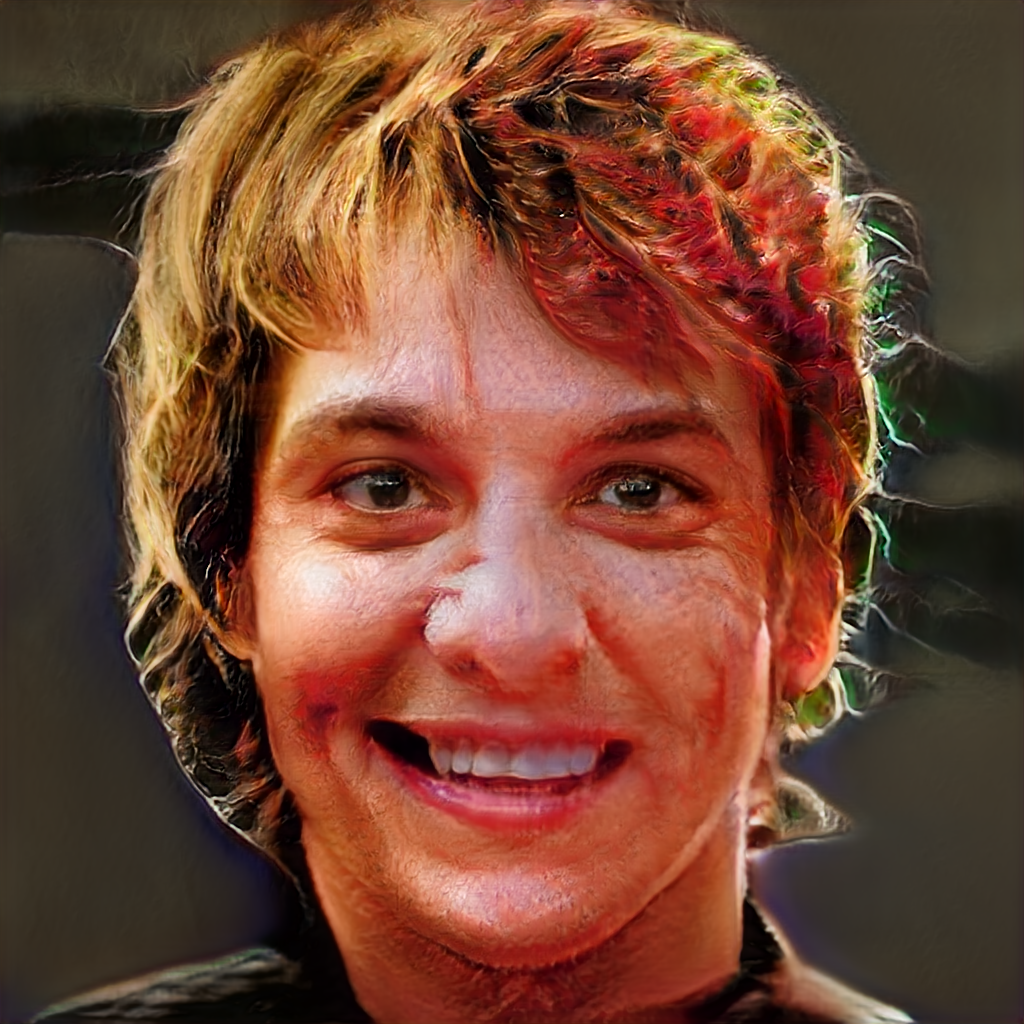}\hfill
    \includegraphics[width=0.2\textwidth, height=110pt]{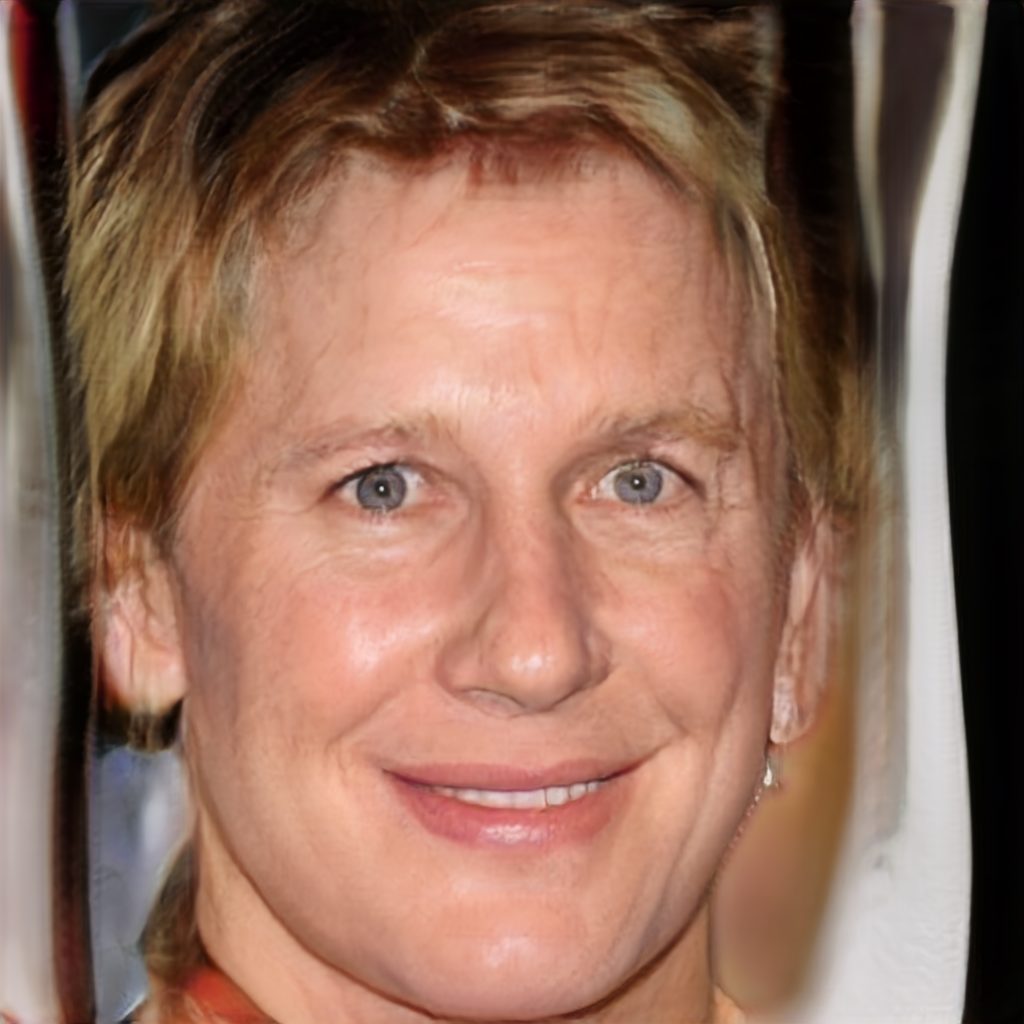}\hfill
    \includegraphics[width=0.2\textwidth, height=110pt]{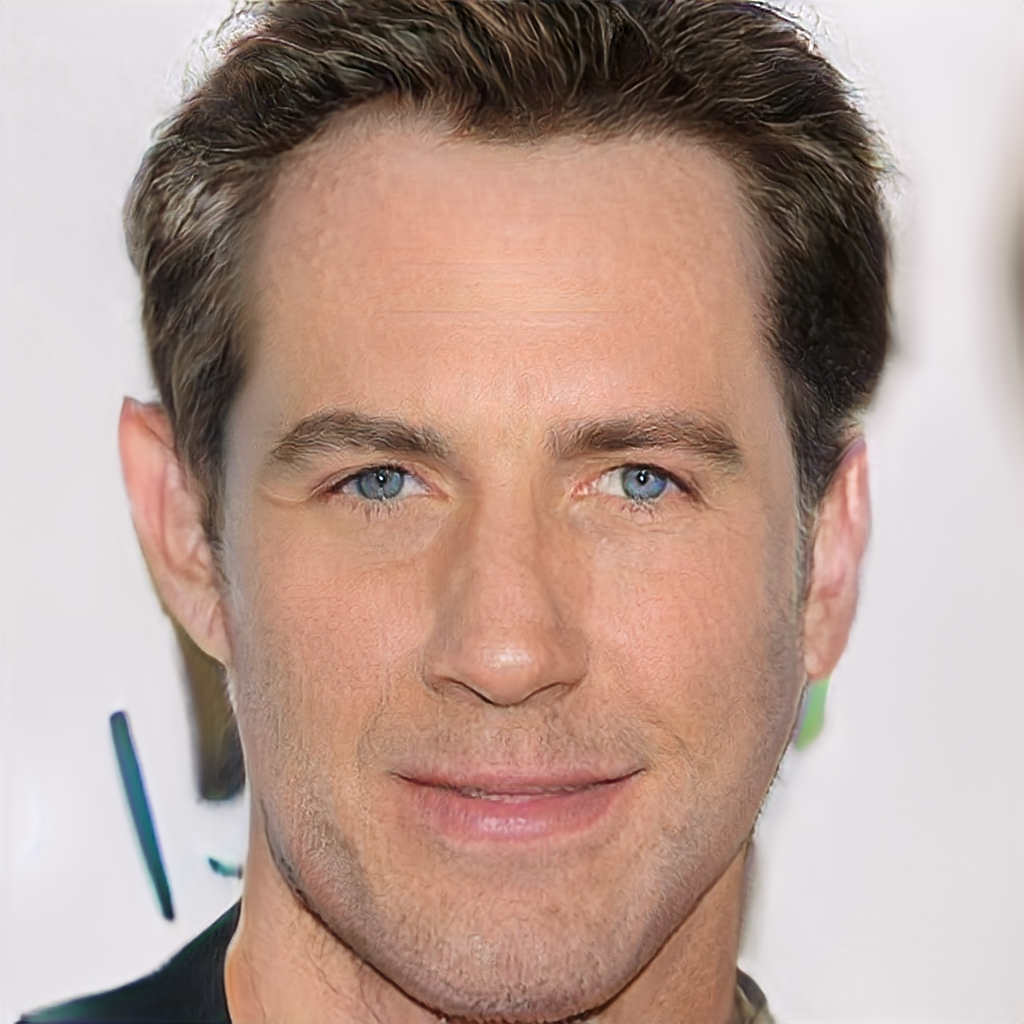}\hfill
    \includegraphics[width=0.2\textwidth, height=110pt]{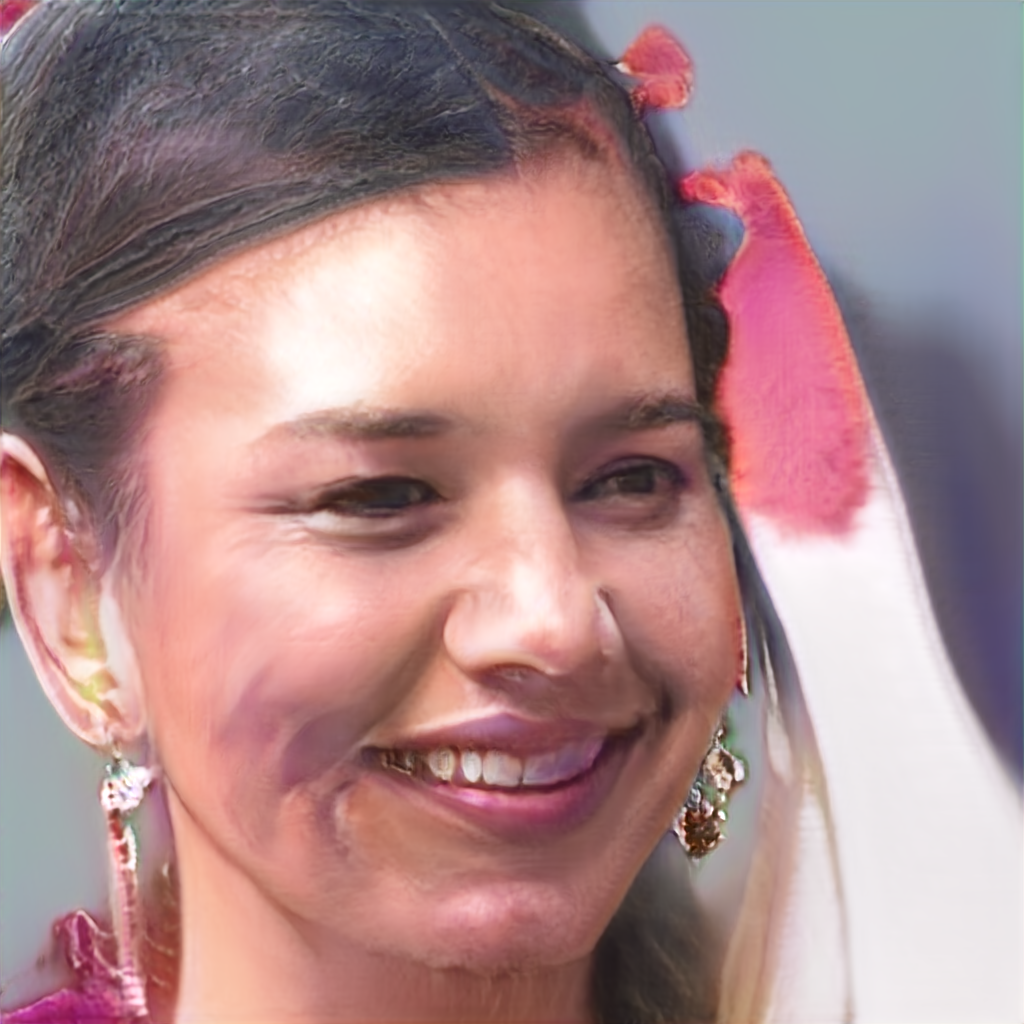}\hfill
    \includegraphics[width=0.2\textwidth, height=110pt]{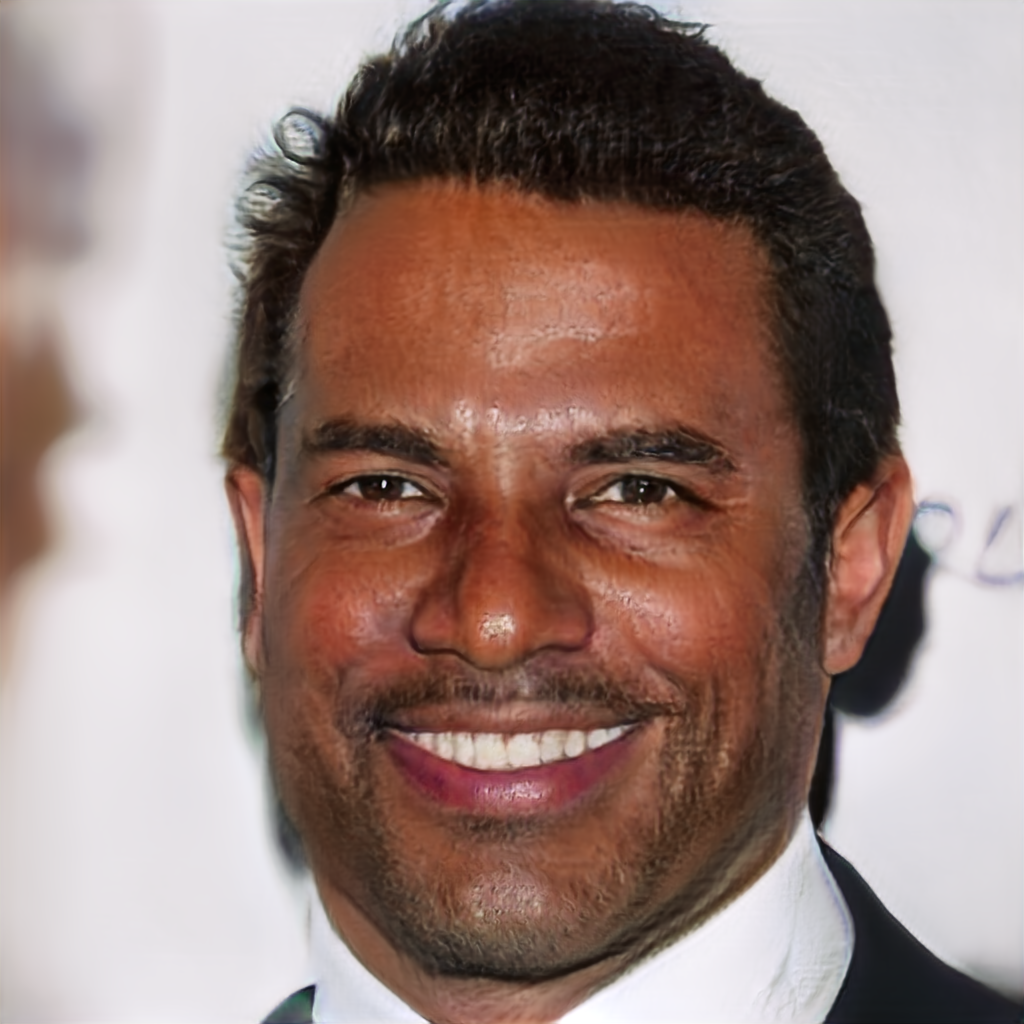}\hfill
    \includegraphics[width=0.2\textwidth, height=110pt]{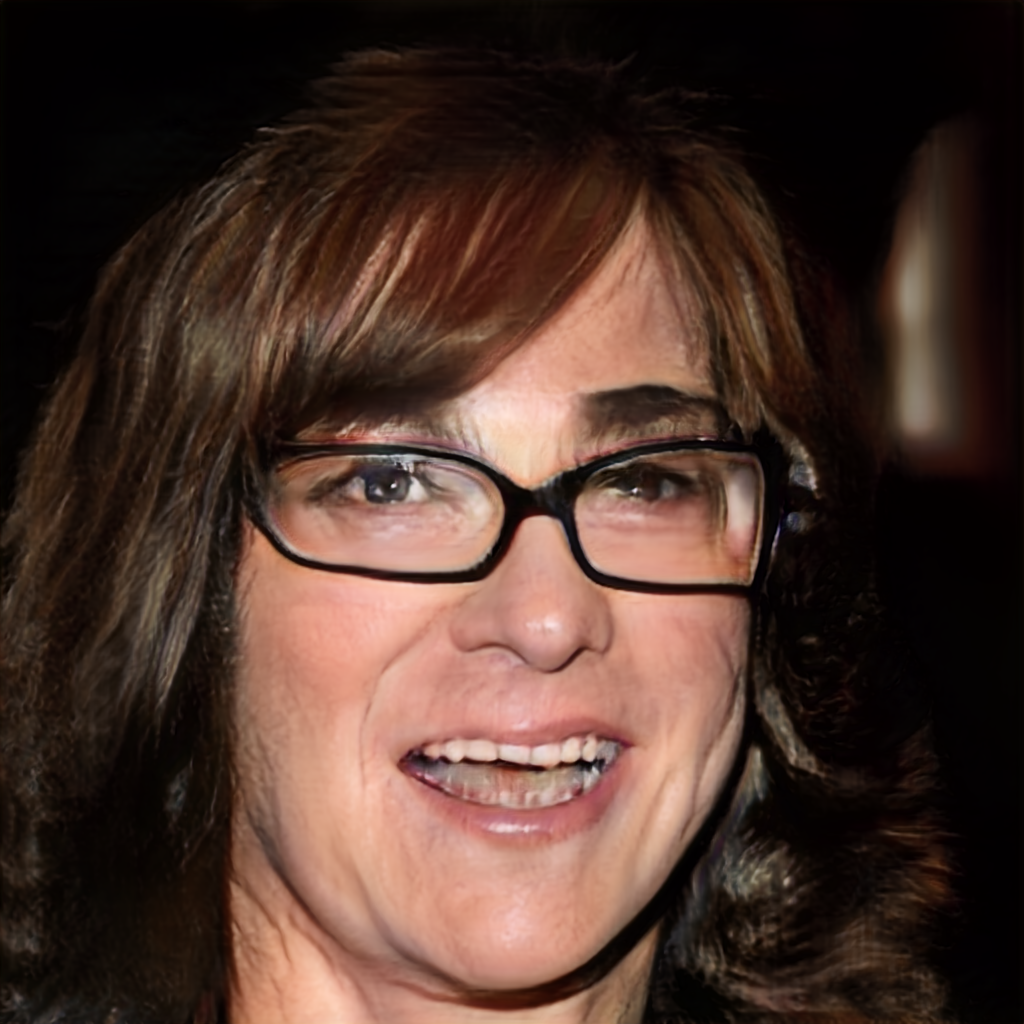}\hfill
    \includegraphics[width=0.2\textwidth, height=110pt]{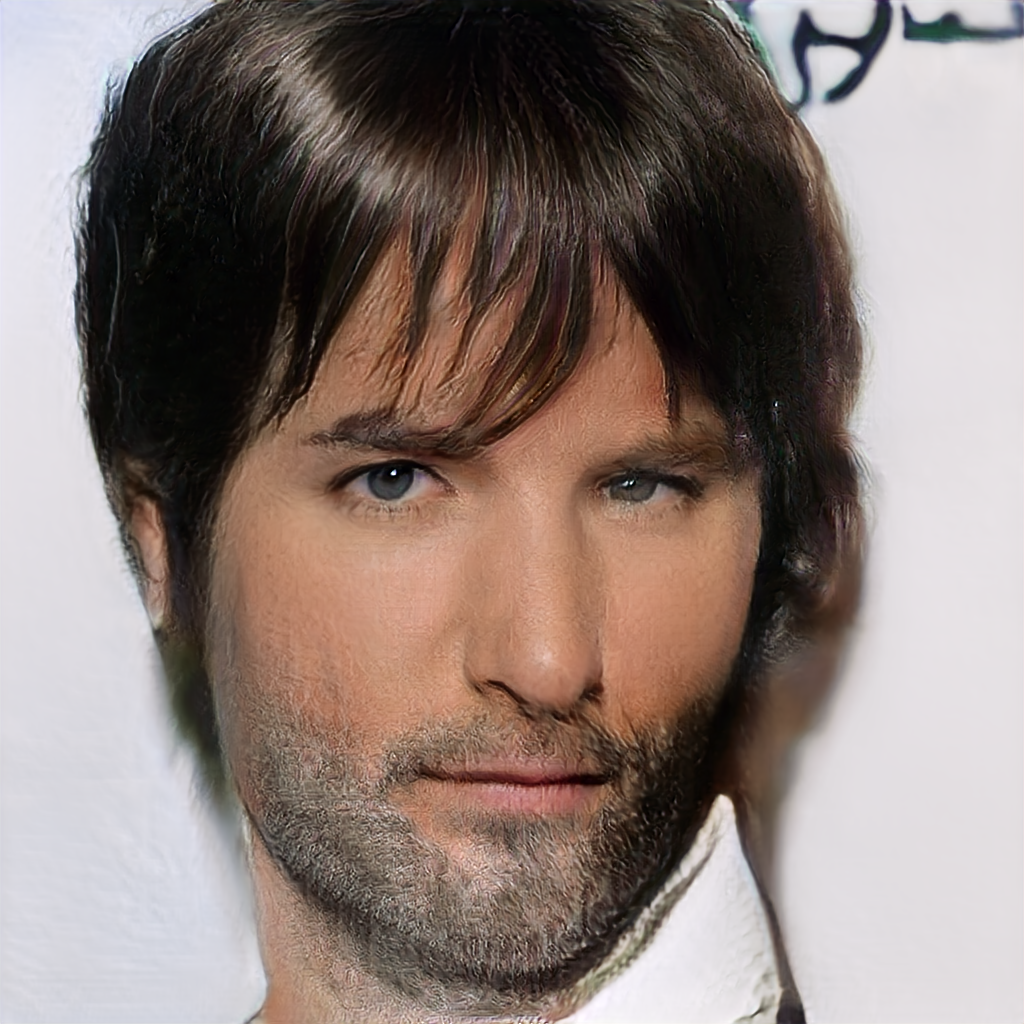}\hfill
    \includegraphics[width=0.2\textwidth, height=110pt]{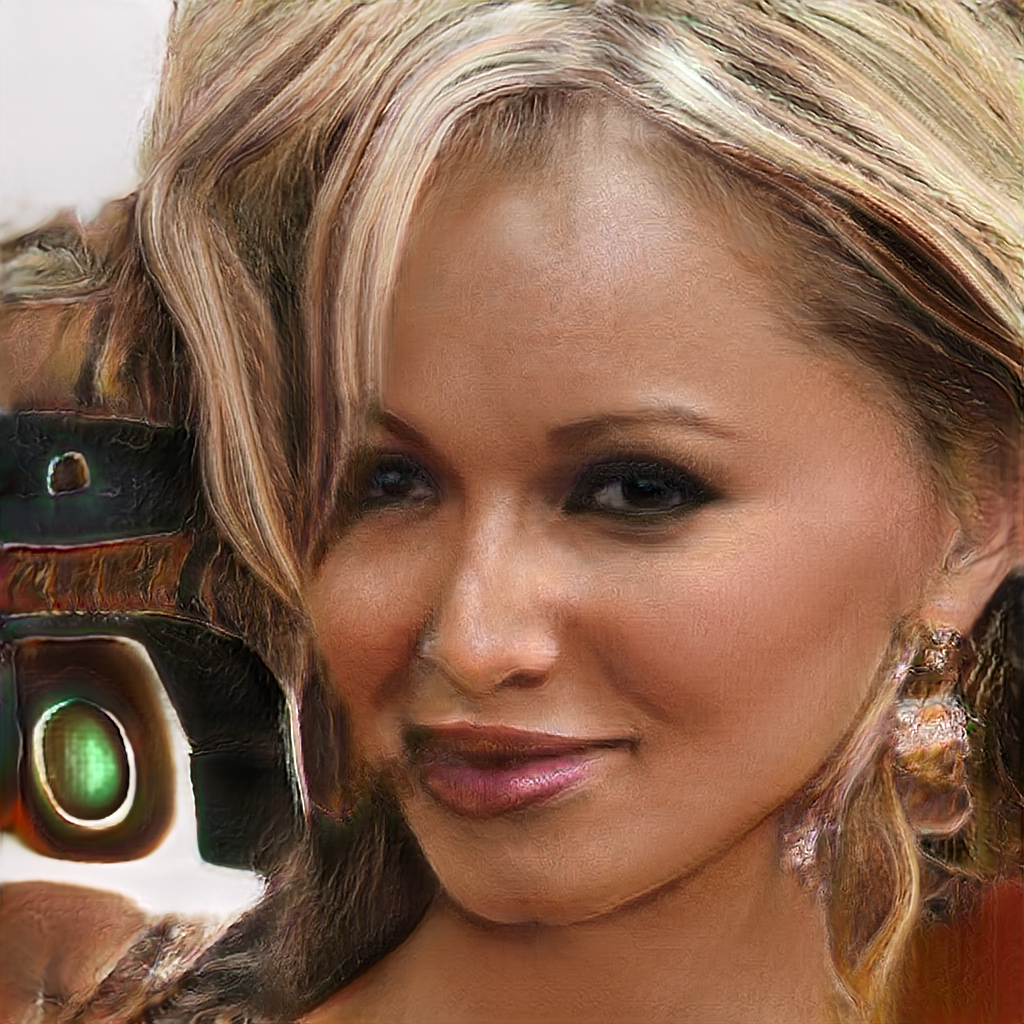}\hfill
    \caption{Samples generated by PgGAN}
    \label{pggan_samples}
\end{figure}

\subsection{Linear Interpolation by PyGAN}

\begin{figure}[H]
    \includegraphics[width=0.2\textwidth, height=110pt]{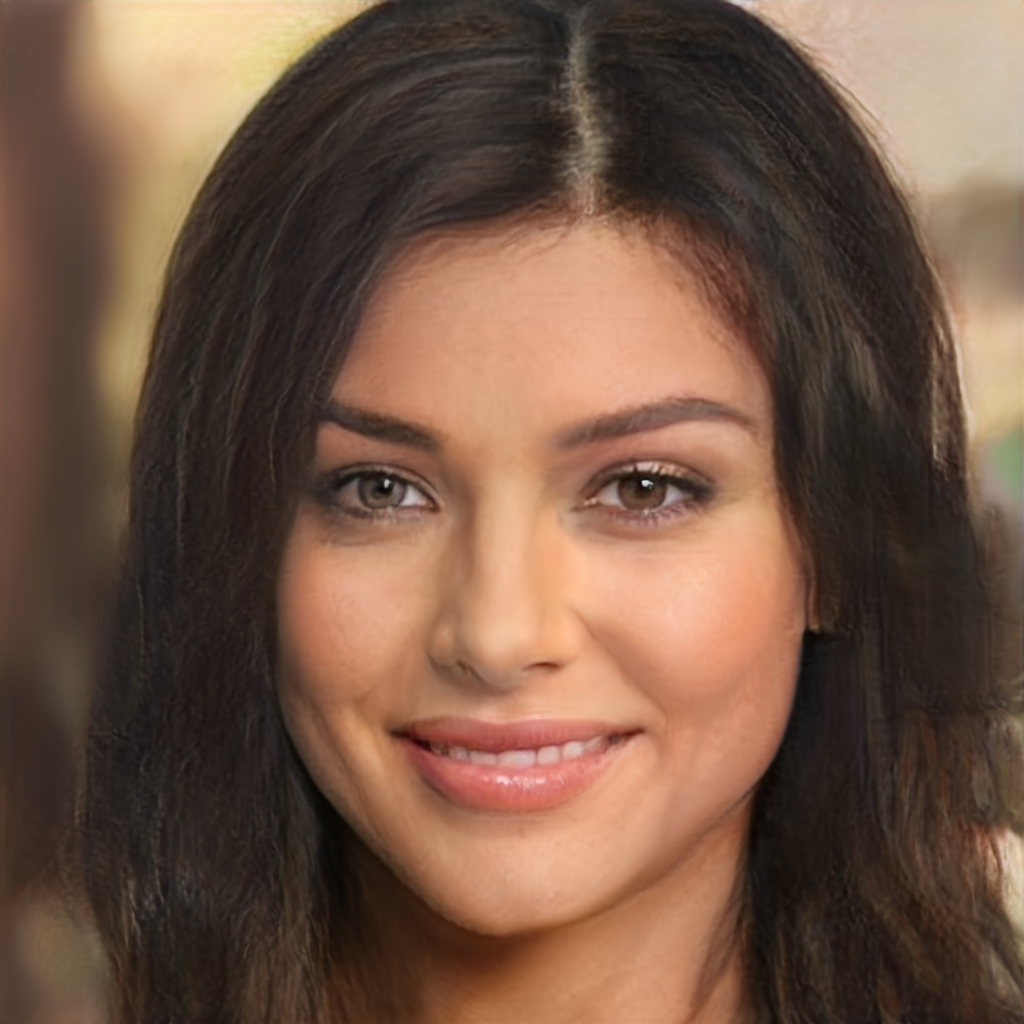}\hfill
    \includegraphics[width=0.2\textwidth, height=110pt]{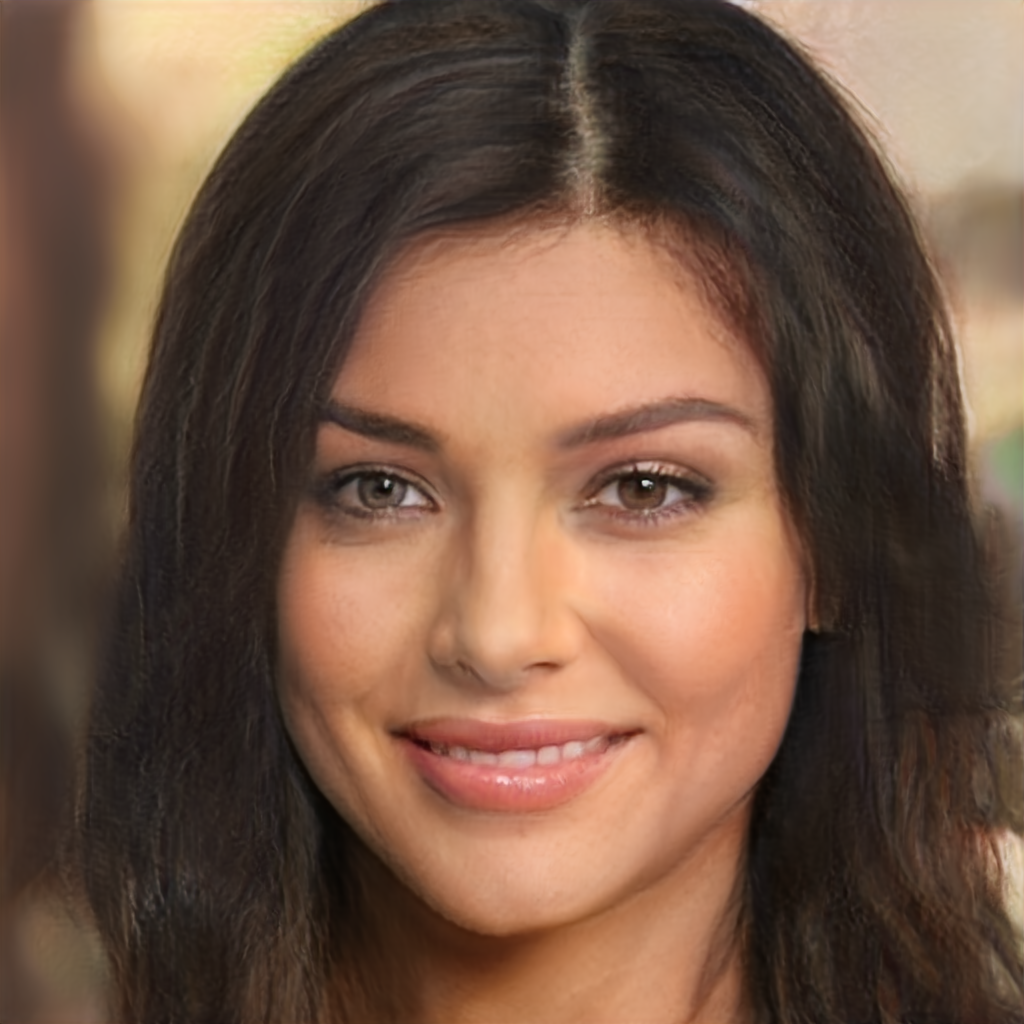}\hfill
    \includegraphics[width=0.2\textwidth, height=110pt]{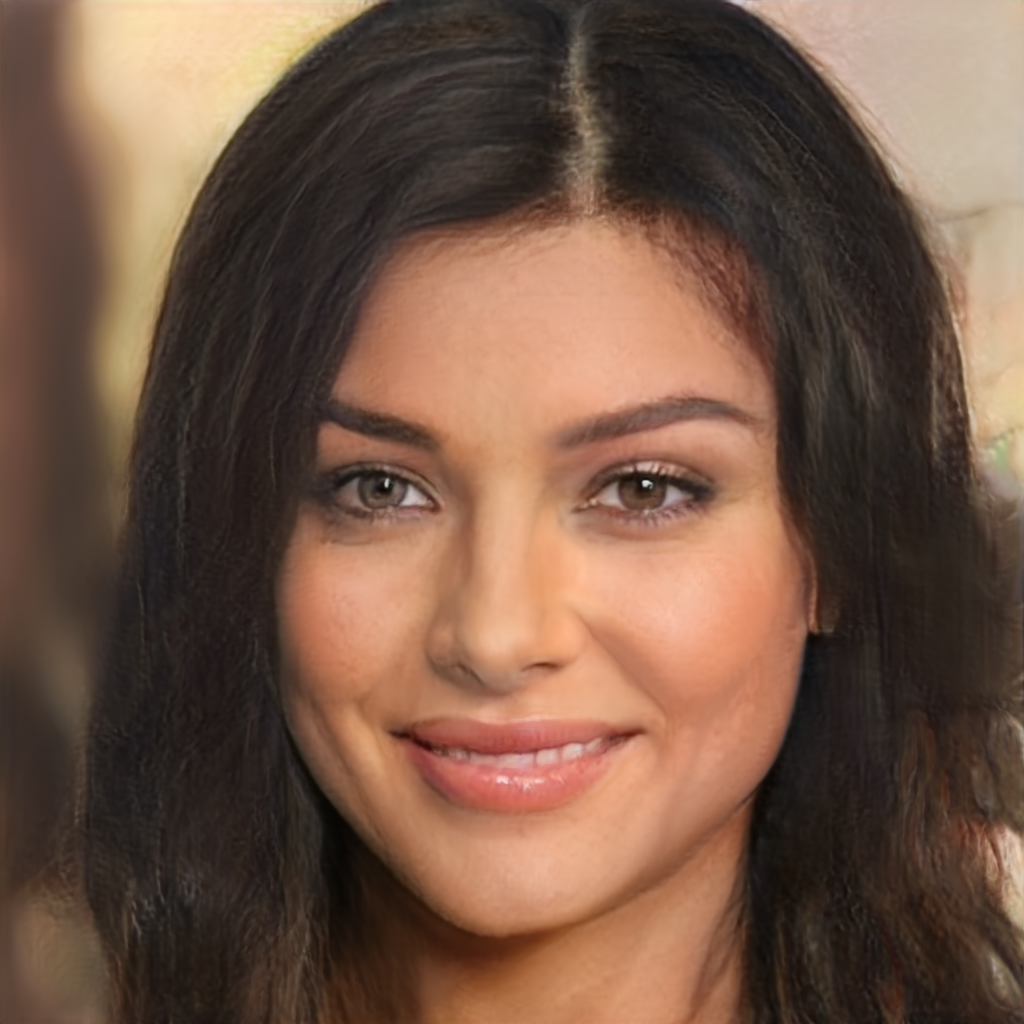}\hfill
    \includegraphics[width=0.2\textwidth, height=110pt]{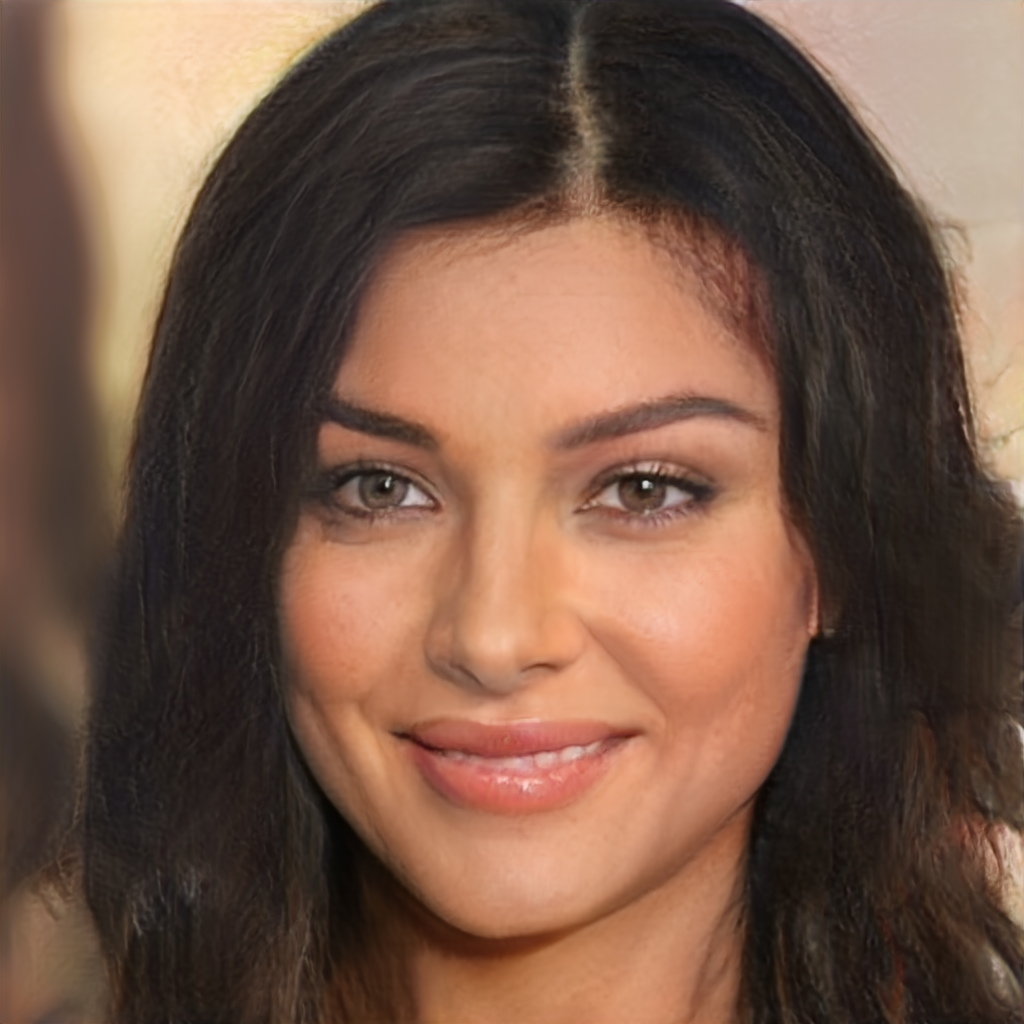}\hfill
    \includegraphics[width=0.2\textwidth, height=110pt]{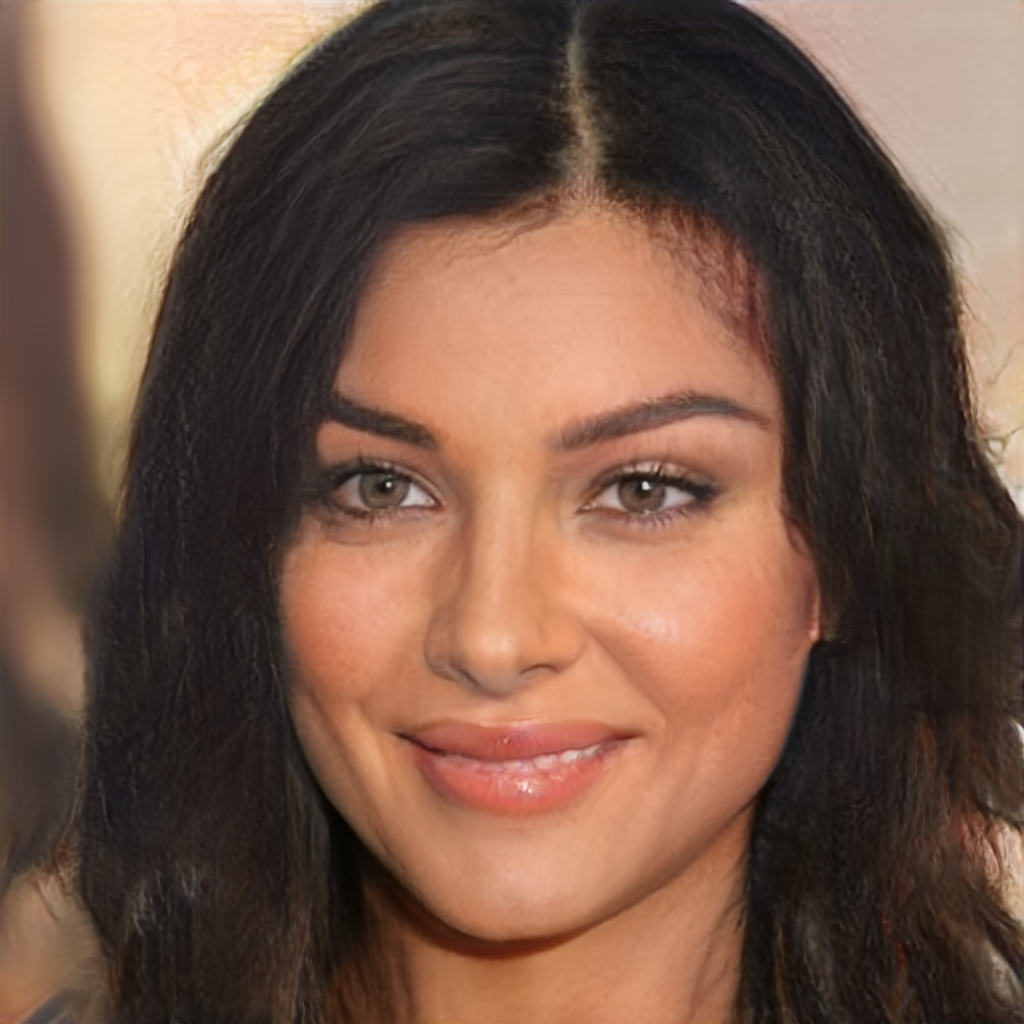}\hfill
    \includegraphics[width=0.2\textwidth, height=110pt]{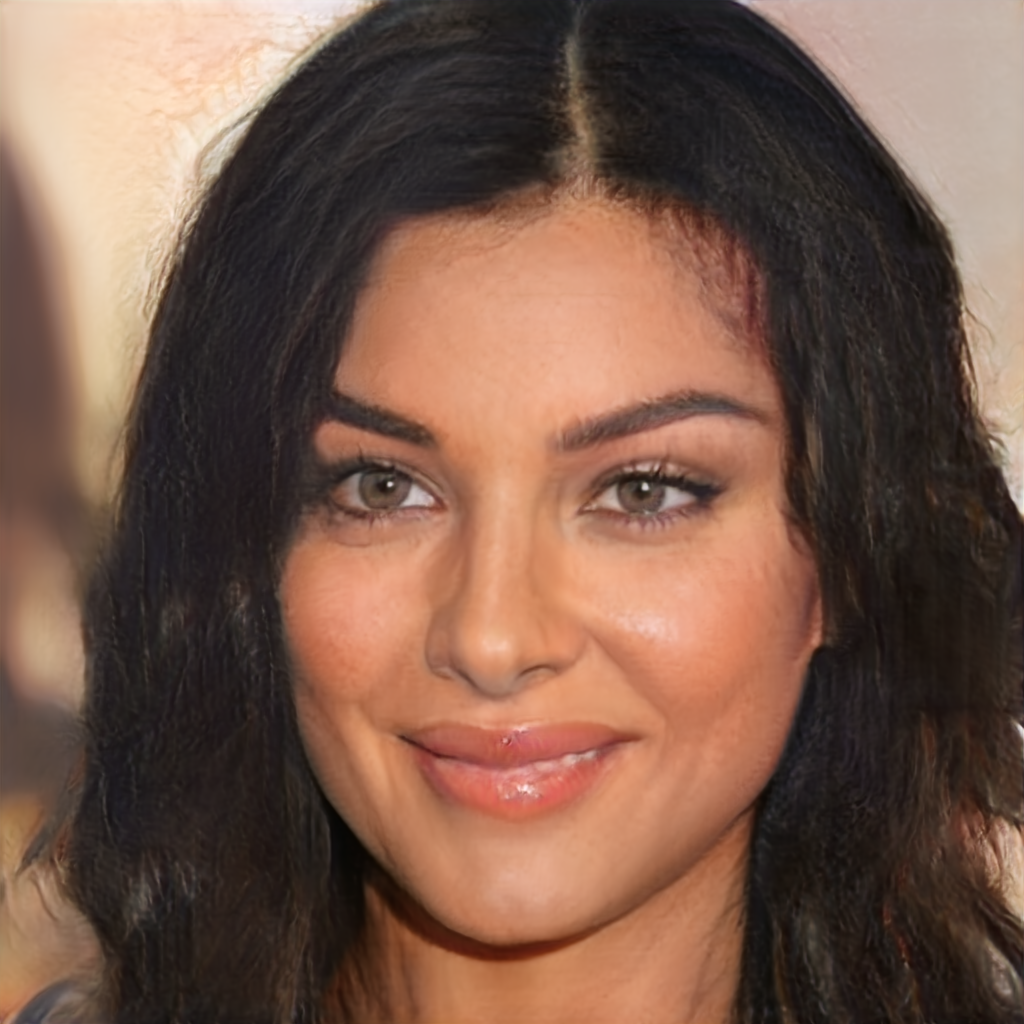}\hfill
    \includegraphics[width=0.2\textwidth, height=110pt]{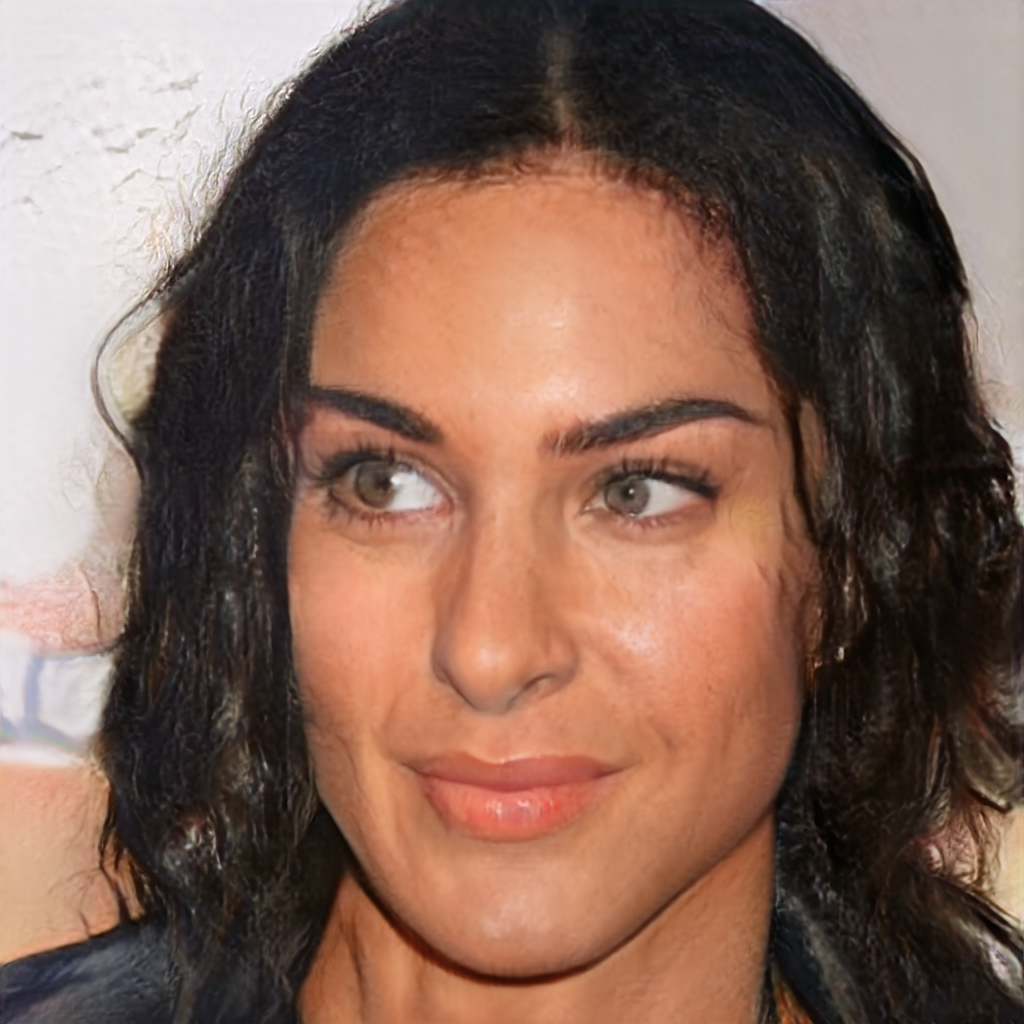}\hfill
    \includegraphics[width=0.2\textwidth, height=110pt]{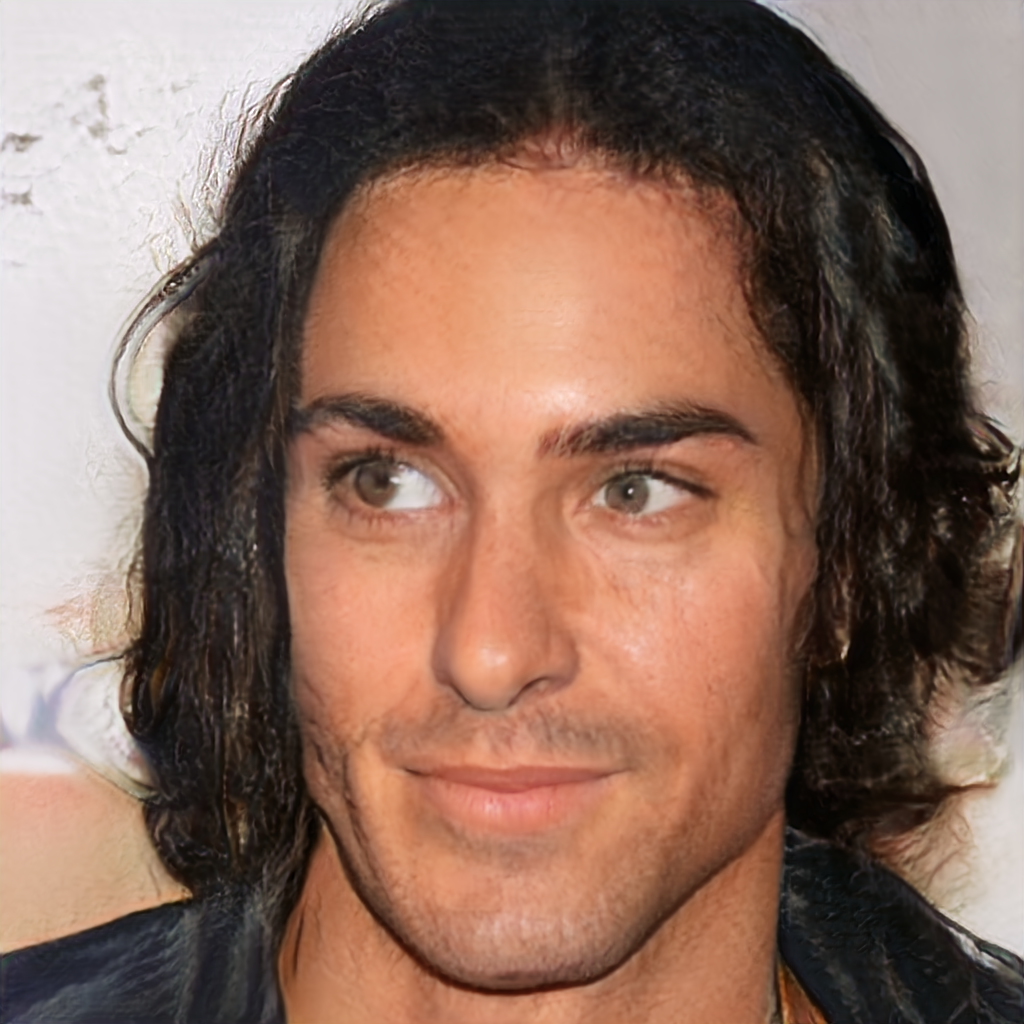}\hfill
    \includegraphics[width=0.2\textwidth, height=110pt]{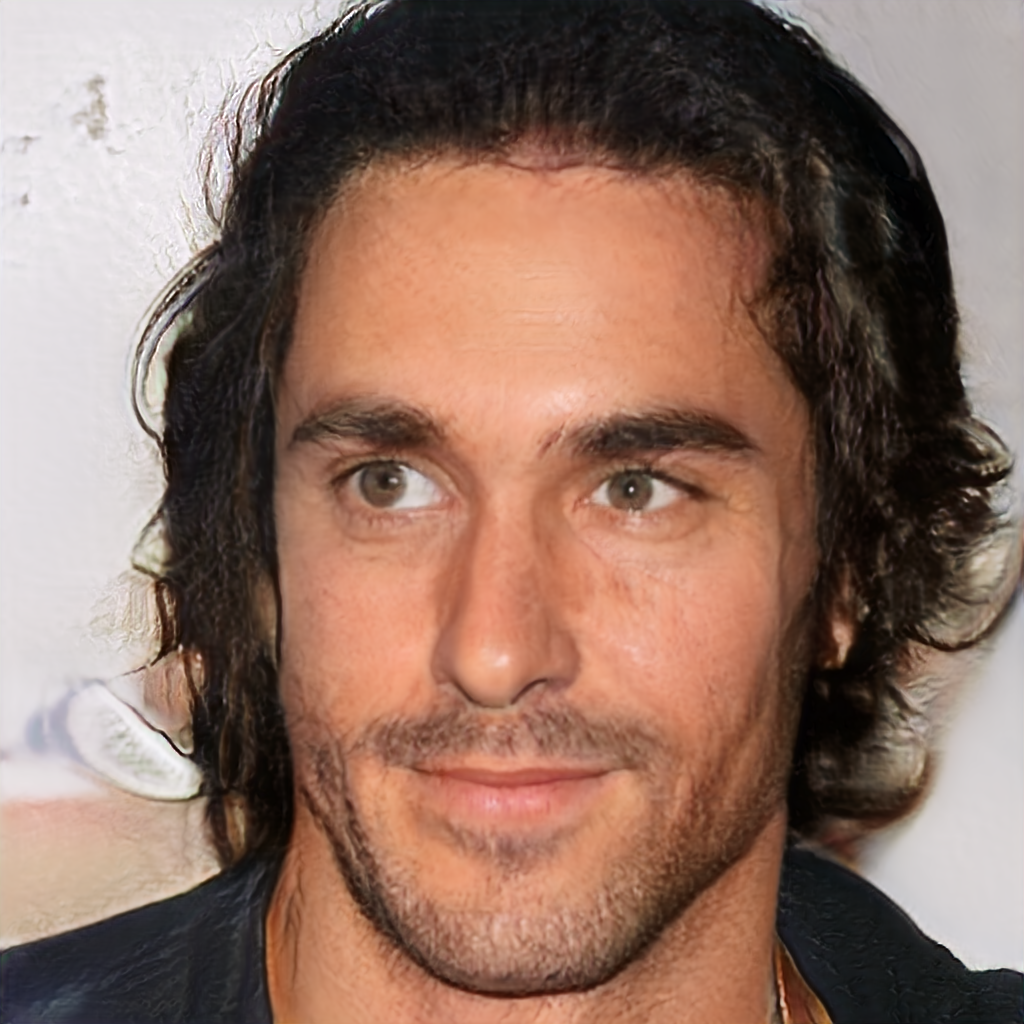}\hfill
    \includegraphics[width=0.2\textwidth, height=110pt]{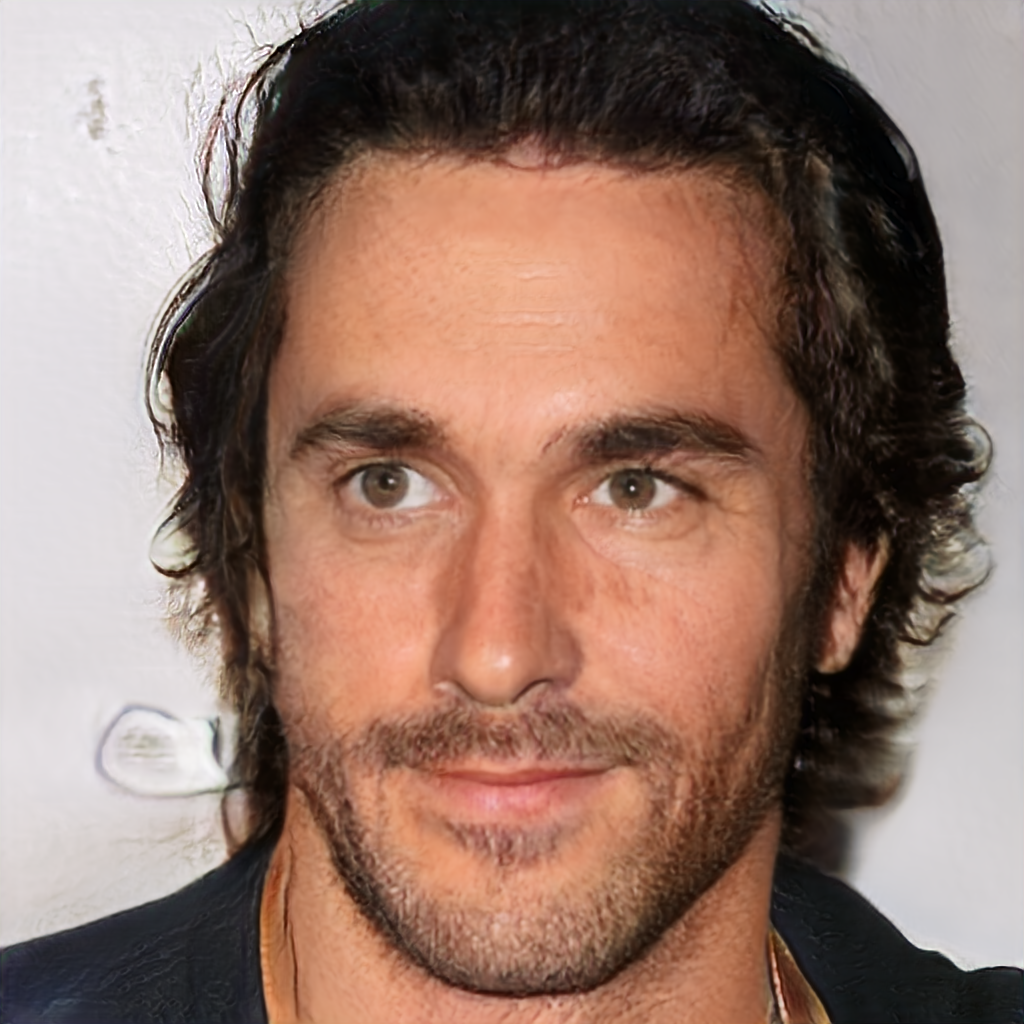}\hfill
    \caption{Interpolation from top left to bottom right generated by PgGAN.}
    \label{pggan_interpolate1}
\end{figure}

\begin{figure}[H]
    \includegraphics[width=0.2\textwidth, height=110pt]{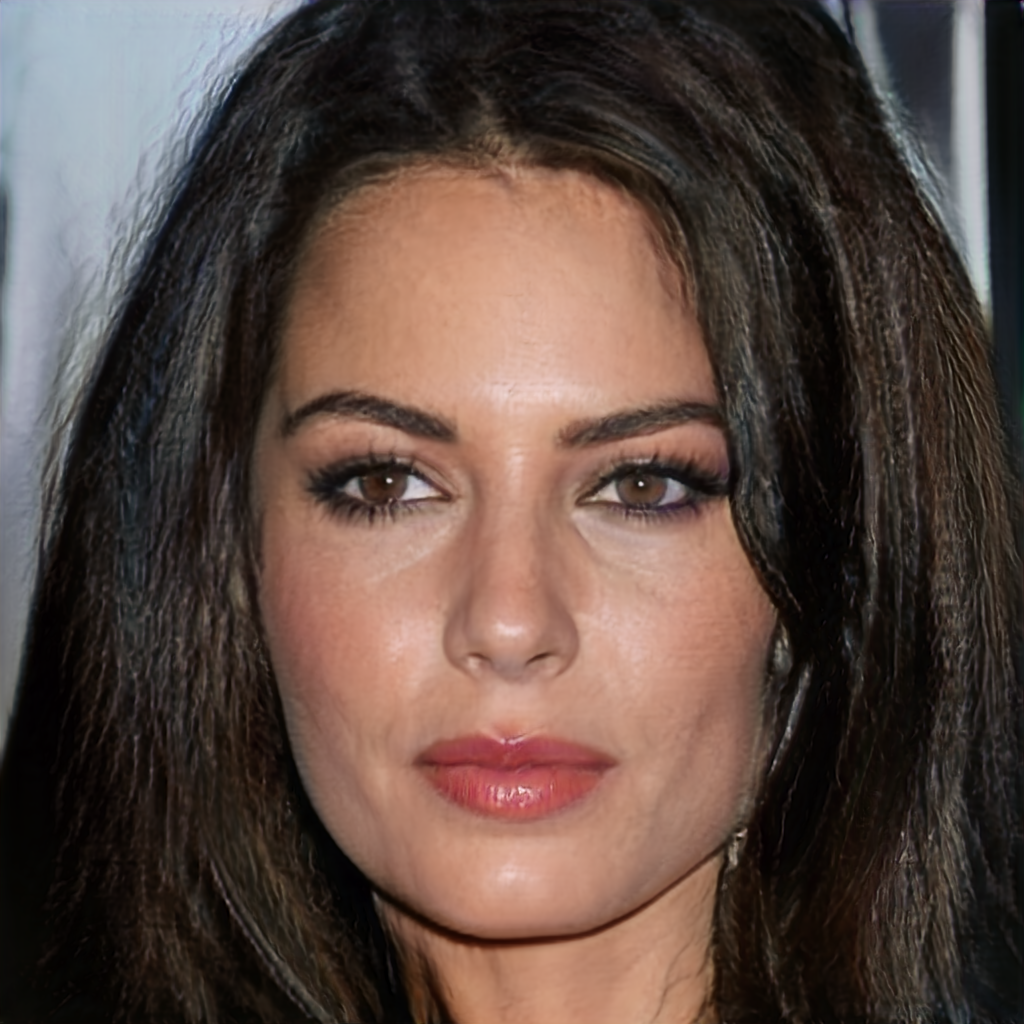}\hfill
    \includegraphics[width=0.2\textwidth, height=110pt]{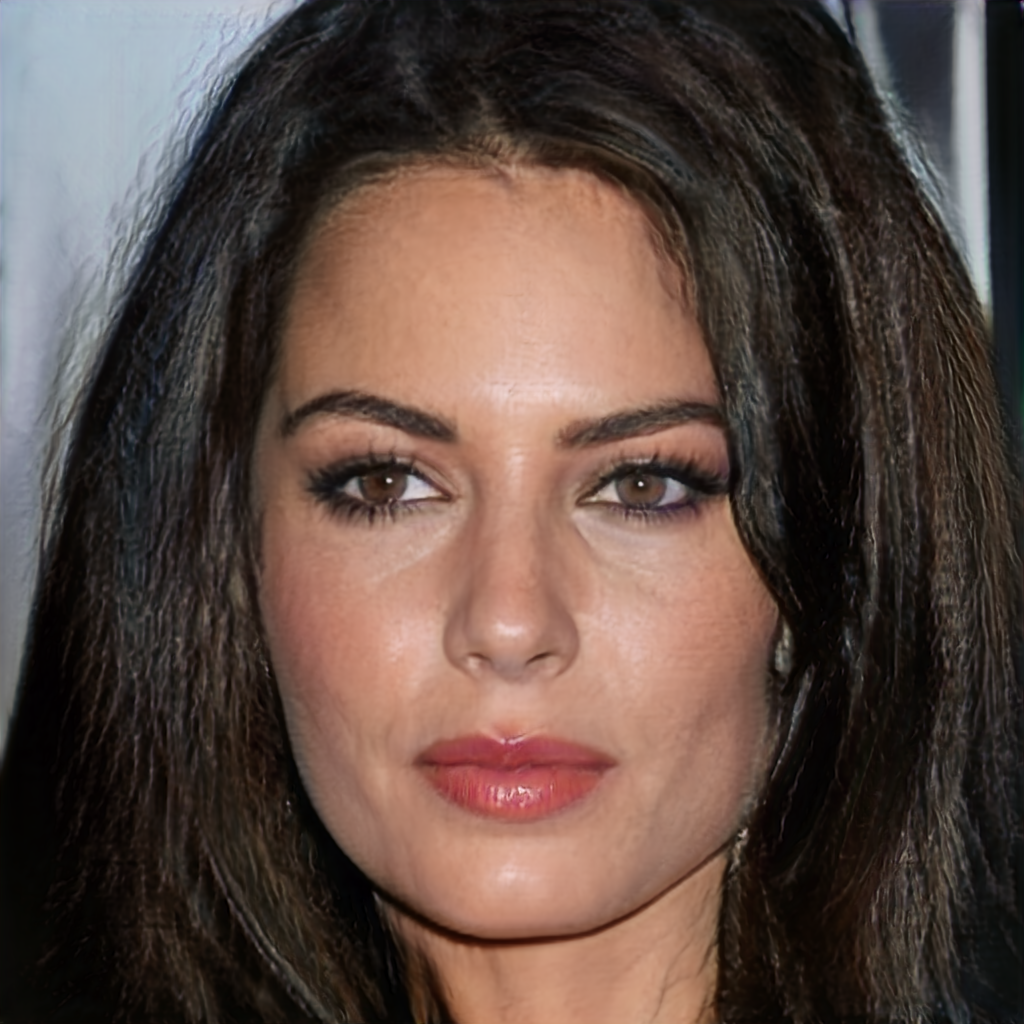}\hfill
    \includegraphics[width=0.2\textwidth, height=110pt]{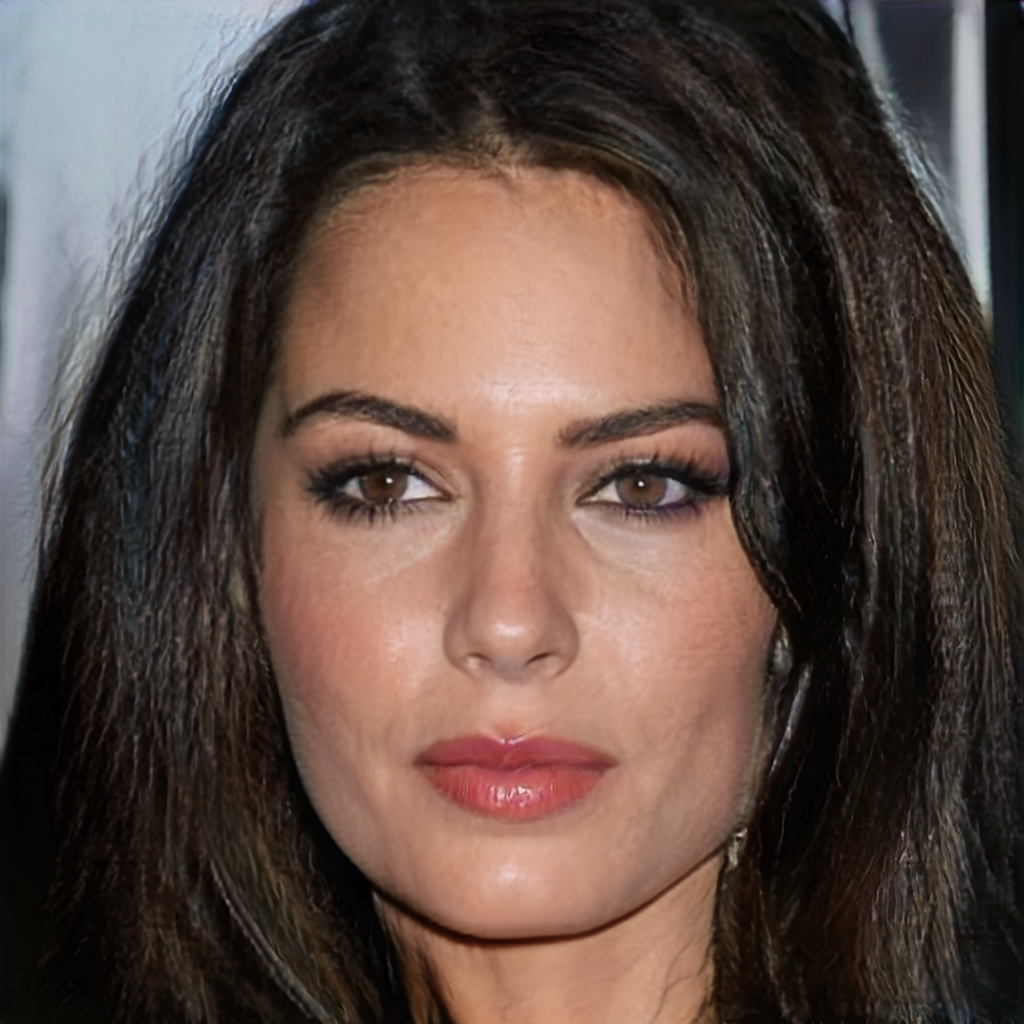}\hfill
    \includegraphics[width=0.2\textwidth, height=110pt]{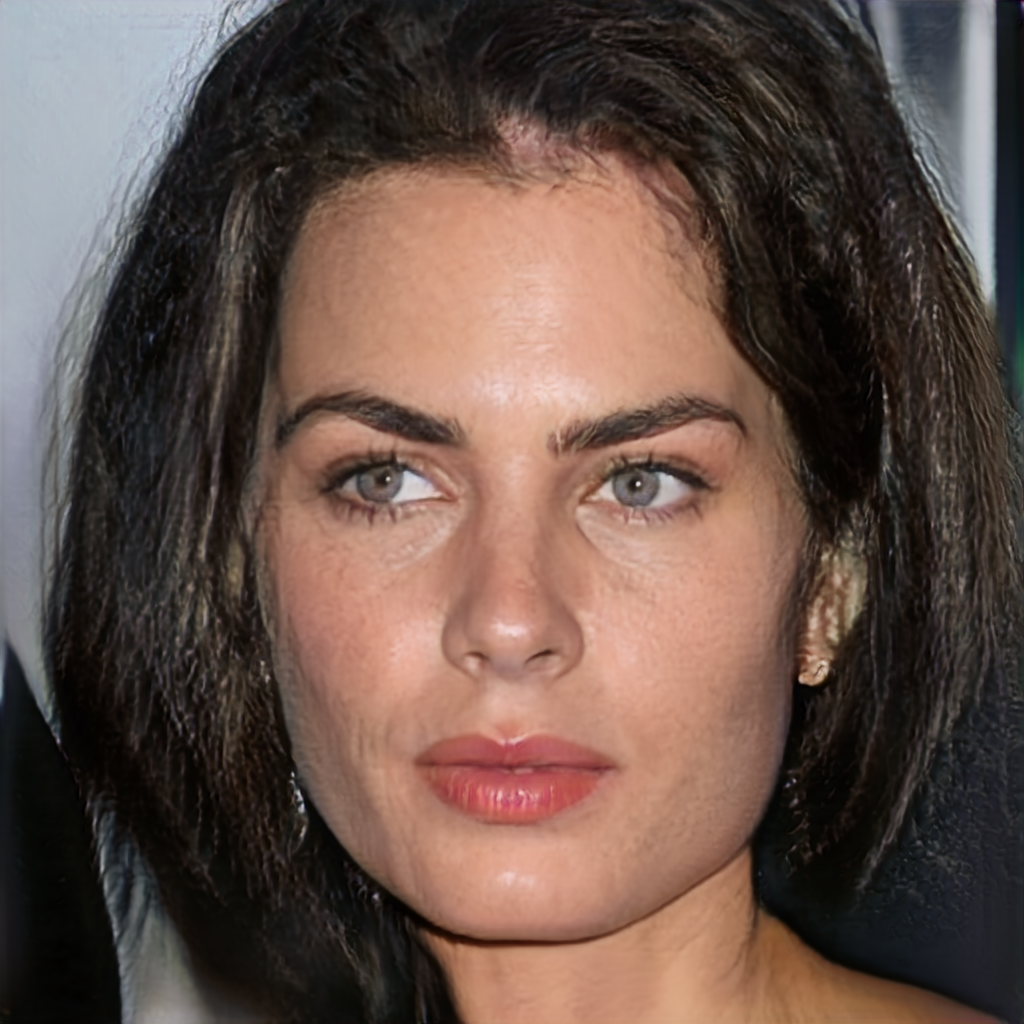}\hfill
    \includegraphics[width=0.2\textwidth, height=110pt]{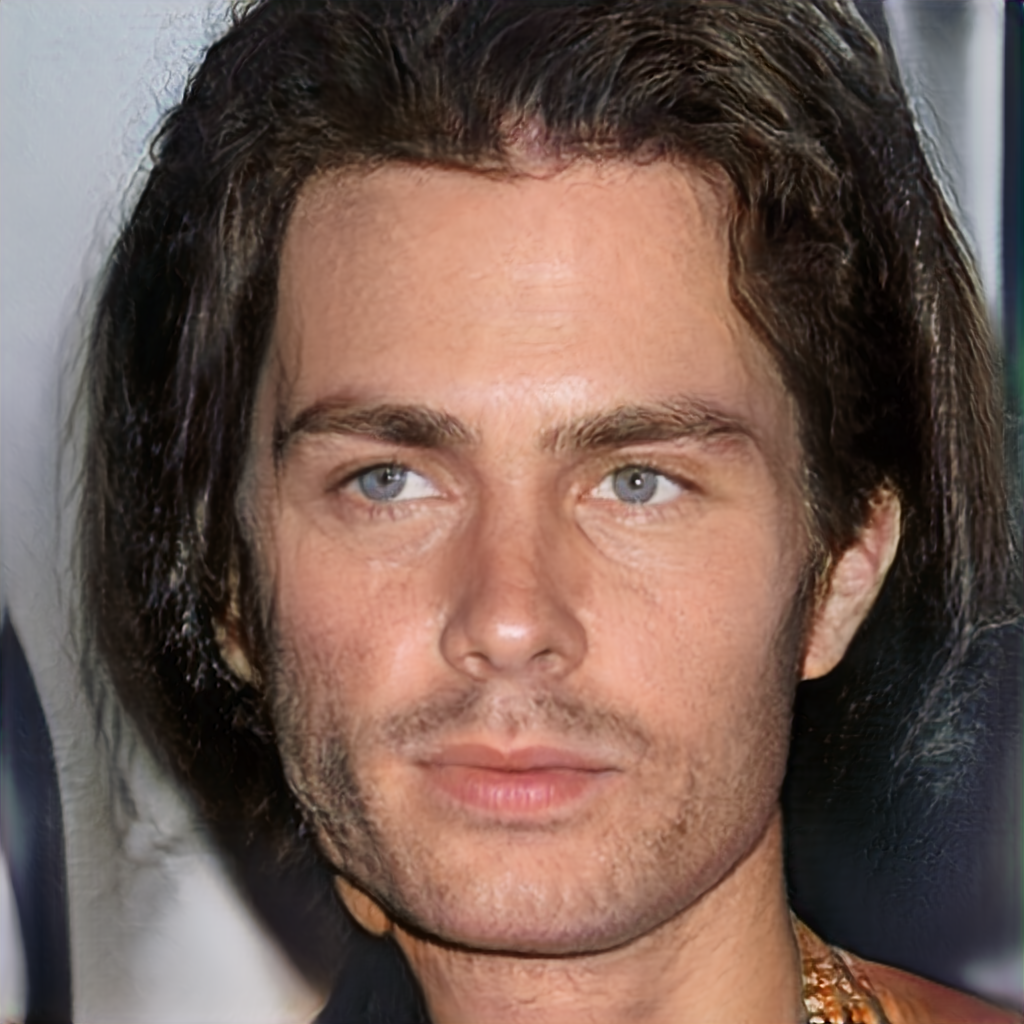}\hfill
    \includegraphics[width=0.2\textwidth, height=110pt]{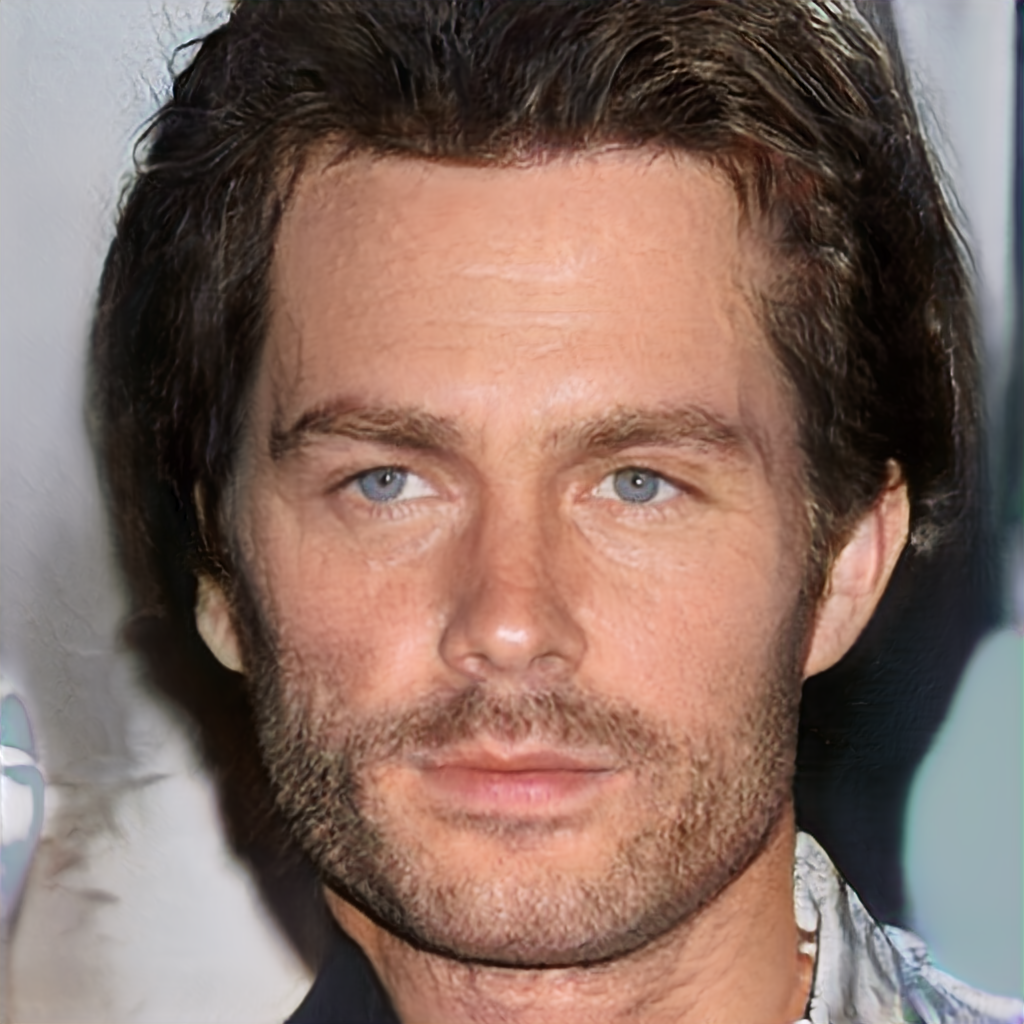}\hfill
    \includegraphics[width=0.2\textwidth, height=110pt]{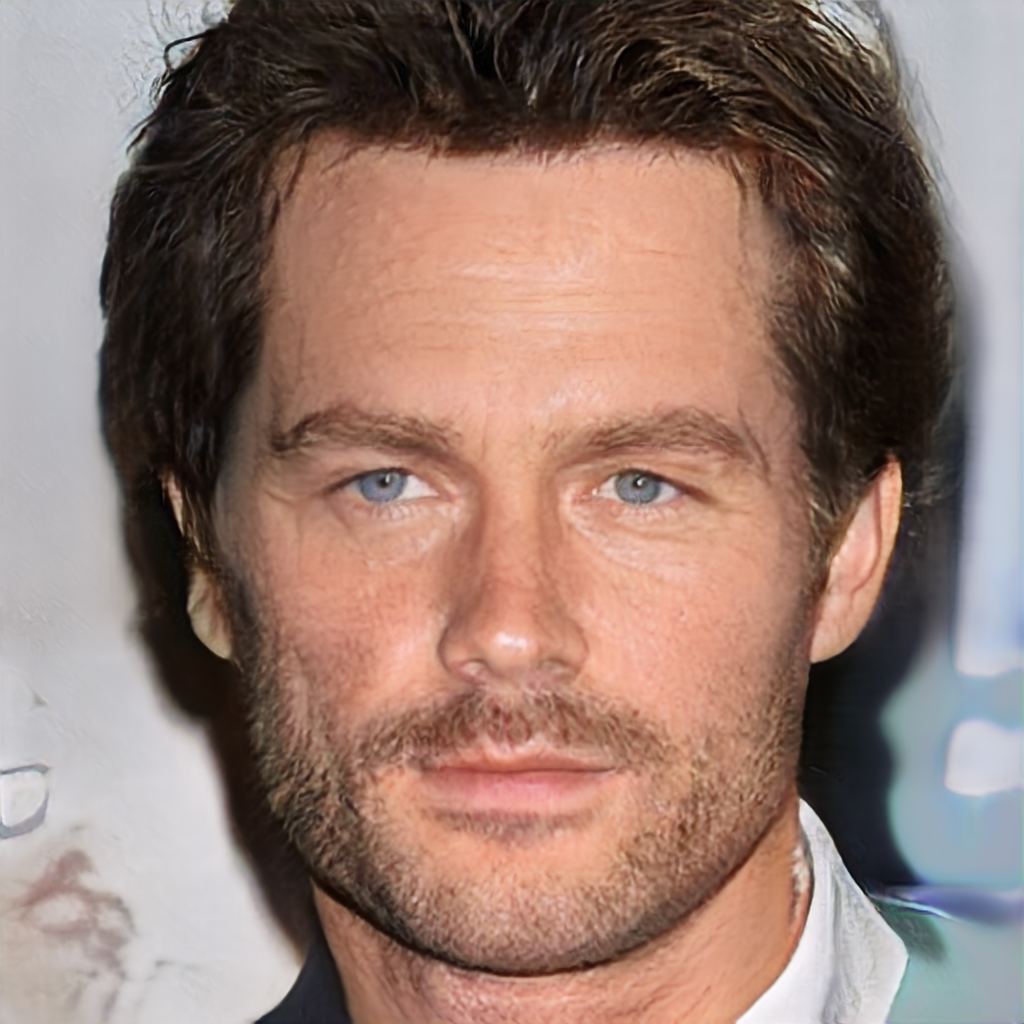}\hfill
    \includegraphics[width=0.2\textwidth, height=110pt]{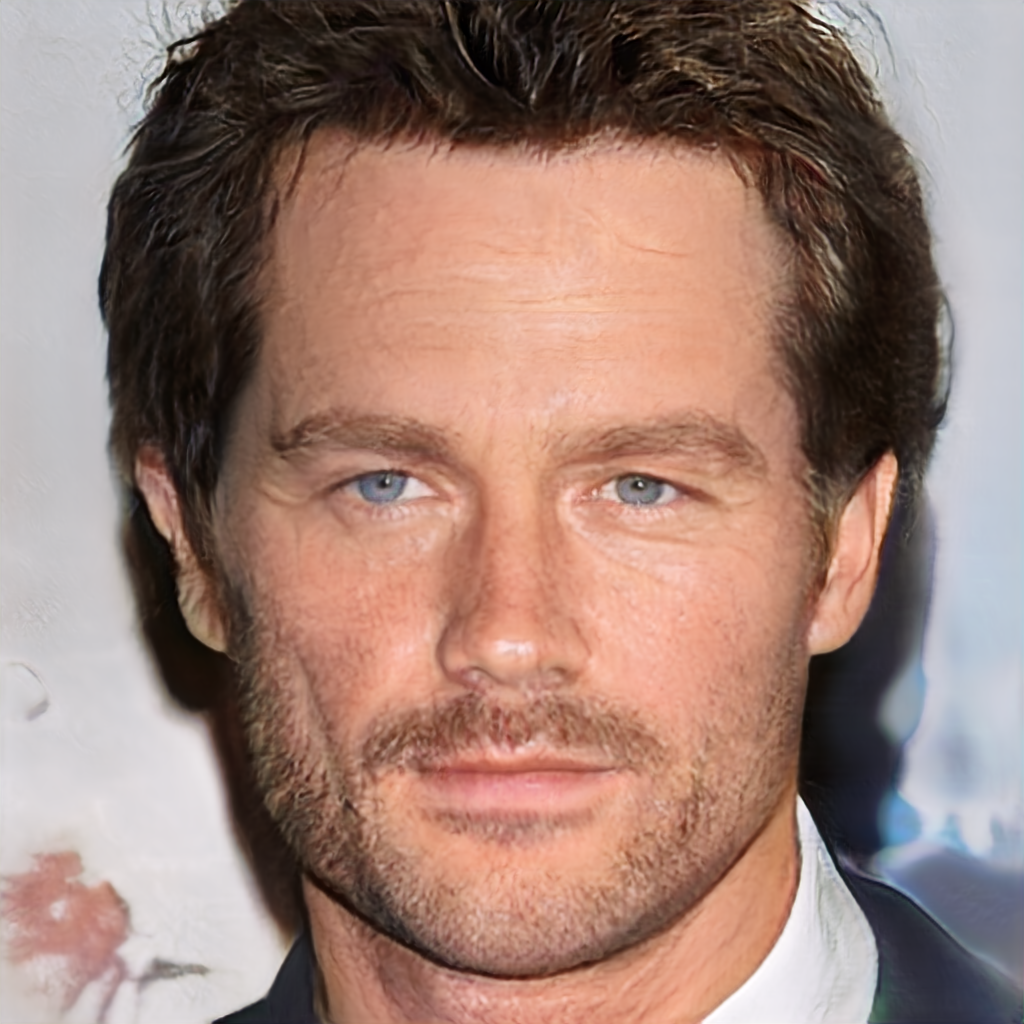}\hfill
    \includegraphics[width=0.2\textwidth, height=110pt]{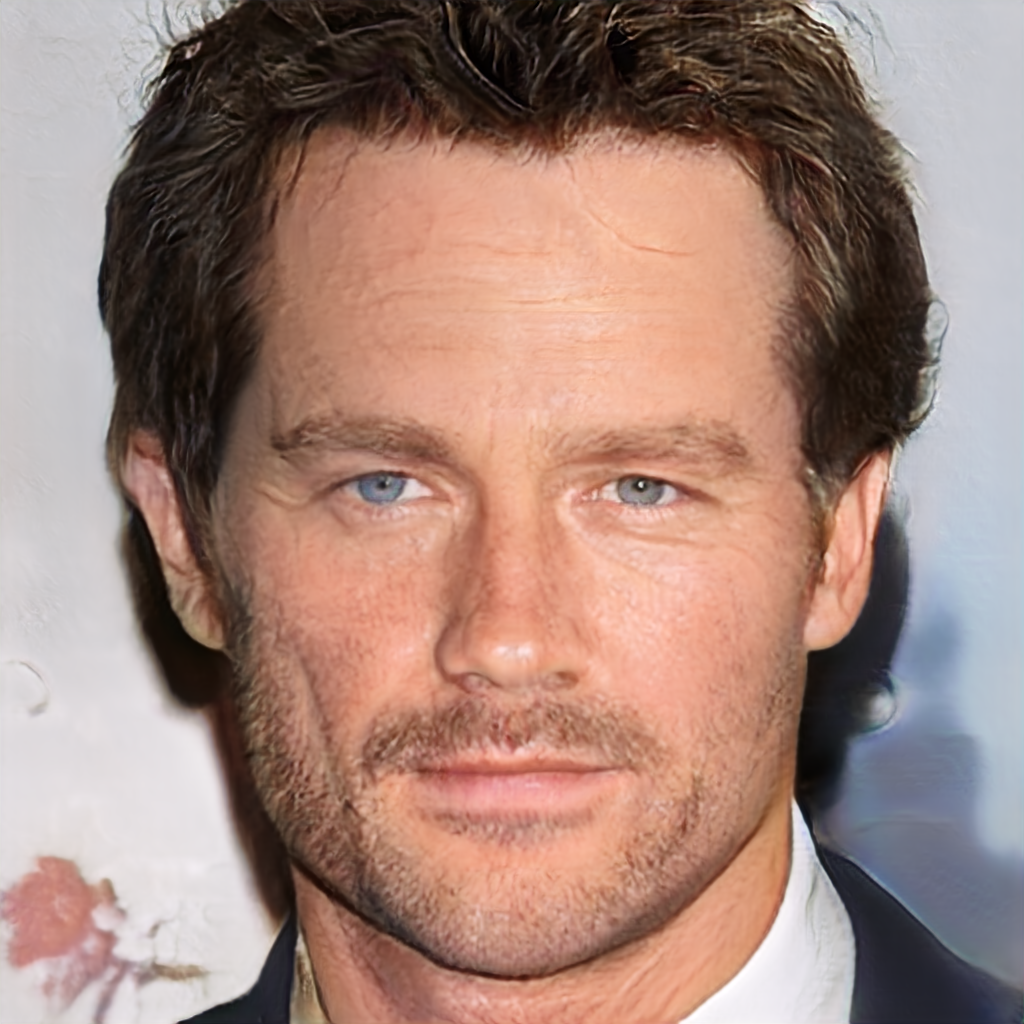}\hfill
    \includegraphics[width=0.2\textwidth, height=110pt]{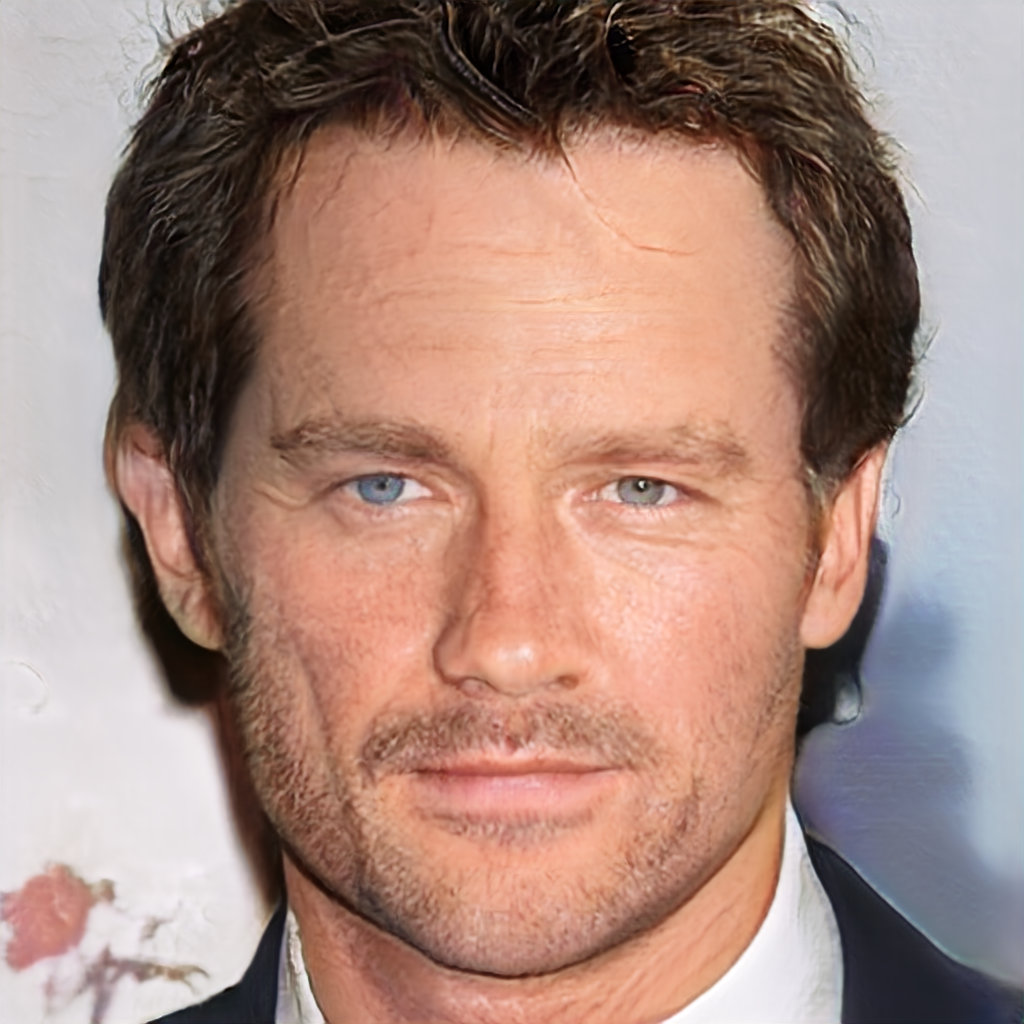}\hfill
    \caption{Interpolation from top left to bottom right generated by PgGAN.}
    \label{pggan_interpolate2}
\end{figure}

\subsection{Circular Interpolation}

\begin{figure}[H]
    \includegraphics[width=0.2\textwidth, height=110pt]{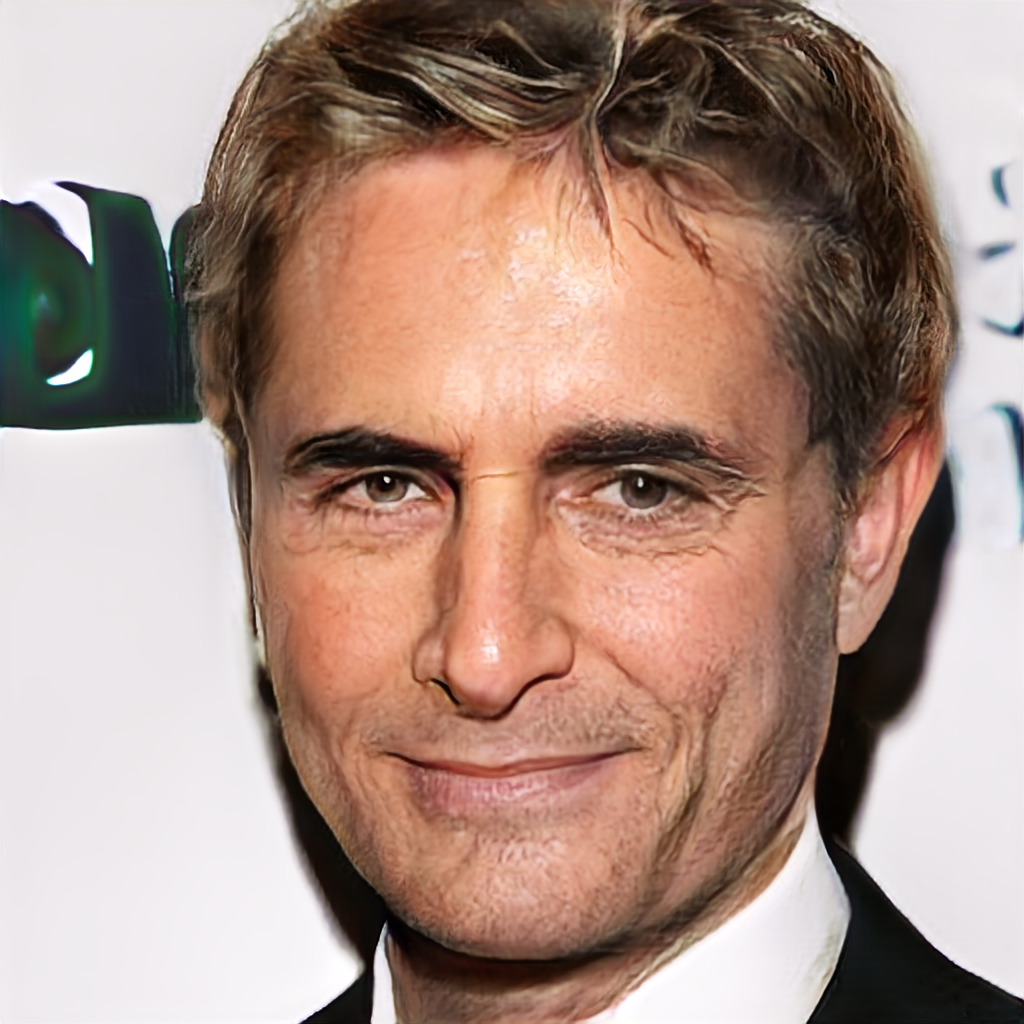}\hfill
    \includegraphics[width=0.2\textwidth, height=110pt]{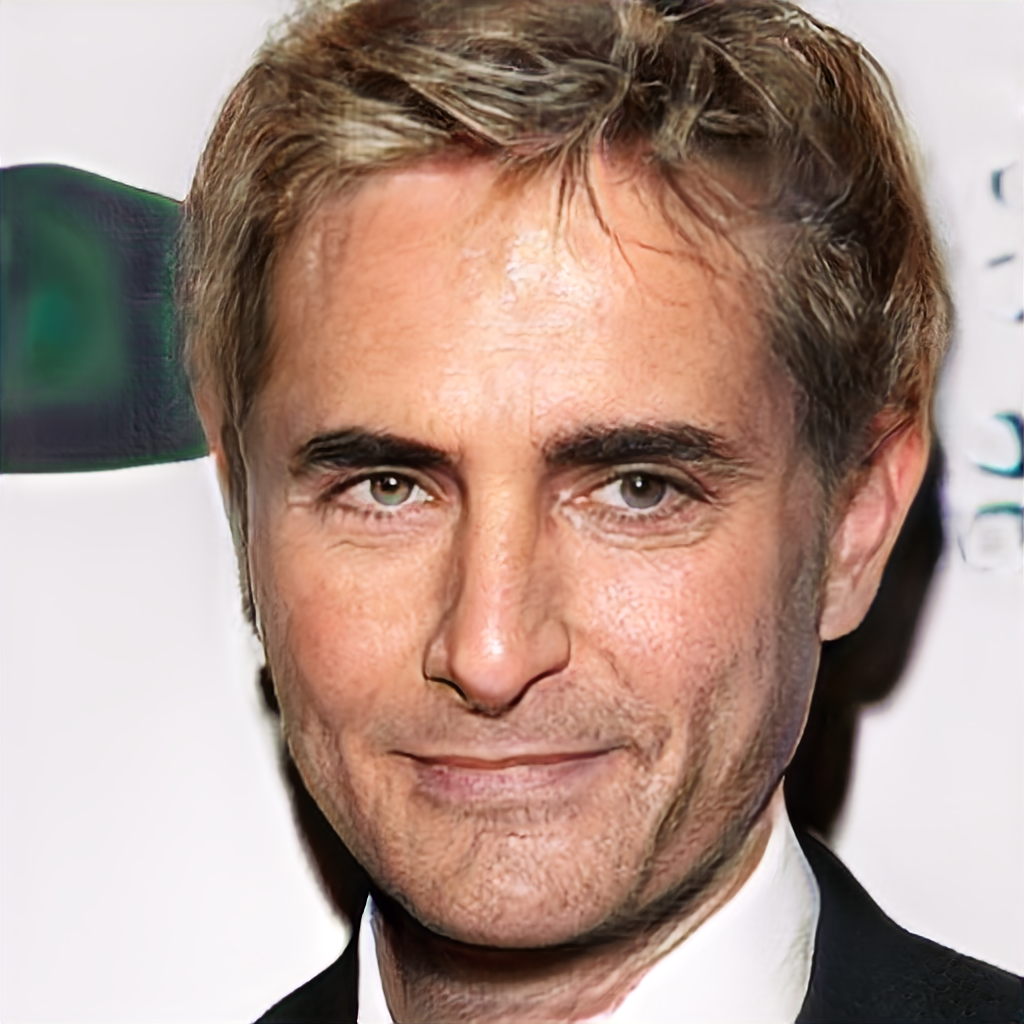}\hfill
    \includegraphics[width=0.2\textwidth, height=110pt]{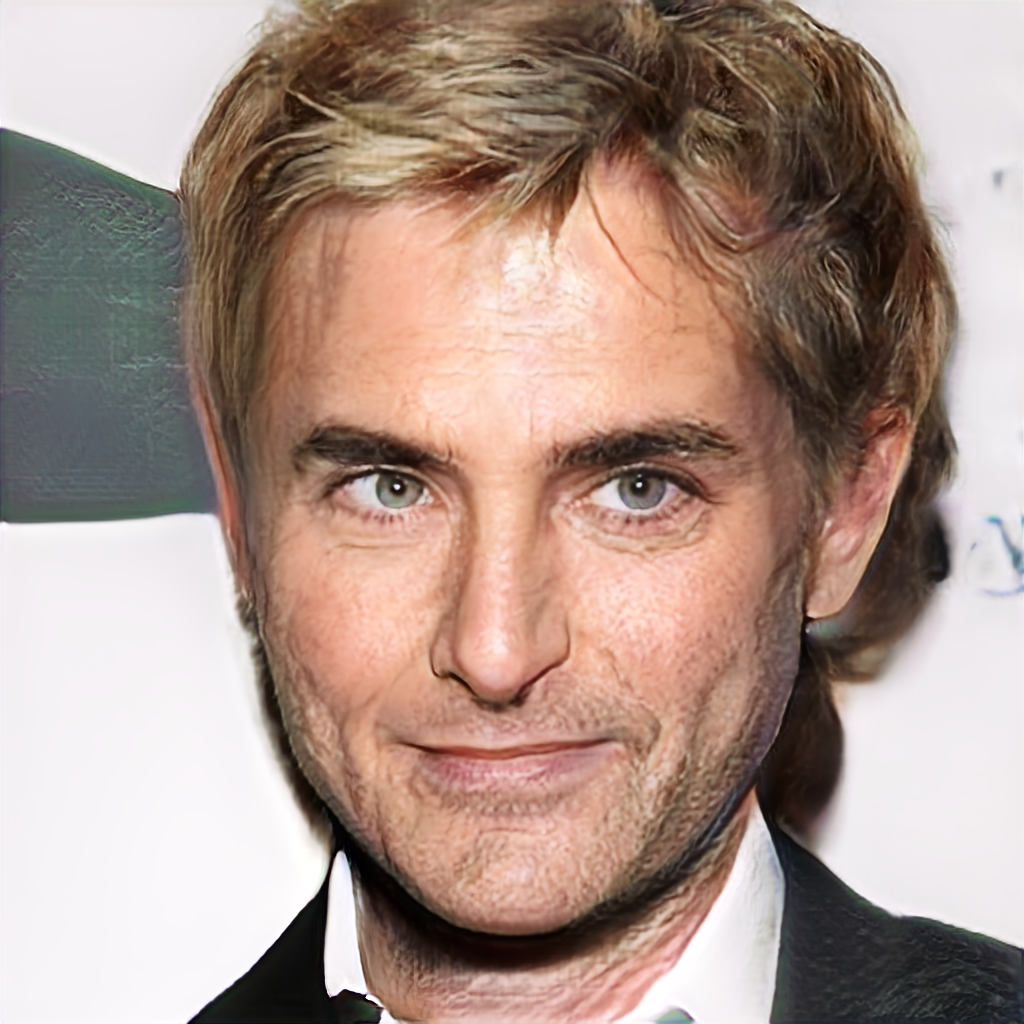}\hfill
    \includegraphics[width=0.2\textwidth, height=110pt]{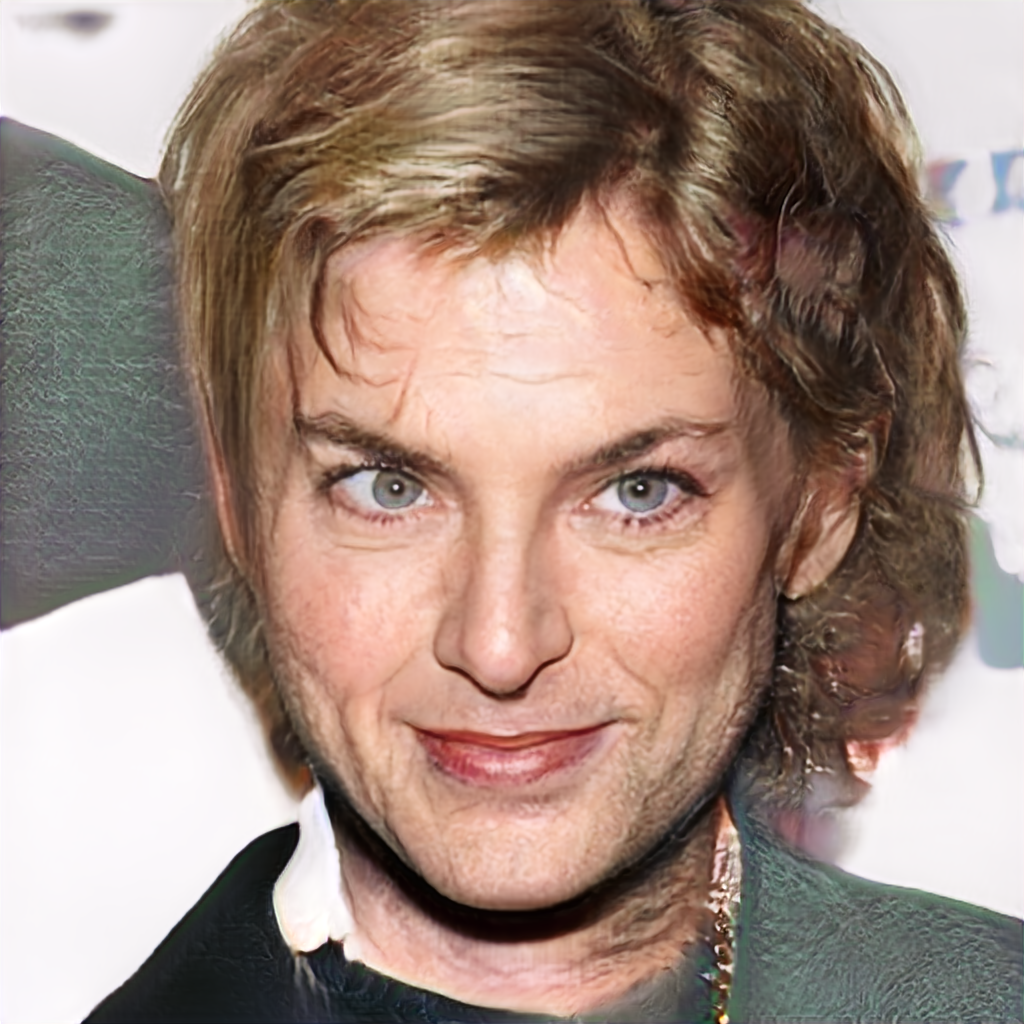}\hfill
    \includegraphics[width=0.2\textwidth, height=110pt]{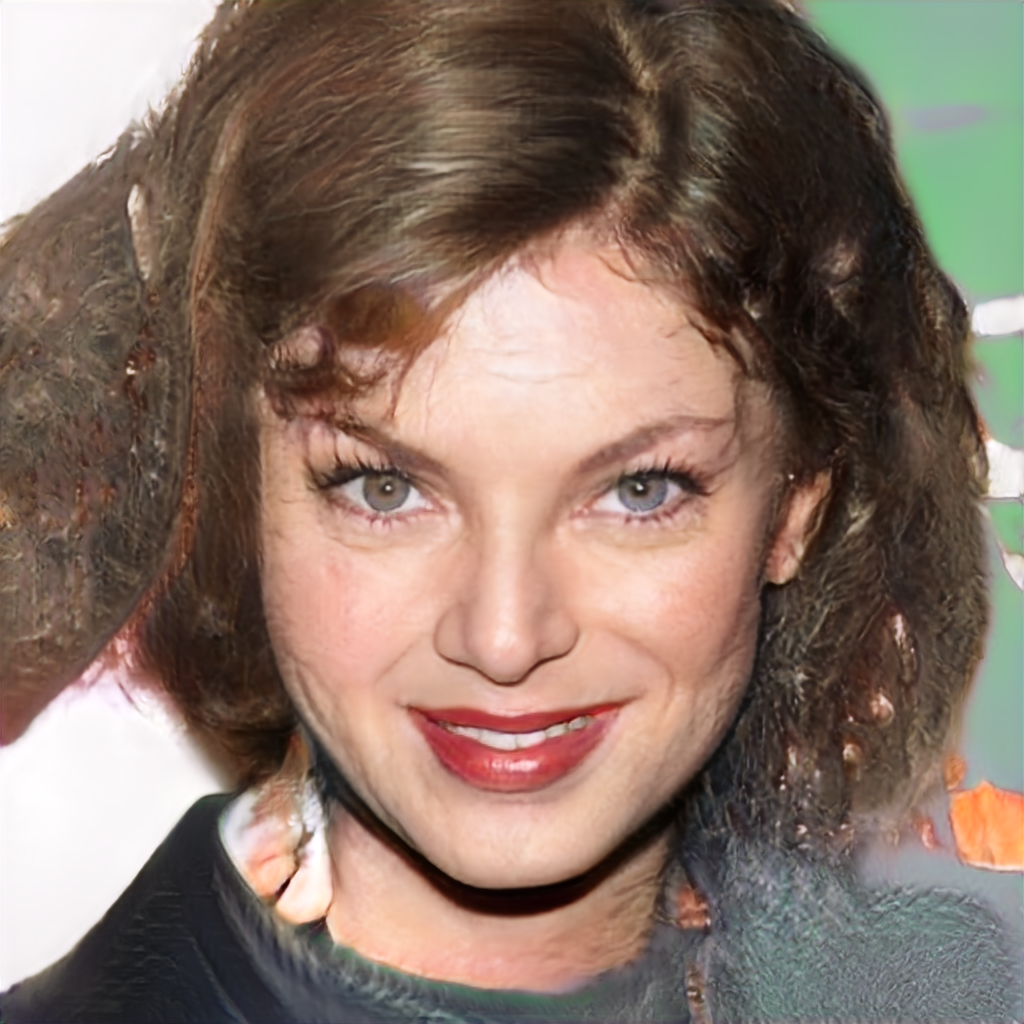}\hfill
    \includegraphics[width=0.2\textwidth, height=110pt]{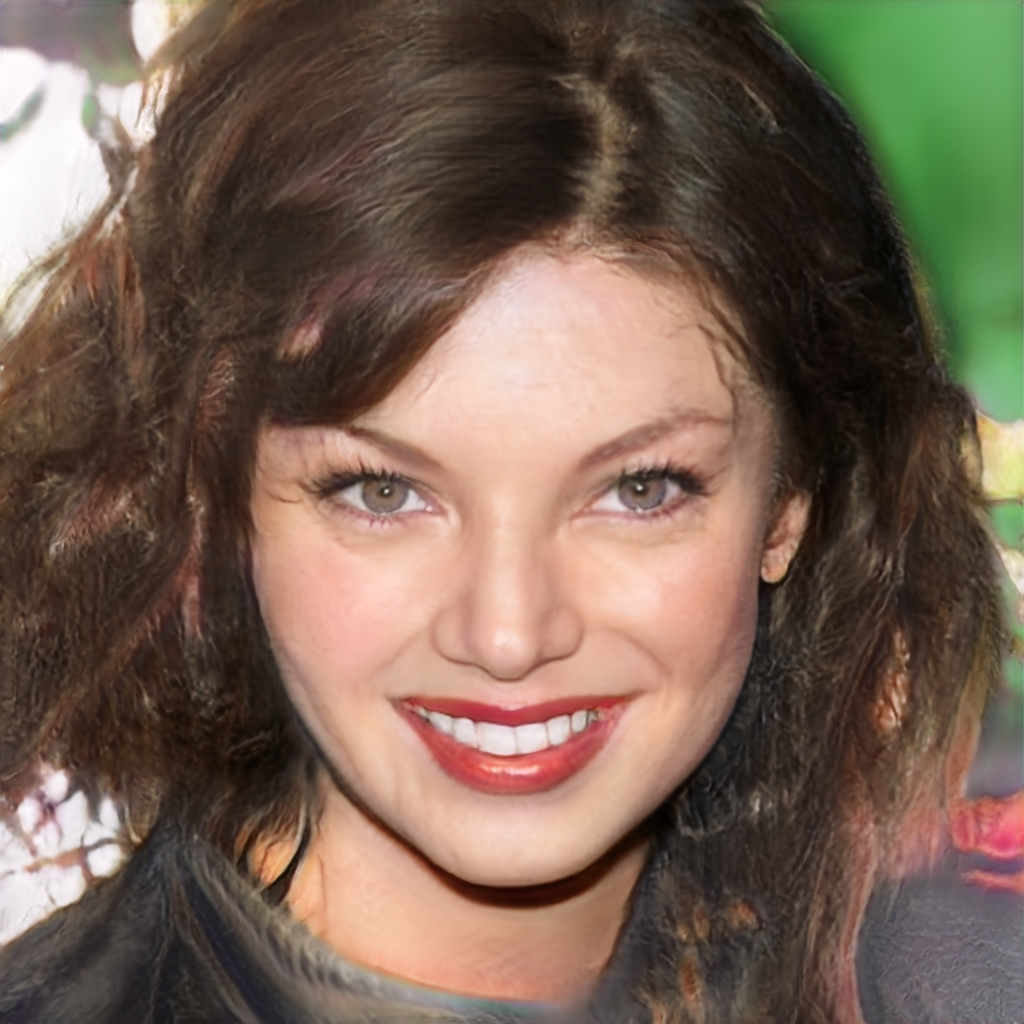}\hfill
    \includegraphics[width=0.2\textwidth, height=110pt]{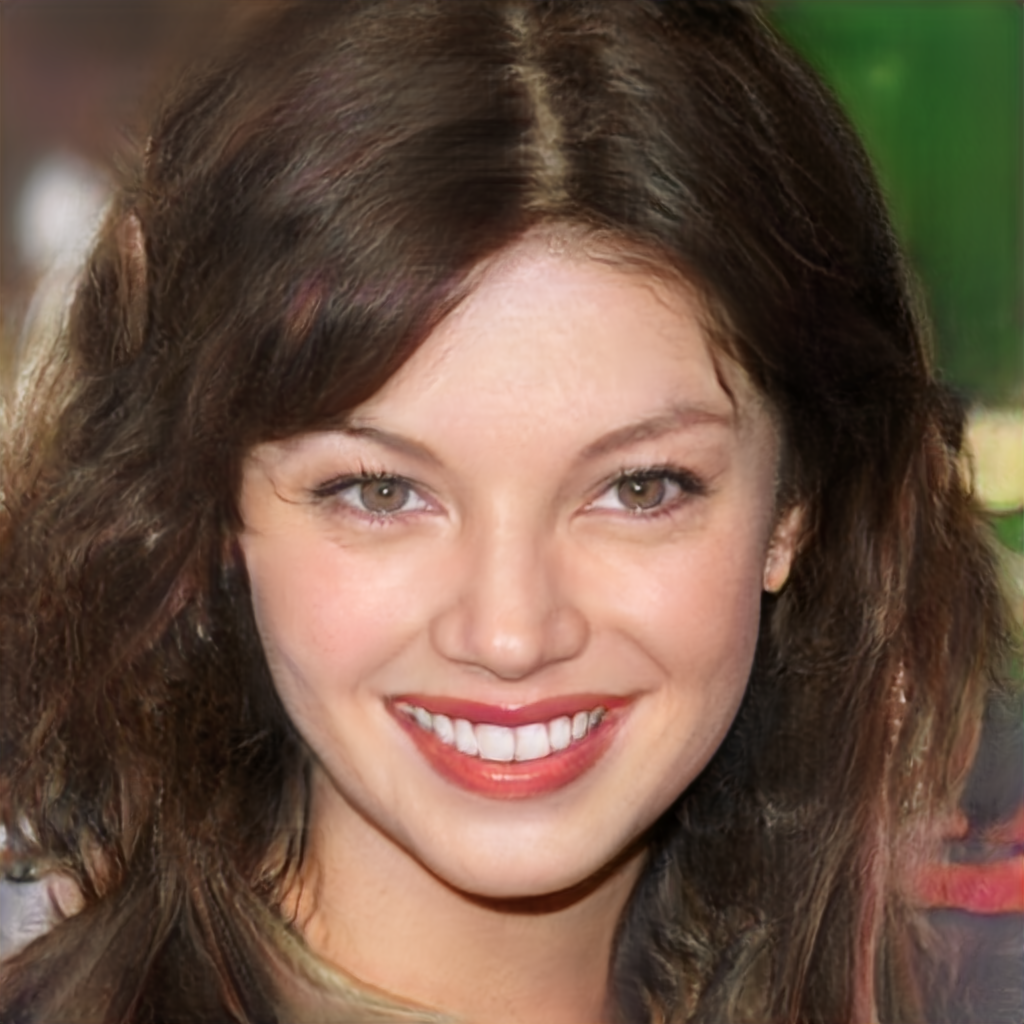}\hfill
    \includegraphics[width=0.2\textwidth, height=110pt]{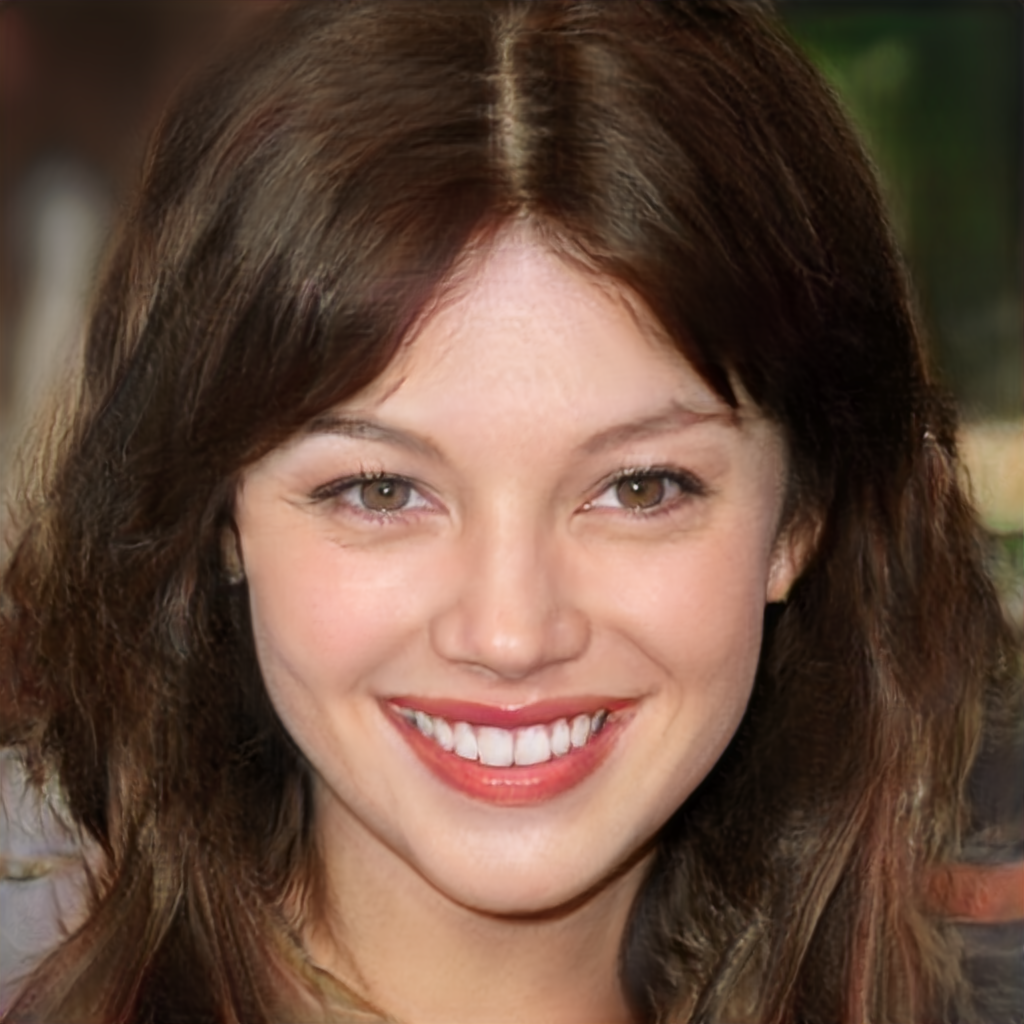}\hfill
    \includegraphics[width=0.2\textwidth, height=110pt]{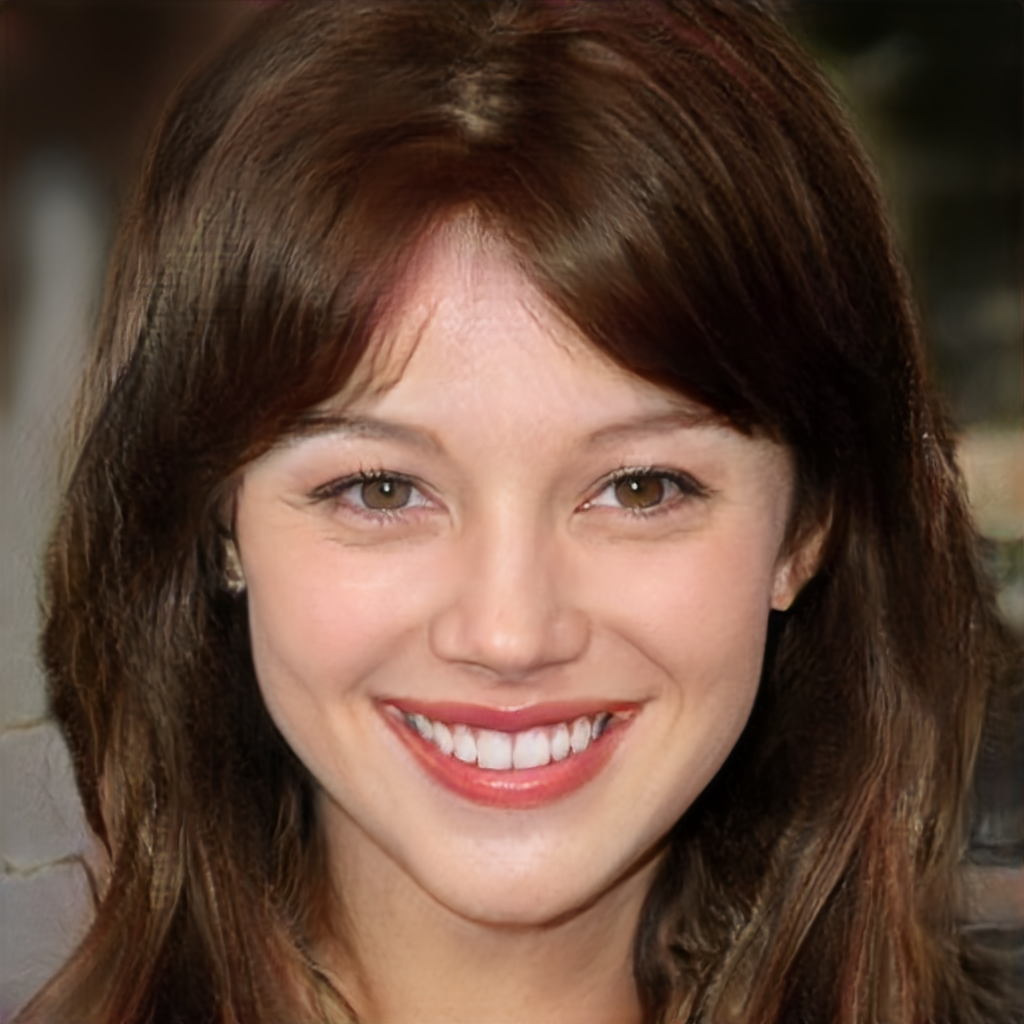}\hfill
    \includegraphics[width=0.2\textwidth, height=110pt]{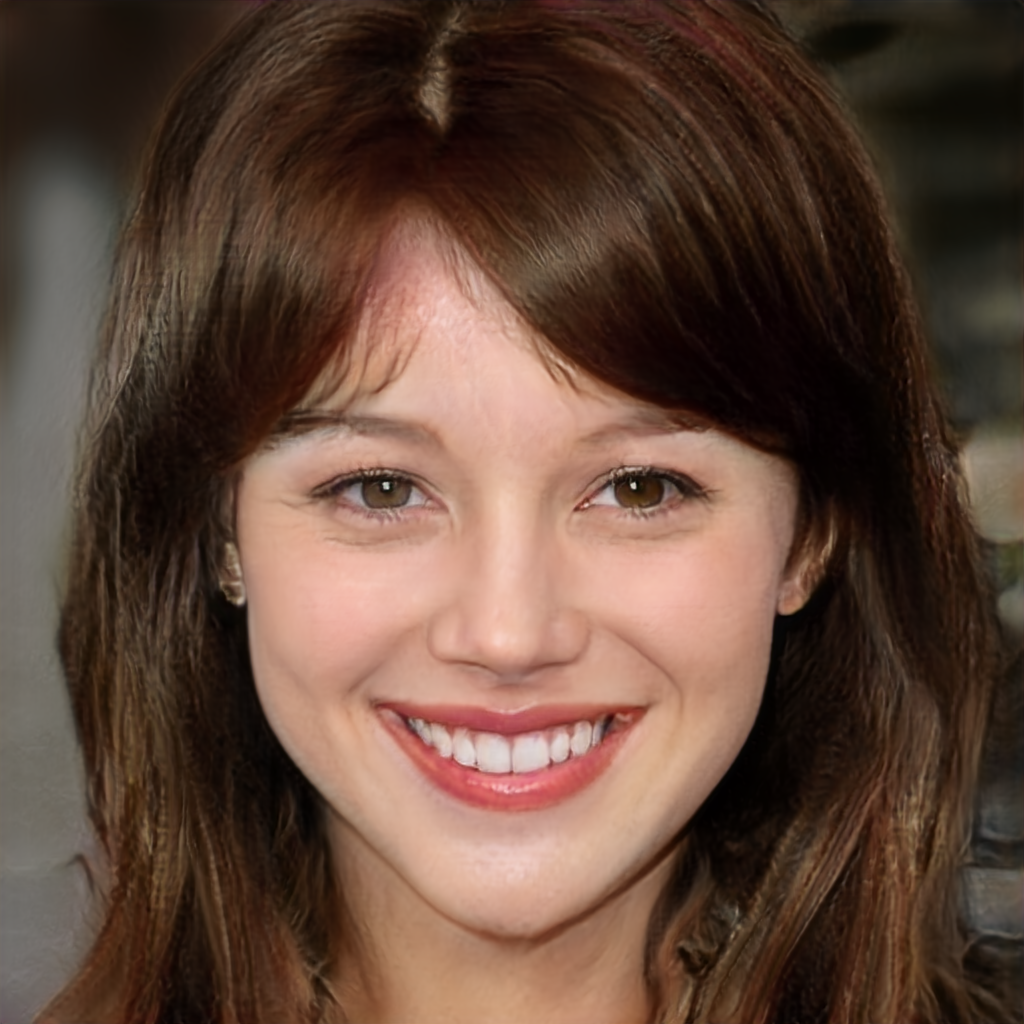}\hfill
    \caption{Circular interpolation over angle of $\pi$ from top left to bottom right generated by PgGAN.}
    \label{pggan_circle_interpolate1}
\end{figure}

\end{document}